\pdfoutput=1

\documentclass{article}

\usepackage[preprint]{neurips_2021}

\usepackage[utf8]{inputenc} 
\usepackage[T1]{fontenc}    

\usepackage{amsfonts}       
\usepackage{nicefrac}       
\usepackage{microtype}      

\usepackage{authblk}
\usepackage{graphicx}
\usepackage{amsmath}
\usepackage[hyphens]{url}
\usepackage{hyperref}
\usepackage{longtable}
\usepackage[justification=justified]{caption}
\usepackage{subcaption}
\usepackage{pbox}
\usepackage{amssymb}
\usepackage{multirow}
\usepackage[table]{xcolor}
\usepackage{natbib}
\usepackage[symbol]{footmisc}
\usepackage{array}
\usepackage{booktabs}
\usepackage{lineno}

\definecolor{orange1}{rgb}{0.99, 0.66, 0.09}
\definecolor{orange2}{rgb}{1, 0.7333, 0.3294}
\definecolor{orange3}{rgb}{1, 0.9237, 0.7804}
\definecolor{orange4}{rgb}{0.97647, 0.76863, 0.6}
\definecolor{orange5}{rgb}{0.98824, 0.92157, 0.93725}
\definecolor{blue1}{rgb}{0.0667, 0.576, 0.941}
\definecolor{blue2}{rgb}{0.290, 0.741, 0.8784}
\definecolor{blue3}{rgb}{0.8667, 0.9216, 0.9529}
\definecolor{blue4}{rgb}{0.57255, 0.80392, 0.86275}
\definecolor{blue5}{rgb}{0.85882, 0.93333, 0.95294}
\definecolor{blue6}{rgb}{0.02353, 0, 0.51373}
\definecolor{grey1}{rgb}{0.867, 0.894, 0.941}
\definecolor{grey2}{rgb}{0.9608, 0.9294, 0.9608}
\definecolor{grey3}{rgb}{0.717, 0.744, 0.791}
\definecolor{grey4}{rgb}{0.8108, 0.7794, 0.8108}
\definecolor{grey5}{rgb}{0.9008, 0.9008, 0.9008}
\definecolor{green1}{rgb}{0.682, 0.902, 0.851}
\definecolor{green2}{rgb}{0.537, 0.8, 0.741}
\definecolor{yellow1}{rgb}{0.9803, 1, 0.8392}
\definecolor{pink1}{rgb}{1, 0.8314, 0.8391}
\definecolor{red1}{rgb}{0.85098, 0.58431, 0.56078}
\definecolor{white}{rgb}{1, 1, 1}

\newcolumntype{x}{>{\columncolor{grey1}}m}
\newcolumntype{y}{>{\columncolor{white}}m}

\hypersetup{
    colorlinks=true,
    linkcolor=blue6,
    filecolor=blue6, 
    citecolor=blue6,
    urlcolor=blue6,
}

\title{Recent Advances in Deep Learning Based Dialogue Systems: A Systematic Survey}

\author[1]{Jinjie Ni}
\author[1]{Tom Young\protect\footnotemark[1]\ \ }
\author[1]{Vlad Pandelea\protect\footnotemark[1]\ \ }
\author[1]{Fuzhao Xue}
\author[1]{Erik Cambria\protect\footnotemark[3]\ \ }

\affil[1]{Nanyang Technological University, Singapore. \{jinjie001, yang0552, fuzhao001\}@e.ntu.edu.sg, \{vlad.pandelea, cambria\}@ntu.edu.sg}

\begin{document}
\addtocounter{footnote}{1}
\footnotetext{Equal contribution}
\addtocounter{footnote}{2}
\footnotetext{Corresponding author}
\addtocounter{footnote}{-3}
\renewcommand*{\thefootnote}{\arabic{footnote}}

\maketitle

\begin{abstract}
Dialogue systems are a popular natural language processing (NLP) task as it is promising in real-life applications. It is also a complicated task since many NLP tasks deserving study are involved. As a result, a multitude of novel works on this task are carried out, and most of them are deep learning based due to the outstanding performance. In this survey, we mainly focus on the deep learning based dialogue systems. We comprehensively review state-of-the-art research outcomes in dialogue systems and analyze them from two angles: model type and system type. Specifically, from the angle of model type, we discuss the principles, characteristics, and applications of different models that are widely used in dialogue systems. This will help researchers acquaint these models and see how they are applied in state-of-the-art frameworks, which is rather helpful when designing a new dialogue system. From the angle of system type, we discuss task-oriented and open-domain dialogue systems as two streams of research, providing insight into the hot topics related. Furthermore, we comprehensively review the evaluation methods and datasets for dialogue systems to pave the way for future research. Finally, some possible research trends are identified based on the recent research outcomes. To the best of our knowledge, this survey is the most comprehensive and up-to-date one at present for deep learning based dialogue systems, extensively covering the popular techniques\footnote{The frameworks, topics, and datasets discussed are originated from the extensive literature review of state-of-the-art research. We have tried our best to cover all but may still omit some works. Readers are welcome to provide suggestions regarding the omissions and mistakes in this article. We also intend to update this article with time as and when new approaches or definitions are proposed and used by the community}. We speculate that this work is a good starting point for academics who are new to the dialogue systems or those who want to quickly grasp up-to-date techniques in this area.\\

\end{abstract}
\newpage

\section{Introduction}
\label{intro}
Dialogue systems (or chatbots) are playing a bigger role in the world. People may still have a stereotype that chatbots are those rigid agents in their phone calls to a bank. However, thanks to the revival of artificial intelligence, the modern chatbots can converse with rich topics ranging from your birthday party to a speech given by Biden, and, if you want, they can even book a place for your party or play the speech video. At present, dialogue systems are one of the hot topics in NLP and are highly demanded in industry and daily life. The market size of chatbot is projected to grow from \$2.6 billion in 2021 to \$9.4 billion by 2024 at a compound annual growth rate (CAGR) of 29.7\% \footnote{Statistic source: \url{https://markets.businessinsider.com}} and 80\% of businesses are expected to be equipped with chatbot automation by the end of 2021 \footnote{Statistic source: \url{https://outgrow.co}}.

Dialogue systems perform chit-chat with human or serve as an assistant via conversations. By their applications, dialogue systems are commonly divided into two categories: task-oriented dialogue systems (TOD) and open-domain dialogue systems (OOD). Task-oriented dialogue systems solve specific problems in a certain domain such as movie ticket booking, restaurant table reserving, etc. Instead of focusing on task completion, open-domain dialogue systems aim to chat with users without the task and domain restrictions~\citep{ritter2011data}, which are usually fully data-driven. Both task-oriented and open-domain dialogue systems can be seen as a mapping $\varphi$ from user message $U = \{\mathrm{\mathbf{u}}^{(1)}, \mathrm{\mathbf{u}}^{(2)}, ... , \mathrm{\mathbf{u}}^{(i)}\}$ to agent response $R = \{\mathrm{\mathbf{r}}^{(1)}, \mathrm{\mathbf{r}}^{(2)}, ... , \mathrm{\mathbf{r}}^{(j)}\}$: $R = \varphi(U)$, where $\mathrm{\mathbf{u}}^{(i)}$ and $\mathrm{\mathbf{r}}^{(j)}$ denote the $i$th token of the user message and the $j$th token of the agent response respectively. In many open-domain and task-oriented dialogue systems, this mapping also considers a source of external knowledge/database $K$ as input: $R = \varphi(U, K)$. Table \ref{An example of inputs and outputs of task-oriented and open-domain dialogue systems in datasets} presents examples of inputs and outputs of task-oriented and open-domain dialogue systems. More specific details and works will be discussed in Section \ref{Task-oriented Dialogue Systems} and \ref{Open-Domain Dialogue Systems}.

\begin{table}[h]
\caption{Examples of inputs and outputs of task-oriented and open-domain dialogue systems in datasets. Some datasets provide external knowledge annotations for each dialogue pair, e.g., in task-oriented dialogue systems, the external knowledge can be retrieved from restaurant databases; in open-domain dialogue systems, it can be retrieved from commonsense knowledge graphs (KG).}

\label{An example of inputs and outputs of task-oriented and open-domain dialogue systems in datasets} 
\centering
\begin{tabular}{m{2cm}m{3cm}m{3cm}m{3.7cm}}
\bottomrule
\multicolumn{1}{c}{\textbf{Category}} & \multicolumn{1}{c}{\textbf{User message ($U$)}} & \multicolumn{1}{c}{\textbf{Agent response ($R$)}} & \multicolumn{1}{c}{\textbf{External Knowledge ($K$)}}\\\midrule
\multicolumn{1}{c}{Task-oriented} & I need to find a nice restaurant in Madrid that serves expensive Thai food. & There is a restaurant called \textit{Bangkok City} locating at 9 Red Ave. & \multicolumn{1}{c}{restaurant database} \\\hline
\multicolumn{1}{c}{Open-domain} & I love the grilled fish so much! & Yeah. it's a famous \textit{Chinese dish}. & \multicolumn{1}{c}{commonsense KG} \\
\bottomrule
\end{tabular}
\end{table}

Traditional task-oriented dialogue systems are organized in a pipeline structure and consist of four functional modules: Natural Language Understanding, Dialogue State Tracking, Policy Learning, and Natural Language Generation, which will be discussed in detail in Section~\ref{Task-oriented Dialogue Systems}. Many state-of-the-art works design end-to-end task-oriented dialogue systems to achieve better optimization compared with pipeline methods. Open-domain dialogue systems are generally divided into three categories: generative systems, retrieval-based systems, and ensemble systems. Generative systems apply sequence-to-sequence models (see Section \ref{Vanilla Sequence-to-sequence Models (Encoder-decoder Models)}) to map the user message and dialogue history into a response sequence that may not appear in the training corpus. By contrast, retrieval-based systems try to select a pre-existing response from a certain response set. Ensemble systems combine generative methods and retrieval-based methods in two ways: retrieved responses can be compared with generated responses to choose the best among them; generative models can also be used to refine the retrieved responses~\citep{zhu2018retrieval, song2016two, qiu2017alime, serban2017deep}. Generative systems can produce flexible and dialogue context-related responses while sometimes they lack coherence \footnote{The quality of being logical and consistent not only between words/subwords but also between responses of different timesteps.} and tend to make dull responses \citep{serban2016building, vinyals2015neural, sordoni2015neural}. Retrieval-based systems select responses from human response sets and thus are able to achieve better coherence in surface-level language. However, retrieval systems are restricted by the finiteness of the response sets and sometimes the responses retrieved show a weak correlation with the dialogue context~\citep{zhu2018retrieval}. 

\begin{figure}[t]
\begin {center}
\includegraphics[width=1.0\textwidth]{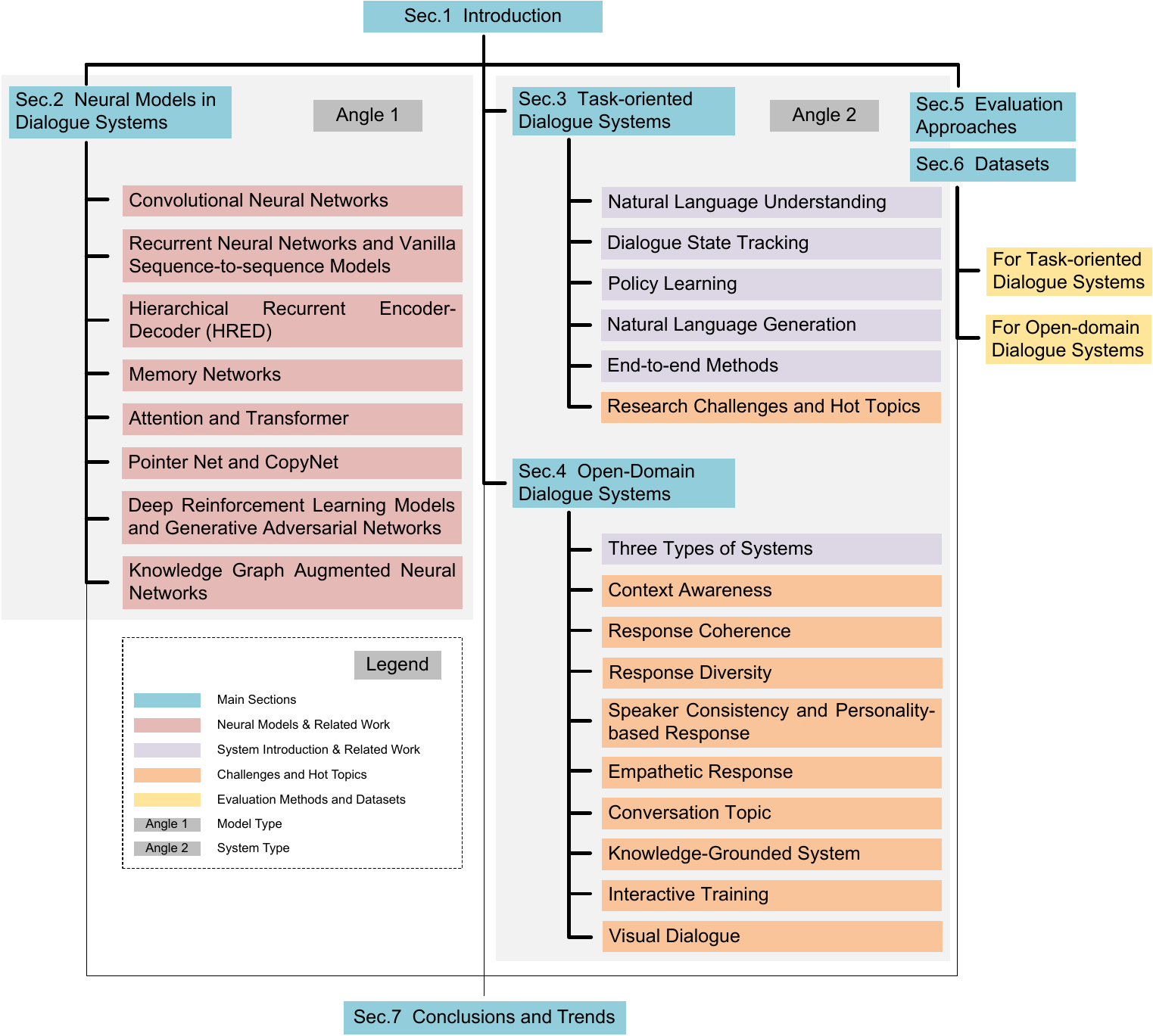}
\caption{The overall diagram of this article}
\label{The overall diagram of this article}
\end {center}
\end{figure}

For dialogue systems, existing surveys~\citep{arora2013dialogue, wang2016recent, mallios2016survey, chen2017survey, gao2018neural} are either outdated or not comprehensive. Some definitions in these papers are no longer being used at present, and a lot of new works and topics are not covered. In addition, most of them lack a multi-angle analysis. Thus, in this survey, we comprehensively review high-quality works in recent years with a focus on deep learning-based approaches and provide insight into state-of-the-art research from both model angle and system angle. Moreover, this survey updates the definitions/names according to state-of-the-art research. E.g., we name "open-domain dialogue systems" instead of "chit-chat dialogue systems" because most of the articles (roughly 70\% according to our survey) name them as the prior one. We also extensively cover the diverse hot topics in dialogue systems and extend some new topics that are popular in current research community (such as Domain Adaptation, Dialogue State Tracking Efficiency, End-to-end methods for task-oriented dialogue systems; Controllable Generation, Interactive Training, and Visual Dialogue for open-domain dialogue systems).

Traditional dialogue systems are mostly rule-based~\citep{arora2013dialogue} and non-neural machine learning based systems. Rule-based systems are easy to implement and can respond naturally, which contributed to their popularity in earlier industry products. However, the dialogue flows of these systems are predetermined, which keeps the applications of the dialogue systems within certain scenarios. Non-neural machine learning based systems usually perform template filling to manage certain tasks. These systems are more flexible compared with rule-based systems because the dialogue flows are not predetermined. However, they cannot achieve high F1 scores~\citep{Powers2020EvaluationFP} in template filling\footnote{Template filling is an efficient approach to extract and structure complex information from text to fill in a pre-defined template. They are mostly used in task-oriented dialogue systems.} and are also restricted in application scenarios and response diversity because of the fixed templates. Most if not all state-of-the-art dialogue systems are deep learning-based systems (neural systems). The rapid growth of deep learning improves the performance of dialogue systems~\citep{chen2017survey}. Deep learning can be viewed as representation learning with multilayer neural networks. Deep learning architectures are widely used in dialogue systems and their subtasks. Section~\ref{Neural Models in Dialogue Systems} discusses various popular deep learning architectures. 

Apart from dialogue systems, there are also many dialogue-related tasks in NLP, including but not limited to question answering, reading comprehension, dialogue disentanglement, visual dialogue, visual question answering, dialogue reasoning, conversational semantic parsing, dialogue relation extraction, dialogue sentiment analysis, hate speech detection, MISC detection, etc. In this survey, we also touch on some works tackling these dialogue-related tasks, since the design of dialogue systems can benefit from advances in these related areas. 

We produced a diagram for this article to help readers familiarize the overall structure (Figure~\ref{The overall diagram of this article}). In this survey, Section~\ref{intro} briefly introduces dialogue systems and deep learning; Section~\ref{Neural Models in Dialogue Systems} discusses the neural models popular in modern dialogue systems and the related work; Section~\ref{Task-oriented Dialogue Systems} introduces the principles and related work of task-oriented dialogue systems and discusses the research challenges and hot topics; Section~\ref{Open-Domain Dialogue Systems} briefly introduces the three kinds of systems and then focuses on hot topics in open-domain dialogue systems; Section~\ref{Evaluation Approaches} reviews the main evaluation methods for dialogue systems; Section~\ref{Data sets} comprehensively summarizes the datasets commonly used for dialogue systems; finally, Section~\ref{Conclusions and Trends} concludes the paper and provides some insight on research trends. 

\section{Neural Models in Dialogue Systems}
\label{Neural Models in Dialogue Systems}
In this section, we introduce neural models that are popular in state-of-the-art dialogue systems and related subtasks. We also discuss the applications of these models or their variants in modern dialogue systems research to provide readers with a picture from the model's perspective. This will help researchers acquaint these models and see how they are applied in state-of-the-art frameworks, which is rather helpful when designing a new dialogue system. The models discussed include: Convolutional Neural Networks (CNNs), Recurrent Neural Networks (RNNs), Vanilla Sequence-to-sequence Models, Hierarchical Recurrent Encoder-Decoder (HRED), Memory Networks, Attention Networks, Transformer, Pointer Net and CopyNet, Deep Reinforcement Learning models, Generative Adversarial Networks (GANs), Knowledge Graph Augmented Neural Networks. We start from some classical models (e.g., CNNs and RNNs), and readers who are familiar with their principles and corresponding applications in dialogue systems can choose to read selectively.
\subsection{Convolutional Neural Networks}
\label{Convolutional Neural Network}
Deep neural networks have been considered as one of the most powerful models. `Deep' refers to the fact that they are multilayer, which extracts features by stacking feed-forward layers. Feed-forward layers can be defined as: $y = \sigma(Wx + b)$. Where the $\sigma$ is an activation function; $W$ and $b$ are trainable parameters. The feed-forward layers are powerful due to the activation function, which makes the otherwise linear operation, non-linear. Whereas there exist some problems when using feed-forward layers. Firstly, the operations of feed-forward layers or multilayer neural networks are just template matching, where they do not consider the specific structure of data. Furthermore, the fully connected mechanism of traditional multilayer neural networks causes an explosion in the number of parameters and thus leads to generalization problems. \cite{lecun1998gradient} proposed LeNet-5, an early CNN. The invention of CNNs mitigates the above problems to some extent.

\begin{figure}
\begin {center}
\includegraphics[width=0.65\textwidth]{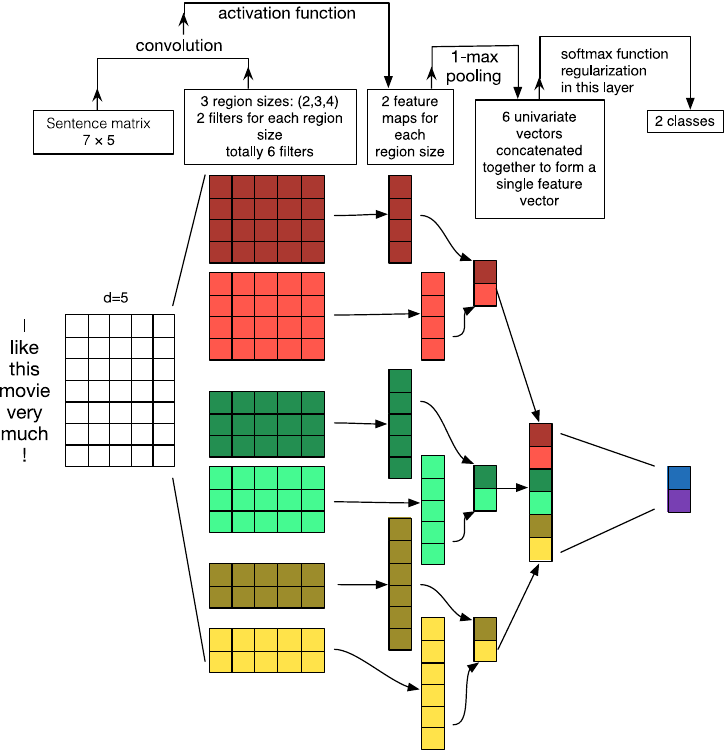}
\caption{A CNN architecture for text classification \citep{zhang-wallace-2017-sensitivity}}
\label{Illustration of a CNN architecture for text classification}
\end {center}
\end{figure} 

CNNs (Figure~\ref{Illustration of a CNN architecture for text classification}) usually consist of convolutional layers, pooling layers and feed-forward layers. Convolutional layers apply convolution kernels to perform the convolution operation: 
\begin{equation}
G(m, n) = (f*h)(m, n) = \sum_{j}\sum_{k}h(j, k)f(m-j, n-k)
\label{1}
\end{equation}

Where $m$ and $n$ are respectively the indexes of rows and columns of the result matrix. $f$ denotes the input matrix and $h$ denotes the convolutional kernel. The pooling layers perform down-sampling on the result of convolutional layers to get a higher level of features and the feed-forward layers map them into a probability distribution to predict class scores. 

A sliding window feature enables convolution layers to capture local features and the pooling layers can produce hierarchical features. These two mechanisms give CNNs the local perception and global perception ability, helping to capture some specific inner structures of data. The parameter sharing mechanism eases the parameter explosion problem and overfitting problem because the reduction of trainable parameters leads to less model complexity, improving the generalization ability. 

Due to these good properties, CNNs have been widely applied in many works. Among them, the Computer Vision tasks benefit the most for that the Spatio-temporal data structures of images or videos are perfectly captured by CNNs. For more detailed mechanism illustrations and other variants of CNNs, readers can refer to these representative algorithm papers or surveys:~\citep{krizhevsky2012imagenet, zeiler2014visualizing, simonyan2014very, szegedy2015going, he2016deep, aloysius2017review, rawat2017deep}. In this survey, we focus on dialogue systems.

Recent years have seen a dramatic increase in applications of CNNs in NLP. Many tasks take words as basic units. However, phrases, sentences, or even paragraphs are also useful to semantic representations. As a result, CNNs are an ideal tool for the hierarchical modeling of language~\citep{conneau2016very}. \\

CNNs are good textual feature extractors, but they may not be ideal sequential encoders. Some dialogue systems~\citep{qiu2019training, bi2019fine, ma2020conversational} directly used CNNs as the encoder of utterances or knowledge, but most of the state-of-the-art dialogue systems such as~\cite{feng2019learning, wu2016sequential, tao2019one, wang2019persuasion, chauhan2019ordinal, feldman2019multi, chen2019working, lu2019constructing} and~\cite{coope2020span} chose to use CNNs as a hierarchical feature extractor after encoding the text information, instead of directly applying them as encoders. This is due to the fixed input length and limited convolution span of CNNs. Generally, there are two main situations where CNNs are used to process encoded information in dialogue systems. The first situation is applying CNNs to extract features directly based on the feature vectors from the encoder~\citep{wang2019persuasion, chauhan2019ordinal, feldman2019multi, chen2019working} and~\cite{coope2020span}. Within the works above,~\cite{feldman2019multi} extracted features from character-level embeddings, illustrating the hierarchical extraction capability of CNNs. Another situation in which CNNs are used is extracting feature maps in response retrieval tasks. Some works built retrieval-based dialogue systems~\citep{wu2016sequential, feng2019learning, tao2019one, lu2019constructing}. They used separate encoders to encode dialogue context and candidate responses and then used a CNN as an extractor of the similarity matrix calculated from the encoded dialogue context and candidate responses. Their experiments showed that this method can achieve good performance in response retrieval tasks. 

The main reason why more recent works do not choose CNNs as dialogue encoders is that they fail to extract the information across temporal sequence steps continuously and flexibly~\citep{krizhevsky2012imagenet}. Some models introduced later do not process data points independently, which are desirable models for encoders.

\subsection{Recurrent Neural Networks and Vanilla Sequence-to-sequence Models}
\label{RNNs and Vanilla Sequence-to-sequence Models}
NLP tasks including dialogue-related tasks try to process and analyze sequential language data points. Even though standard neural networks, as well as CNNs, are powerful learning models, they have two main limitations~\citep{lipton2015critical}. One is that they assume the data points are independent of each other. While it is reasonable if the data points are produced independently, essential information can be missed when processing interrelated data points (e.g., text, audio, video). Additionally, their inputs are usually of fixed length, which is a limitation when processing sequential data varying in length. Thus, a sequential model being able to represent the sequential information flow is desirable. 

Markov models like Hidden Markov Models (HMMs) are traditional sequential models, but due to the time complexity of the inference algorithm~\citep{viterbi1967error} and because the size of transition matrix grows significantly with the increase of the discrete state space, in practice they are not applicable in dealing with problems involving large possible hidden states. The property that the hidden states of Markov models are only affected by the immediate hidden states further limits the power of this model.

RNN models are not proposed recently, but they greatly solve the above problems and some variants can amazingly achieve state-of-the-art performance in dialogue-related tasks as well as many other NLP tasks. The inductive bias of recurrent models is non-replaceable in many scenarios, and many up-to-date models incorporate the recurrence.

\subsubsection{Jordan-Type and Elman-Type RNNs}

In 1982, Hopfield introduced an early family of RNNs to solve pattern recognition tasks~\citep{hopfield1982neural}. \cite{jordan1986serial} and~\cite{elman1990finding} introduced two kinds of RNN architectures respectively. Generally, modern RNNs can be classified into Jordan-type RNNs and Elman-type RNNs. 

\begin{figure}
\begin {center}
     \begin{subfigure}[b]{0.48\textwidth}
         \centering
         \includegraphics[width=0.6\textwidth]{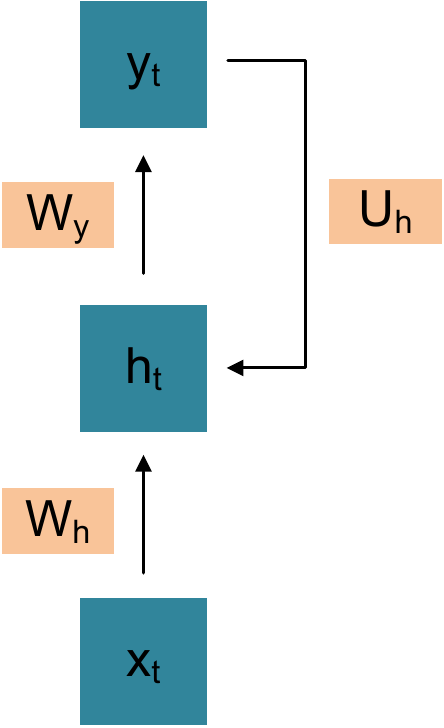}
         \caption{Jordan-type RNNs}
         \label{Jordan type RNNs}
     \end{subfigure}
\hfill
     \begin{subfigure}[b]{0.48\textwidth}
         \centering
         \includegraphics[width=0.6\textwidth]{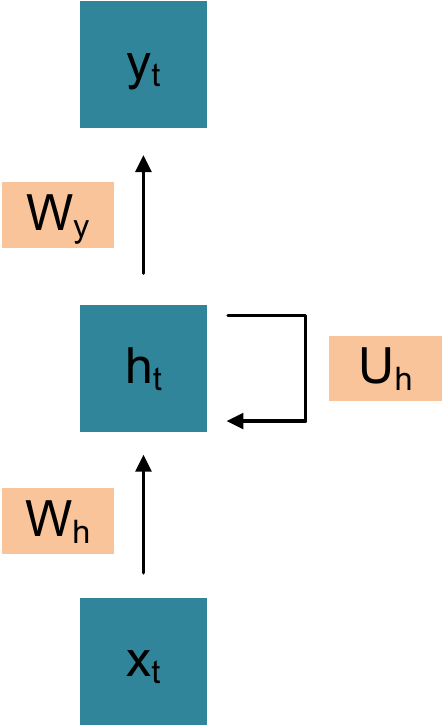}
         \caption{Elman-type RNNs}
         \label{Elman type RNNs}
     \end{subfigure}
\caption{Graphical models of two basic types of RNNs}
\label{Two basic types of RNNs}
\end {center}
\end{figure}

The Jordan-type RNNs are shown in Figure~\ref{Jordan type RNNs}. $x_t$, $h_t$, and $y_{t}$ are the inputs, hidden state, and output of time step $t$, respectively. $W_h$, $W_y$ and $U_h$ are weight matrixes. Each update of hidden state is decided by the current input and the output of last time step while each output is decided by current hidden state. Thus the hidden state and output of time step $t$ can be calculated as: 
\begin{equation}
h_t = \sigma_h (W_h x_t + U_h y_{t-1} + b_h)
\label{2}
\end{equation}
\begin{equation}
y_t = \sigma_y (W_y h_t + b_y)
\label{3}
\end{equation}
Where $b_h$ and $b_y$ are biases. $\sigma_h$ and $\sigma_y$ are activation functions. 

The Elman-type RNNs are shown in Figure~\ref{Elman type RNNs}. The difference is that each hidden state is decided by the current input and the hidden state of last time step. Thus the hidden state and output of time step $t$ can be calculated as: 

\begin{equation}
h_t = \sigma_h (W_h x_t + U_h h_{t-1} + b_h)
\label{4}
\end{equation}
\begin{equation}
y_t = \sigma_y (W_y h_t + b_y)
\label{5}
\end{equation}

Simple RNNs can model long-term dependencies theoretically. But in practical training, long-range dependencies are difficult to learn~\citep{bengio1994learning, hochreiter2001gradient}. When backpropagating errors over many time steps, simple RNNs suffer from problems known as gradient vanishing and gradient explosion~\citep{hochreiter1997long}. Some solutions were proposed to solve these problems~\citep{williams1989learning, pascanu2013difficulty}, which led to the inventions of some variants of traditional recurrent networks.

\subsubsection{LSTM}

~\cite{hochreiter1997long} introduced gate mechanisms in LSTM mainly to address the gradient vanishing problem. Input gate, forget gate and output gate were introduced to decide how much information from new inputs and past memories should be reserved. The model can be described by the following equations: 
\begin{equation}
\hat{h}^{(t)} = tanh\left(W^{\hat{h}x} x^{(t)} + W^{\hat{h}h} h^{(t-1)} + b_{\hat{h}}\right)
\end{equation}
\begin{equation}
i^{(t)} = \sigma\left(W^{ix} x^{(t)} + W^{ih} h^{(t-1)} + b_i\right)
\end{equation}
\begin{equation}
f^{(t)} = \sigma\left(W^{fx} x^{(t)} + W^{fh} h^{(t-1)} + b_f\right)
\end{equation}
\begin{equation}
o^{(t)} = \sigma\left(W^{ox} x^{(t)} + W^{oh} h^{(t-1)} + b_o\right)
\end{equation}
\begin{equation}
s^{(t)} = \hat{h}^{(t)} \odot i^{(t)} + s^{(t-1)} \odot f^{(t)}
\end{equation}
\begin{equation}
h^{(t)} = tanh (s^{(t)})\odot o^{(t)}
\end{equation}

Where $t$ represents time step $t$. $i$, $f$ and $o$ are gates, denoting input gate, forget gate and output gate respectively. $x$, $\hat{h}$, $s$ and $h$ are input, short-term memory, long-term memory and output respectively. $b$ is bias and $W$ is weight matrix. $\odot$ denotes element-wise multiplication. 

The intuition of the term ``Long Short-Term Memory" is that the proposed model applies both long-term and short-term memory vectors to encode the sequential data, and uses gate mechanisms to control the information flow. The performance of LSTM is impressive since that it achieved state-of-the-art results in many NLP tasks as a backbone model although this model was proposed in 1997. 

\subsubsection{GRU}

Inspired by the gating mechanism,~\cite{cho2014learning} proposed Gated Recurrent Unit (GRU), which can be modeled by the equations:
\begin{equation}
z^{(t)} = \sigma\left(W^z x^{(t)} + U^{z} h^{(t-1)} + b_z\right)
\end{equation}
\begin{equation}
r^{(t)} = \sigma\left(W^r x^{(t)} + U^{r} h^{(t-1)} + b_r\right)
\end{equation}
\begin{equation}
\hat{h}^{(t)} = tanh\left(W^h x^{(t)} + U^{h} (r^{(t)} \odot h^{(t-1)}) + b_h\right)
\end{equation}
\begin{equation}
h^{(t)} = (1-z^{(t)})\odot h^{(t-1)} + z^{(t)} \odot \hat{h}^{(t)}
\end{equation}

Where $t$ represents time step $t$. $z$ and $r$ are gates, denoting update gate and reset gate respectively. $x$, $\hat{h}$ and $h$ are input, candidate activation vector and output respectively. $b$ is bias while $W$ and $U$ are weight matrixes. $\odot$ denotes element-wise multiplication. 

LSTM and GRU, as two types of gating units, are very similar to each other~\citep{chung2014empirical}. The most prominent common point between them is that from time step $t$ to time step $t+1$, an additive component is introduced to update the state whereas simple RNNs always replace the activation. Both LSTM and GRU keep certain old components and mix them with new contents. This property enables the units to remember the information of history steps farther back and, more importantly, avoid gradient vanishing problems when backpropagating the error. 

There also exist several differences between them. LSTM exposes its memory content under the control of the output gate, while the same content in GRU is in an uncontrolled manner. Additionally, different from LSTM, GRU does not independently gate the amount of new memory content being added. And if looking from experimental perspective, GRU has fewer parameters, which contributes to its faster convergence and better generalization ability. It has also been shown that GRU can achieve better performance in smaller datasets~\citep{chung2014empirical}. However,~\cite{gruber2020gru} showed that LSTM cells exhibited consistently better performance in a large-scale analysis of Neural Machine Translation. 

\subsubsection{Bidirectional Recurrent Neural Networks}

In sequence learning, not only the past information is essential to the model inference, the future information should also be considered to achieve a better inference ability. \cite{schuster1997bidirectional} proposed the bi-directional recurrent neural networks (BRNNs), which had two kinds of hidden layers: the first encoded information from past time steps while the second encoded information in a flipped direction. The model can be described using the equations: 
\begin{equation}
h^{(t)} = \sigma\left(W^{hx} x^{(t)} + W^{hh} h^{(t-1)} + b_h\right)
\end{equation}
\begin{equation}
z^{(t)} = \sigma\left(W^{zx} x^{(t)} + W^{zz} z^{(t+1)} + b_z\right)
\end{equation}
\begin{equation}
\hat{y}^{(t)} = softmax\left(W^{yh} h^{(t)} + W^{yz} z^{(t)} + b_y\right)
\end{equation}

Where $h$ and $z$ are the two hidden layers. Other variables are defined in the same way as in the case of LSTMs and GRUs.

\subsubsection{Vanilla Sequence-to-sequence Models (Encoder-decoder Models)}
\label{Vanilla Sequence-to-sequence Models (Encoder-decoder Models)}

~\cite{sutskever2014sequence} first proposed the sequence-to-sequence model to solve the machine translation tasks. The sequence-to-sequence model aimed to map an input sequence to an output sequence by first using an encoder to map the input sequence into an intermediate vector and a decoder further generated the output based on the intermediate vector and history generated by the decoder. The equations below illustrate the encoder-decoder model: 

\begin{equation}
Encoder: h_t = E(h_{t-1}, x_{t}) 
\end{equation}

\begin{equation}
Decoder: y_t = D(h_t, y_{t-1})
\end{equation}

Where $t$ is the time step, $h$ is the hidden vector and $y$ is the output vector. $E$ and $D$ are the sequential cells used by the encoder and decoder respectively. The last hidden state of the encoder is the intermediate vector, and this vector is usually used to initialize the first hidden state of the decoder. At encoding time, each hidden state is decided by the hidden state of the previous time step and the input at the current time step, while at decoding time, each hidden state is decided by the current hidden state and the output of the previous time step. 

This model is powerful because it is not restricted to fixed-length inputs and outputs. Instead, the length of the source sequence and target sequence can differ. Based on this model, many more advanced sequence-to-sequence models have been developed, which will be discussed in this and subsequent sections. \\

RNNs play an essential role in neural dialogue systems for their strong ability to encode sequential text information. RNNs and their variants are found in many dialogue systems. Task-oriented systems apply RNNs as encoders of dialogue context, dialogue state, knowledge base entries, and domain tags~\citep{moon2019opendialkg, chen2019semantically, wu2019self, wu2019transferable}. Open-domain systems apply RNNs as dialogue history encoders~\citep{sankar2019neural, du2019boosting, ji2020cross, chen2020bridging}, among which retrieval-based systems model dialogue history and candidate responses together~\citep{zhu2018retrieval, tang2019target, feldman2019multi, lu2019constructing}. In knowledge-grounded systems, RNNs are encoders of outside knowledge sources (e.g., background, persona, topic, etc.)~\citep{shuster2019dialogue, majumder2020interview, chen2020bridging, cho2020grounding}. 

Furthermore, as the decoder of sequence-to-sequence models in dialogue systems~\citep{huang2020meta, song2019generating, liu2019vocabulary, lin2019unified}, RNNs usually decode the hidden state of utterance sequences by greedy search or beam search~\citep{aubert1994large}. These decoding mechanisms cause problems like generic responses, which will be discussed in later sections. 

Some works~\citep{liu2019vocabulary, mehri2019pretraining, chen2019working, ma2020conversational} combined RNNs as a part of dialogue representation models to train dialogue embeddings and further improved the performance of dialogue-related tasks. These embedding models were trained on dialogue tasks and present more dialogue features. They consistently outperformed state-of-the-art contextual representation models (e.g., BERT, ELMo, and GPT) in some dialogue tasks when these contextual representation models were not fine-tuned for the specific tasks. 

\subsection{Hierarchical Recurrent Encoder-Decoder (HRED)}
\label{Hierarchical Recurrent Encoder-Decoder (HRED)}
Hierarchical Recurrent Encoder-Decoder (HRED) is a context-aware sequence-to-sequence model. It was first proposed by~\cite{sordoni2015hierarchical} to address the context-aware online query suggestion problem. It was designed to be aware of history queries and the proposed model can provide rare and high-quality results. 

With the popularity of the sequence-to-sequence model,~\cite{serban2016building} extended HRED to the dialogue domain and built an end-to-end context-aware dialogue system. HRED achieved noticeable improvements in dialogue and end-to-end question answering. This work attracted even more attention than the original paper for that dialogue systems are a perfect setting for the application of HRED. Traditional dialogue systems~\citep{ritter2011data} generated responses based on the single-turn messages, which sacrificed the information in the dialogue history. \cite{sordoni2015neural} combined dialogue history turns with a window size of 3 as the input of a sequence-to-sequence model for response generation, which is limited as well for that they encode the dialogue history only in token-level. The ``turn-by-turn" characteristic of dialogue indicated that the turn-level information also matters. The HRED learned both token-level and turn-level representation, thus exhibiting promising dialogue context awareness. 

\begin{figure}
\begin {center}
\includegraphics[width=1\textwidth]{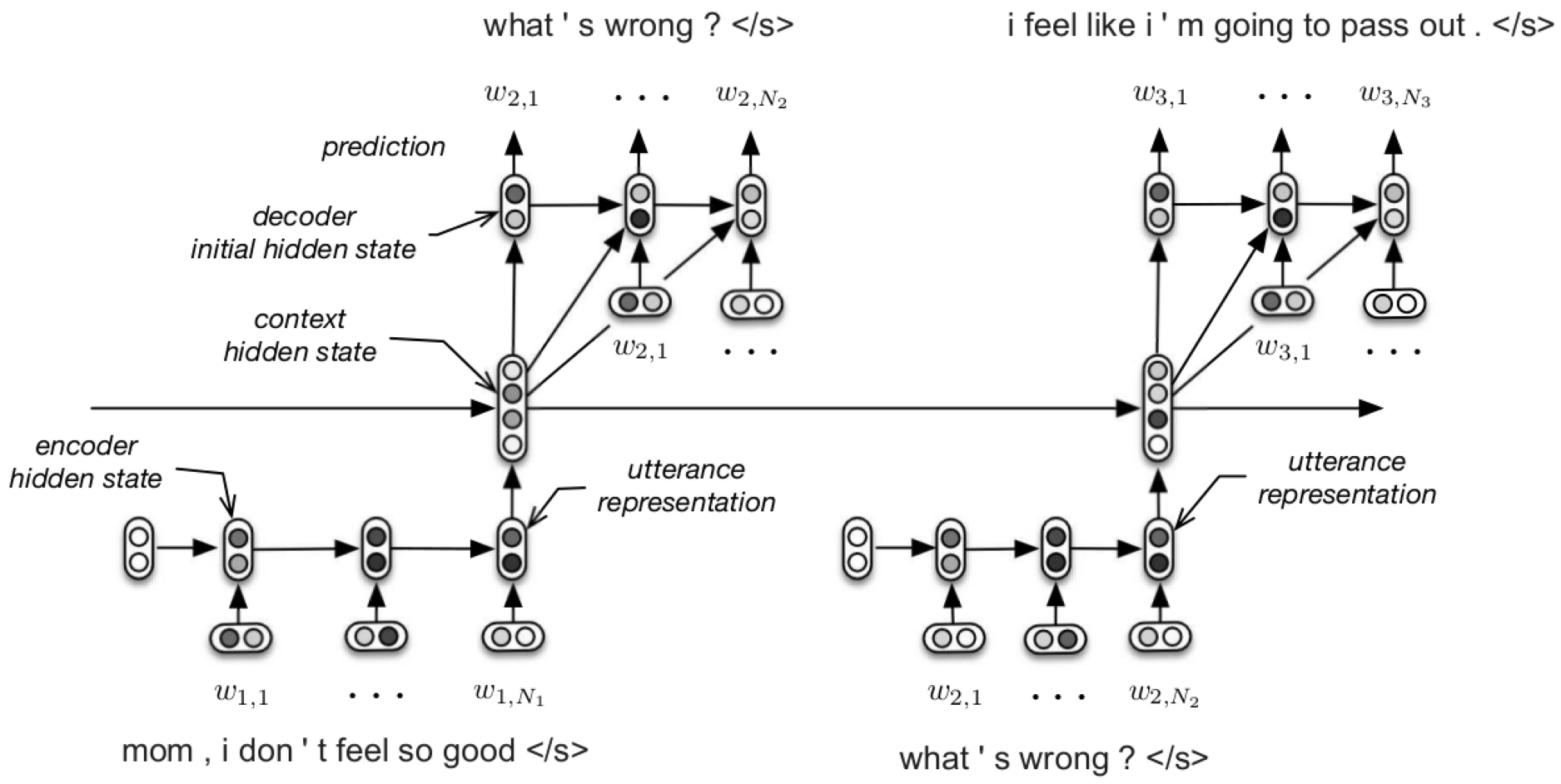}
\caption{The HRED model in a dialogue setting~\citep{serban2016building}}
\label{The HRED model in a dialogue setting}
\end {center}
\end{figure} 

Figure~\ref{The HRED model in a dialogue setting} represents the HRED in a dialogue setting. HRED models the token-level and turn-level sequences hierarchically with two levels of RNNs: a token-level RNN consisting of an encoder and a decoder, and a turn-level context RNN. The encoder RNN encodes the utterance of each turn token by token into a hidden state. This hidden state is then taken as the input of the context RNN at each turn-level time step. Thus the turn-level context RNN iteratively keeps track of the history utterances. The hidden state of context RNN at turn $t$ represents a summary of the utterances up to turn $t$ and is used to initialize the first hidden state of decoder RNN, which is similar to a standard decoder in sequence-to-sequence models~\citep{sutskever2014sequence}. All of the three RNNs described above apply GRU cells as the recurrent unit, and the parameters of encoder and decoder are shared for each utterance. 

~\cite{serban2017hierarchical} further proposed Latent Variable Hierarchical Recurrent Encoder-Decoder (VHRED) to model complex dependencies between sequences. Based on HRED, VHRED combined a latent variable into the decoder and turned the decoding process into a two-step generation process: sampling a latent variable at the first step and then generating the response conditionally. VHRED was trained with a variational lower bound on the log-likelihood and exhibited promising improvement in diversity, length, and quality of generated responses. \\

Many recent works in dialogue-related tasks apply HRED-based frameworks to capture hierarchical dialogue features. \cite{zhang2019recosa} argued that standard HRED processed all contexts in dialogue history indiscriminately. Inspired by the architecture of Transformer~\citep{vaswani2017attention}, they proposed ReCoSa, a self-attention-based hierarchical model. It first applied LSTM to encode token-level information into context hidden vectors and then calculated the self-attention for both the context vectors and masked response vectors. At the decoding stage, the encoder-decoder attention was calculated to facilitate the decoding. \cite{shen2019modeling} proposed a hierarchical model consisting of 3 hierarchies: the discourse-level which captures the global knowledge, the pair-level which captured the topic information in utterance pairs, and the utterance level which captured the content information. Such a multi-hierarchy structure contributed to its higher quality responses in terms of diversity, coherence, and fluency. \cite{chauhan2019ordinal} applied HRED and VGG-19 as a multimodal HRED (MHRED). The HRED encoded hierarchical dialogue context while VGG-19 extracted visual features for all images in the corresponding turn. With the addition of a position-aware attention mechanism, the model showed more diverse and accurate responses in a visually grounded setting. \cite{mehri2019pretraining} learned dialogue context representations via four sub-tasks, three of which (next-utterance generation, masked-utterance retrieval, and inconsistency identification) made uses of HRED as the context encoder, and good performance was achieved. \cite{cao2019observing} used HRED to encode the dialogue history between therapists and patients to categorize therapist and client MI behavioral codes and predict future codes. \cite{qiu2020structured} applied an LSTM-based VHRED to address the two-agent and multi-agent dialogue structure induction problem in an unsupervised fashion. On top of that, they applied a Conditional Random Field model in two-agent dialogues and a non-projective dependency tree in multi-agent dialogues, both of them achieving better performance in dialogue structure modeling. 

\subsection{Memory Networks}
\label{Memory Networks}
Memory is a crucial component when addressing problems regarding past experiences or outside knowledge sources. The hippocampus of human brains and the hard disk of computers are the components that humans and computers depend on for reading and writing memories. Traditional models rarely have a memory component, thus lacking the ability of knowledge reusing and reasoning. RNNs iteratively pass history information across time steps, which, to some extent, can be viewed as a memory model. However, even for LSTM, which is a powerful variant of RNN equipped with a long-term and short-term memory, the memory module is too small and facts are not explicitly discriminated, thus not being able to compress specific knowledge facts and reuse them in tasks. 

~\cite{weston2014memory} proposed memory networks, a model that is endowed with a memory component. As described in their work, a memory network has five modules: a memory module which stores the representations of memory facts; an `I' module which maps the input memory facts into embedded representations; a `G' module which decides the update of the memory module; an `O' module which generates the output conditioned on the input representation and memory representation; an `R' module which organizes the final response based on the output of `O' module. This model needs a strong supervision signal for each module and thus is not practical to train in an end-to-end fashion. 

~\cite{sukhbaatar2015end} extended their prior work to an end-to-end memory network, which was commonly accepted as a standard memory network being easy to train and apply. 

\begin{figure}
\begin {center}
\includegraphics[width=1\textwidth]{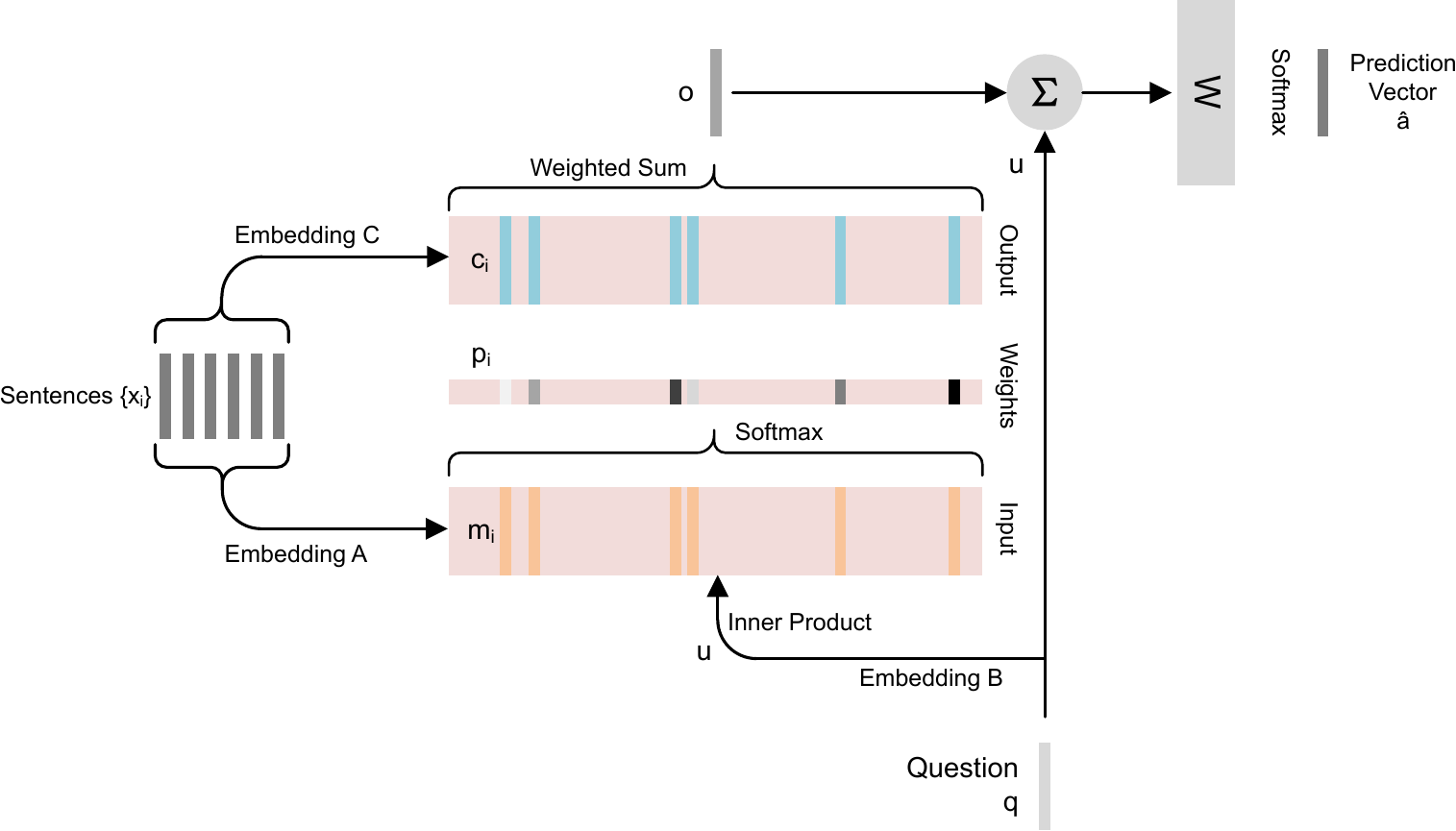}
\caption{The structure of end-to-end memory networks~\citep{sukhbaatar2015end}}
\label{The structure of end-to-end memory network}
\end {center}
\end{figure} 

Figure~\ref{The structure of end-to-end memory network} represents the proposed end-to-end memory networks. Its architecture consists of three stages: weight calculation, memory selection, and final prediction. 

\textbf{Weight calculation.} The model first converts the input memory set $\{x_i\}$ into memory representations $\{m_i\}$ using a representation model $A$. Then it maps the input query into its embedding space using another representation model $B$, obtaining an embedding vector $u$. The final weights are calculated as follows: 
\begin{equation}
p_i = Softmax(u^Tm_i)
\end{equation}
Where $p_i$ is the weight corresponding to each input memory $x_i$ conditioned on the query. 

\textbf{Memory selection.} Before generating the final prediction, a selected memory vector is generated by first encoding the input memory $x_i$ into an embedded vector $c_i$ using another representation model $C$, then calculating the weighted sum over the $\{c_i\}$ using the weights calculated in the previous stage: 
\begin{equation}
o = \sum_i p_ic_i
\end{equation}
Where o represents the selected memory vector. This vector cannot be found in memory representations. The soft memory selection facilitates differentiability in gradient computing, which makes the whole model end-to-end trainable. 

\textbf{Final prediction.} The final prediction is obtained by mapping the sum vector of the selected memory $o$ and the embedded query $u$ into a probability vector $\hat{a}$:

\begin{equation}
\hat{\alpha} = Softmax(W(o+u))
\end{equation}
\\
Many dialogue-related works incorporate memory networks into their framework, especially for tasks involving an external knowledge base like task-oriented dialogue systems, knowledge-grounded dialogue systems, and QA.

\paragraph*{Memory networks for task-oriented dialogue systems}~\cite{chen2019working} argued that state-of-the-art task-oriented dialogue systems tended to combine dialogue history and knowledge base entries in a single memory module, which influenced the response quality. They proposed a task-oriented system that consists of three memory modules: two long-term memory modules storing the dialogue history and the knowledge base respectively; a working memory module that memorizes two distributions and controls the final word prediction. \cite{he2020amalgamating} trained a task-oriented dialogue system with a ``Two-teacher-one-student" framework to improve the knowledge retrieval and response quality of their memory networks. They first trained two teacher networks using reinforcement learning with complementary goal-specific reward functions respectively. Then with a GAN framework, they trained two discriminators to teach the student memory network to generate responses similar to those of the teachers, transferring the expert knowledge from the two teachers to the student. The advantage is that this training framework needs only weak supervision and the student network can benefit from the complementary targets of teacher networks. \cite{kim2019efficient} solved the dialogue state tracking in task-oriented dialogue systems with a memory network that memorized the dialogue states. Different from other works, they did not update all dialogue states in the memory module from scratch. Instead, their model first predicted which states needed to be updated and then overwrote the target states. By selectively overwriting the memory module, they improved the efficiency of the dialogue state tracking task. \cite{dai2020learning} applied the MemN2N~\citep{sukhbaatar2015end} as task-oriented utterance encoder, memorizing the existing responses and dialogue history. Then they used model-agnostic meta-learning (MAML)~\citep{finn2017model} to train the framework to retrieve correct responses in a few-shot fashion. 

\paragraph*{Memory networks for open-domain dialogue systems}~\cite{tian2019learning} proposed a knowledge-grounded chit-chat system. A memory network was used to store query-response pairs and at the response generation stage, the generator produced the response conditioned on both the input query and memory pairs. It extracted key-value information from the query-response pairs in memory and combined them into token prediction. \cite{xu2019neural} proposed to use meta-words to generate responses in open-domain systems in a controllable way. Meta-words are phrases describing response attributes. Using a goal-tracking memory network, they memorized the meta-words and generated responses based on the user message while incorporating meta-words at the same time. \cite{gan2019multi} performed multi-step reasoning conditioned on a dialogue history memory module and a visual memory module. Two memory modules recurrently refined the representation to perform the next reasoning process. Experimental results illustrated the benefits of combining image and dialogue clues to improve the performance of visual dialogue systems. \cite{han2019episodic} trained a reinforcement learning agent to decide which memory vector can be replaced when the memory module is full to improve the accuracy and efficiency of the document-grounded question-answering task. They solved the scalability problem of memory networks by learning the query-specific value corresponding to each memory. \cite{gao2020explicit} solved the same problem in a conversational machine reading task. They proposed an Explicit Memory Tracker (EMT) to decide whether the provided information in memory is enough for final prediction. Furthermore, a coarse-to-fine strategy was applied for the agent to make clarification questions to request additional information and refine the reasoning. 

\subsection{Attention and Transformer}
\label{Attention and transformer}
As introduced in Section~\ref{RNNs and Vanilla Sequence-to-sequence Models}, traditional sequence-to-sequence models decode the token conditioning on the current hidden state and output vector of last time step, which is formulated as: 
\begin{equation}
\label{P(y_i|y_1,}
P(y_i|y_1, ..., y_{i-1}, x) = g(y_{i-1}, h_i)
\end{equation}
Where g is a sequential model which maps the input vectors into a probability vector.

However, such a decoding scheme is limited when the input sentence is long. RNNs are not able to encode all information into a fixed-length hidden vector. \cite{cho2014properties} proved via experiments that a sequence-to-sequence model performed worse when the input sequence got longer. Also, for the limited-expression ability of a fixed-length hidden vector, the performance of the decoding scheme in Equation (\ref{P(y_i|y_1,}) largely depends on the first few steps of decoding, and if the decoder fails to have a good start, the whole sequence would be negatively affected. 

\subsubsection{Attention}
\begin{figure}[t]
\begin {center}
\includegraphics[width=0.35\textwidth]{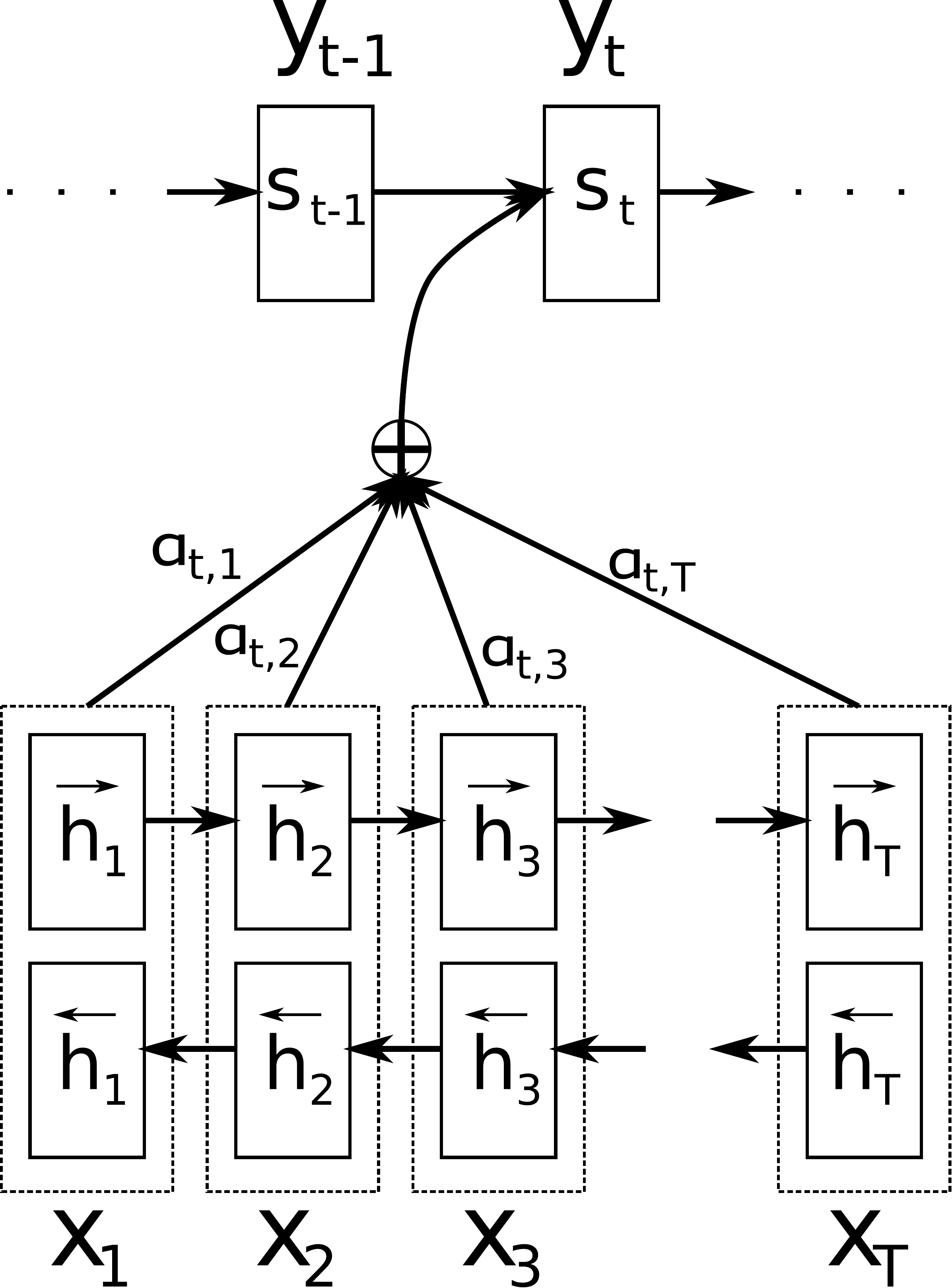}
\caption{The attention model~\citep{bahdanau2014neural}}
\label{The attention model}
\end {center}
\end{figure} 

\cite{bahdanau2014neural} proposed the attention mechanism in the machine translation task. They described the method as ``jointly align and translate", which illustrated the sequence-to-sequence translation model as an encoder-decoder model with attention. At the decoding stage, each decoding state would consider which parts of the encoded source sentence are correlated, instead of depending only on the immediate prior output token. The output probability distribution can be described as: 
\begin{equation}
P(y_i|y_1, ..., y_{i-1}, x) = g(y_{i-1}, s_i, c_i)
\end{equation}
Where $i$ denotes the $i^{th}$ time step; $y_i$ is the output token, $s_i$ is the decoder hidden state and $c_i$ is the weighted source sentence: 
\begin{equation}
s_i = f(s_{i-1}, y_{i-1}, c_i)
\end{equation}
\begin{equation}
c_i = \sum_{j = 1}^{T_x}\alpha_{ij}h_j
\end{equation}
Where $\alpha_{ij}$ is the normalized weight score: 
\begin{equation}
\alpha_{ij} = \frac{exp(e_{ij})}{\sum_{k = 1}^{T_x}exp(e_{ik})}
\end{equation}
$e_{ij}$ is the similarity score between $s_{i-1}$ and $j^{th}$ encoder hidden state $h_j$, where the score is predicted by the similarity model $a$: 
\begin{equation}
e_{ij} = a(s_{i-1}, h_j)
\end{equation}
Figure~\ref{The attention model} illustrates the attention model, where t and T denote time steps of decoder and encoder respectively.

Memory networks are similar to attention networks in the way they operate, except for the choice of the similarity model. In memory networks, the encoded memory can be viewed as the encoded source sentence in attention. However, the memory model proposed by~\cite{sukhbaatar2015end} chose cosine distance as the similarity model while the attention proposed by~\cite{bahdanau2014neural} used a feed-forward network which is trainable together with the whole sequence-to-sequence model. 

\subsubsection{Transformer}

Before transformers, most works combined attention with recurrent units, except for few works such as \cite{parikh2016decomposable} and \cite{gehring2017convolutional}. Recurrent models condition each hidden state on the previous hidden state and the current input and are flexible in sequence length. However, due to their sequential nature, recurrent models cannot be trained in parallel, which severely undermines their potential. \cite{vaswani2017attention} proposed Transformer, which entirely utilized attention mechanisms without any recurrent units and deployed more parallelization to speed up training. It applied self-attention and encoder-decoder attention to achieve local and global dependencies respectively. 

\begin{figure}[h]
\begin {center}
\includegraphics[width=0.7\textwidth]{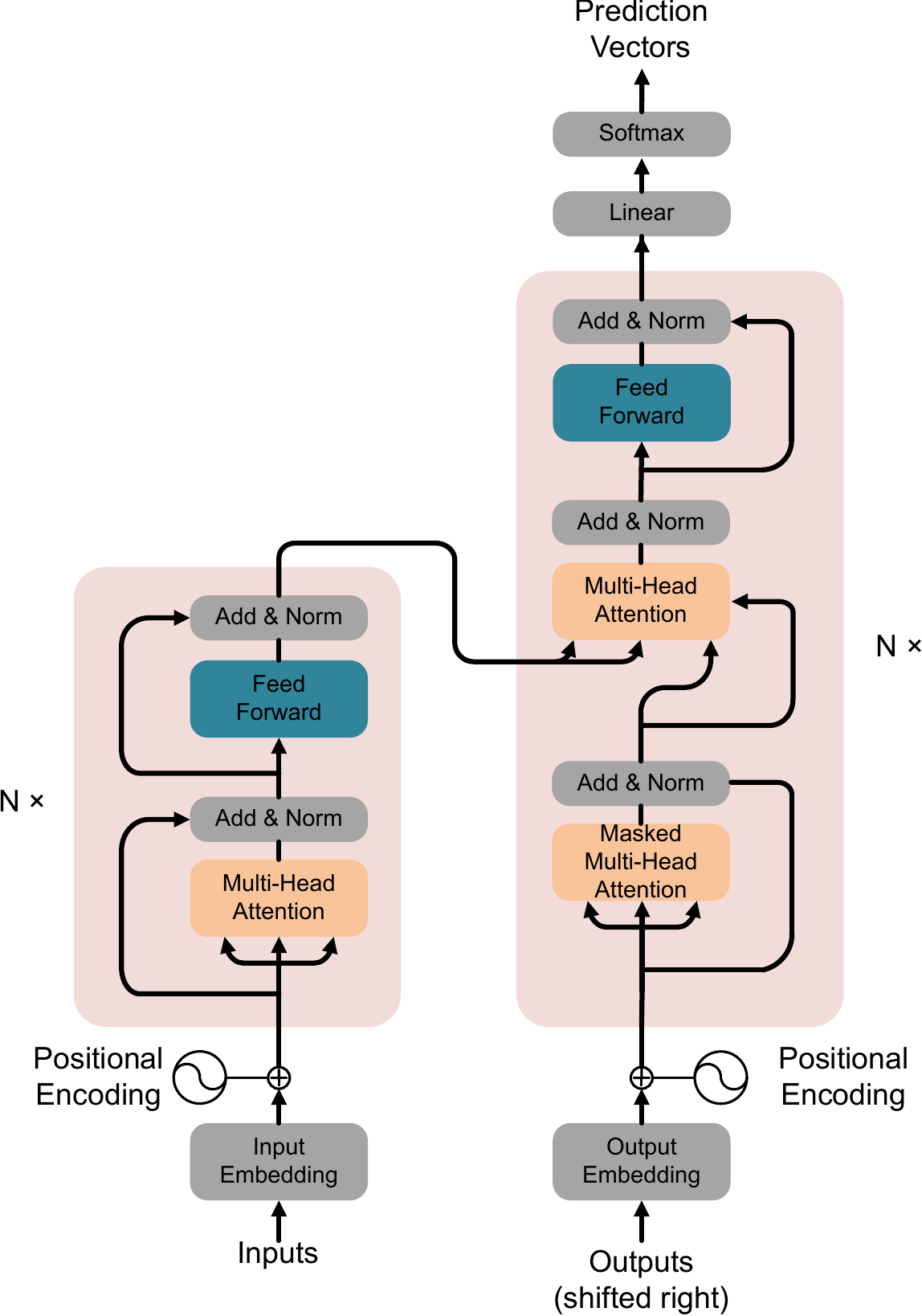}
\caption{The transformer model~\citep{vaswani2017attention}}
\label{The transformer model}
\end {center}
\end{figure}

Figure~\ref{The transformer model} represents the transformer. The following details its key mechanisms. 

\paragraph*{Encoder-decoder} The Transformer consists of an encoder and a decoder. The encoder maps the input sequence $(x_1, \ldots,x_n)$ into continuous hidden states $(z_1, \ldots,z_n)$. The decoder further generates the output sequence $(y_1, \ldots,y_n)$ based on the hidden states of the encoder. The probability model of the Transformer is in the same form as that of the vanilla sequence-to-sequence model introduced in Section~\ref{Vanilla Sequence-to-sequence Models (Encoder-decoder Models)}. \cite{vaswani2017attention} stacked 6 identical encoder layers and 6 identical decoder layers. An encoder layer consists of a multi-head attention component and a simple feed-forward network, both of which apply residual structure. The structure of a decoder layer is almost the same as that of an encoder layer, except for an additional encoder-decoder attention layer, which computes the attention between decoder hidden states of the current time step and the encoder output vectors. The input of the decoder is partially masked to make sure that each prediction is based on the previous tokens, avoiding predicting with the presence of future information. Both inputs of encoder and decoder use a positional encoding mechanism. 

\paragraph*{Self-attention} For an input sentence $x = (x_1, \ldots, x_n)$, each token $x_i$ corresponds to three vectors: query, key, and value. The self-attention computes the attention weight for every token $x_i$ against all other tokens in $x$ by multiplying the query of $x_i$ with the keys of all the remaining tokens one-by-one. For parallel computing, the query, key ,and value vectors of all tokens are combined into three matrices: Query (Q), Key (K) ,and Value (V). The self-attention of an input sentence $x$ is computed by the following formula: 
\begin{equation}
Attention(Q, K, V) = softmax(\frac{QK^T}{\sqrt{d_k}})V
\end{equation}
Where $d_k$ is the dimension of queries or keys. 

\paragraph*{Multi-head attention} To jointly consider the information from different subspaces of embedding, query, key, and value vectors are mapped into $h$ vectors of identical shapes by using different linear transformations, where $h$ denotes the number of heads. Attention is computed on each of these vectors in parallel, and the results are concatenated and further projected. The multi-head attention can be described as: 

\begin{equation}
MultiHead(Q, K, V) = Concat(head_1, ..., head_h)W^O
\end{equation}
Where $head_i = Attention (QW_i^Q, KW_i^K, VW_i^V)$ and $W$ denotes the linear transformations. 

\paragraph*{Positional encoding} The proposed transformer architecture has no recurrent units, which means that the order information of sequence is dismissed. The positional encoding is added with input embeddings to provide positional information. The paper chooses cosine functions for positional encoding: 
\begin{equation}
PE_{(pos, 2i)} = sin(pos/10000^{2i/d_{model}})
\end{equation}
\begin{equation}
PE_{(pos, 2i+1)} = cos(pos/10000^{2i/d_{model}})
\end{equation}
Where $pos$ denotes the position of the target token and $i$ denotes the dimension, which means that each dimension of the positional matrix uses a different wavelength for encoding. 

\paragraph*{Transformer-based pretrain models and Transformer variants} Recently, many transformer-based pretrain models have been developed. Unlike Embeddings from Language Model (ELMo) proposed by~\cite{peters2018deep}, which is an LSTM-based contextual embedding model, transformer-based pretrain models are more powerful. Two most popular models are GPT-2 \footnote{https://openai.com/blog/better-language-models/} and BERT~\citep{devlin2018bert}. GPT-2 and BERT both consist of 12 transformer blocks and BERT is further improved by making the training bi-directional. They are powerful due to their capability of adapting to new tasks after pretraining. This property helped achieve significant improvements in many NLP tasks. There also evolve many Transformer variants \citep{zaheer2020big, dai2019transformer, guo2019star}, which are designed to reduce the model parameters/computational complexity, or improve performance of the original Transformer in diverse scenarios. \cite{lin2021survey} and \cite{tay2020efficient} systematically summarize the state-of-the-art Transformer variants for academics that are interested.\\

 
\paragraph*{Attention for dialogue systems} Attention is a mechanism to catch the importance of different parts in the target sequence. \cite{zhu2018retrieval} applied a two-level attention to generate words. Given the user message and candidate responses selected by a retrieval system, the generator first computes word-level attention weights, then uses sentence-level attention to rescale the weights. This two-level attention helps the generator catch different importance given the encoded context. \cite{liu2019vocabulary} used an attention-based recurrent architecture to generate responses. They designed a multi-level encoder-decoder of which the multi-level encoder tries to map raw words, low-level clusters, and high-level clusters into hierarchical embedded representations while the multi-level decoder leveraged the hierarchical representations using attention and then generated responses. At each decoding stage, the model calculated two attention weights for the output of the higher-level decoder and the hidden state of the current level's encoder. \cite{chen2019semantically} computed multi-head self-attention for the outputs of a dialogue act predictor. Unlike the transformer, which concatenates the outputs of different heads, they passed the outputs directly to the next multi-head layer. The stacked multi-head layers then generated the responses with dialogue acts as the input. 

\paragraph*{Transformers for dialogue systems} Transformers are powerful sequence-to-sequence models and meanwhile, their encoders also serve as good dialogue representation models. \cite{henderson2019training} built a transformer-based response retrieval model for task-oriented dialogue systems. A two-channel transformer encoder was designed for encoding user messages and responses, both of which were initially presented as unigrams and bigrams. A simple cosine distance was then applied to calculate the semantic similarity between the user message and the candidate response. \cite{li2019incremental} built multiple incremental transformer encoders to encode multi-turn conversations and their related document knowledge. The encoded utterance and related document of the previous turn were treated as a part of the input of the next turn's transformer encoder. The pretrained model was adaptable to multiple domains with only a small amount of data from the target domain. \cite{bao2019plato} used stacked transformers for dialogue generation pretraining. Besides the response generation task, they also pretrained the model together with a latent act prediction task. A latent variable was applied to solve the ``one-to-many" problem in response generation. The multi-task training scheme improved the performance of the proposed transformer pretraining model.  

\paragraph*{Transformer-based pretrain models for dialogue systems} Large transformer-based pretrain models are adaptable to many tasks and are thus popular in recent works. \cite{golovanov2019large} used GPT as a sequence-to-sequence model to directly generate utterances and compared the performances under single- and multi-input settings. \cite{majumder2020interview} first used a probability model to retrieve related news corpus and then combined the news corpus and dialogue context as input of a GPT-2 generator for response generation. They proposed that by using discourse pattern recognition and interrogative type prediction as two subtasks for multi-task learning, the dialogue modeling could be further improved. \cite{wu2019proactive} used BERT as an encoder of context and candidate responses in their goal-based response retrieval system while~\cite{zhongtowards2020} built Co-BERT, a BERT-based response selection model, to retrieve empathetic responses given persona-based training corpus. \cite{zhao2020knowledge} built a knowledge-grounded dialogue system in a synthesized fashion. They used both BERT and GPT-2 to perform knowledge selection and response generation jointly, where BERT was for knowledge selection and GPT-2 generated responses based on dialogue context and the selected knowledge. 

\subsection{Pointer Net and CopyNet}
\label{Pointer Net and CopyNet}
\subsubsection{Pointer Net}
\begin{figure}
\begin {center}
     \begin{subfigure}[b]{0.45\textwidth}
         \centering
         \includegraphics[width=\textwidth]{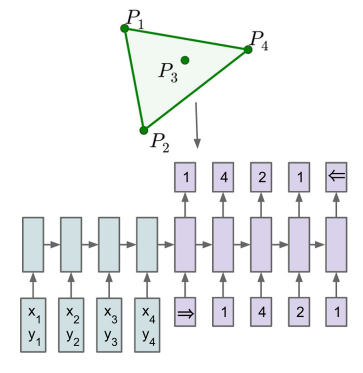}
         \caption{Sequence-to-sequence}
     \end{subfigure}
\hfill
     \begin{subfigure}[b]{0.45\textwidth}
         \centering
         \includegraphics[width=\textwidth]{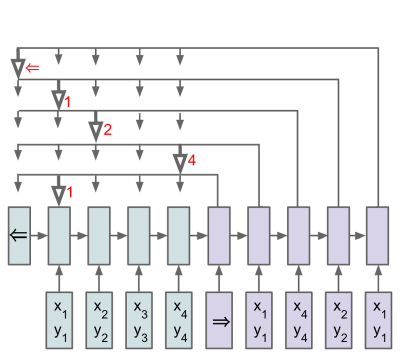}
         \caption{Pointer Net}
     \end{subfigure}
\caption{\textbf{(a)} \textit{Sequence-to-sequence} - The RNN (blue) processes the input sequence to produce a code vector, which is then used by the probability chain rule and another RNN to generate the output sequence (purple). The dimensionality of the problem determines the output dimensionality, which remains constant through training and inference. \textbf{(b)} \textit{Pointer Net} - The input sequence is converted to a code (blue) by an encoding RNN, which is fed to the generating network (purple). The generating network generates a vector at each step that modulates a content-based attention process across inputs. The attention mechanism produces a softmax distribution with a dictionary size equal to the input length.~\citep{oriol2015pointer}}
\label{Pointer Net}
\end {center}
\end{figure}

In some NLP tasks like dialogue systems and question-answering, the agents sometimes need to directly quote from the user message. Pointer Net~\citep{oriol2015pointer} (Figure~\ref{Pointer Net}) solved the problem of directly copying tokens from the input sentence. 

Traditional sequence-to-sequence models~\citep{sutskever2014sequence, graves2014neural} with an encoder-decoder structure map a source sentence to a target sentence. Generally, these models first map source sentence into hidden state vectors with an encoder, and then predict the output sequence based on the hidden states. The sequence prediction is accomplished step-by-step, each step predicting one token using greedy search or beam search. The overall sequence-to-sequence model can be described by the following probability model: 
\begin{equation}
P(C^P|P;\theta) = \prod_{i=1}^{m(P)}p(C_i|C_1, ..., C_{i-1}, P; \theta)
\end{equation}
Where $(P, C_p)$ constitutes a training pair, $P$ = $\{P_1, ..., P_n\}$ denotes the input sequence and $C_p$ = $\{C_1, ..., C_{m(p)}\}$ denotes the ground target sequence. $\theta$ is a decoder model. 

The sequence-to-sequence models have the vanilla backbones and attention-based backbones. Vanilla models predict the target sequence based only on the last hidden state of the encoder and pass it across different decoder time steps. Such a mechanism restricts the information received by the decoder at each decoding stage. Attention-based models consider all hidden states of the encoder at each decoding step and calculate their importance when utilizing them. To compare the mechanism of Pointer Net and Attention, we present the equations explained in Section~\ref{RNNs and Vanilla Sequence-to-sequence Models} here again. The decoder predicts the token conditioned partially on the weighted sum of encoder hidden states $d_i$: 
\begin{equation}
d_i = \sum_{j=1}^{T_x}\alpha_{ij}h_j
\end{equation}
Where $\alpha_{ij}$ is the normalized weight score:
\begin{equation}
\alpha_{ij} = \frac{exp(e_{ij})}{\sum_{k=1}^{T_x}exp(e_{ik})}
\end{equation}
$e_{ij}$ is the similarity score between $s_{i-1}$ and $jth$ encoder hidden state $h_j$, where the score is predicted by the similarity model $a$: 
\begin{equation}
e_{ij} = a(s_{i-1}, h_j)
\end{equation}

At each decoding step, both vanilla and attention-based sequence-to-sequence models predict a distribution over a fixed dictionary $X = \{x_1, ..., x_n\}$, where $x_i$ denotes the tokens and $n$ denotes the total count of different tokens in the training corpus. However, when copying words from the input sentence, we do not need such a large dictionary. Instead, $n$ equals to the number of tokens in the input sequence (including repeated ones) and is not fixed since it changes according to the length of the input sequence. Pointer Net made a simple change to the attention-based sequence-to-sequence models: instead of predicting the token distribution based on the weighted sum of encoder hidden states $d_i$, it directly used the normalized weights $\alpha_i$ as predicted distribution: 
\begin{equation}
P(C_i|C_1, ..., C_{i-1}, P) = \alpha_i
\end{equation}
Where $\alpha_i$ is a set of probability numbers $\{\alpha_i^1, ..., \alpha_i^j\}$ which represents the probability distribution over the tokens of the input sequence. Obviously, the \textit{token prediction} problem is now transformed into \textit{position prediction} problem, where the model only needs to predict a position in the input sequence. This mechanism is like a pointer that points to its target, hence the name ``Pointer Net". 

\subsubsection{CopyNet}

\begin{figure}
\begin {center}
\includegraphics[width=1.0\textwidth]{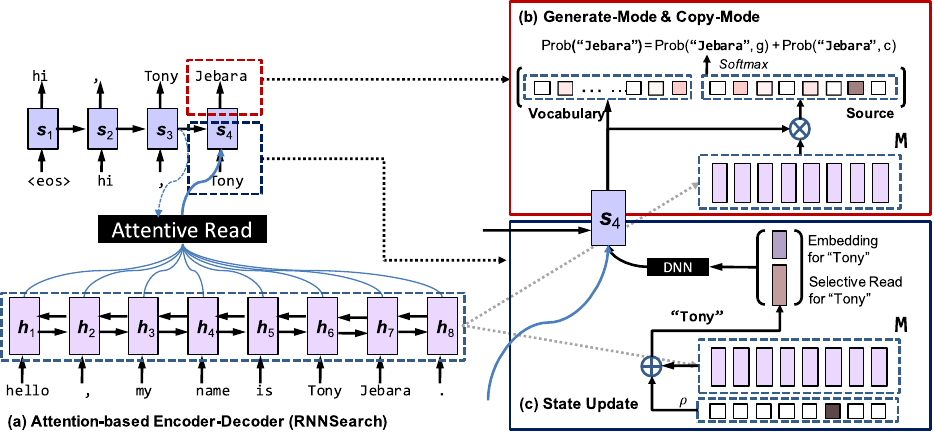}
\caption{The overall architecture of CopyNet~\citep{gu2016incorporating}}
\label{The overall architecture of CopyNet}
\end {center}
\end{figure} 

In real-world applications, simply copying from the source message is not enough. Instead, in tasks like dialogue systems and QA, agents also require the ability to generate words that are not in the source sentence. CopyNet~\citep{gu2016incorporating} (Figure~\ref{The overall architecture of CopyNet}) was proposed to incorporate the copy mechanism into traditional sequence-to-sequence models. The model decides at each decoding stage whether to copy from the source or generate a new token not in the source. 

The encoder of CopyNet is the same as that of a traditional sequence-to-sequence model, whereas the decoder has some differences compared with a traditional attention-based decoder. When predicting the token at time step $t$, it combines the probabilistic models of generate-mode and copy-mode: 
\begin{equation}
P(y_t|s_t, y_{t-1}, c_t, M) = P_g(y_t|s_t, y_{t-1}, c_t, M) + P_c(y_t|s_t, y_{t-1}, c_t, M)
\end{equation}
Where $t$ is the time step. $s_t$ is the decoder hidden state and $y_t$ is the predicted token. $c_t$ and $M$ represent weighted sum of encoder hidden states and encoder hidden states respectively. $g$ and $c$ are generate-mode and copy-mode respectively. 

Besides, though it still uses $y_{t-1}$ and weighted attention vector $c_t$ to update the decoder hidden state, $y_{t-1}$ is uniquely encoded with both its embedding and its location-specific hidden state; also, CopyNet combines attentive read and selective read to capture information from the encoder hidden states, where the selective read is the same method used in Pointer Net. Different from the Neural Turing Machines~\citep{graves2014neural, kurach2015neural}, the CopyNet has a location-based mechanism that enables the model to be aware of some specific details in training data in a more subtle way. \\

Copy mechanism is suitable for dialogues involving terminologies or external knowledge sources, and it is popular in knowledge-grounded or task-oriented dialogue systems. 

\paragraph*{Copy mechanism for knowledge-grounded dialogue systems} For knowledge-grounded systems, external documents or dialogues are sources to copy from. \cite{lin2020generating} combined a recurrent knowledge interactive decoder with a knowledge-aware pointer network to achieve both knowledge-grounded generation and knowledge copy. In the proposed model, they first calculated the attention distribution over external knowledge, then used two pointers referring to dialogue context and knowledge source respectively to copy out-of-vocabulary (OOV) words. \cite{wu2020diverse} applied a multi-class classifier to flexibly fuse three distributions: generated words, generated knowledge entities, and copied query words. They used Context-Knowledge Fusion and Flexible Mode Fusion to perform the knowledge retrieval, response generation, and copying jointly, making the generated responses precise, coherent, and knowledge-infused. \cite{ji2020cross} proposed a Cross Copy Network to copy from internal utterance (dialogue history) and external utterance (similar cases) respectively. They first used pretrained language models for similar case retrieval, then combined the probability distribution of two pointers to make a prediction. They only experimented with court debate and customer service content generation tasks, where similar cases were easy to obtain. 

\paragraph*{Copy mechanism for task-oriented dialogue systems} Many dialogue state tracking tasks generate slots and slot values using a copy component~\citep{wu2019transferable, ouyang2020dialogue, gangadharaiah2020recursive, chen2020parallel, zhang2020probabilistic, li2020slot}. Among them~\cite{wu2019transferable},~\cite{ouyang2020dialogue} and~\cite{chen2020parallel} solved the problem of multi-domain dialogue state tracking. \cite{wu2019transferable} proposed TRAnsferable Dialogue statE generator (TRADE), a copy-based dialogue state generator. The generator decoded the slot value multiple times for each possible (domain, slot) pair, then a slot gate was applied to decide which pair belonged to the dialogue. The output distribution was a copy of the slot values belonging to the selected (domain, slot) pairs from vocabulary and dialogue history. \cite{chen2020parallel} used a different copy strategy from TRADE. Instead of using the whole dialogue history as the copy source, they copied state values from user utterances and system messages respectively, which took the slot-level context as input. \cite{ouyang2020dialogue} proposed slot connection mechanism to efficiently utilize existing states from other domains. Attention weights were calculated to measure the connection between the target slot and related slot-value tuples in other domains. Three distributions over token generation, dialogue context copying, and past state copying were finally gated and fused to predict the next token. \cite{gangadharaiah2020recursive} combined a pointer network with a template-based tree decoder to fill the templates recursively and hierarchically. Copy mechanisms also alleviated the problem of expensive data annotation in end-to-end task-oriented dialogue systems. Copy-augmented dialogue generation models were proven to perform significantly better than strong baselines with limited domain-specific or multi-domain data~\citep{zhang2020probabilistic, li2020slot, gao2020paraphrase}. 

\paragraph*{Copy mechanism for dialogue-related tasks} Pointer networks and CopyNet are also used to solve other dialogue-related tasks. \cite{yu2020online} applied a pointer net for online conversation disentanglement. The pointer module pointed to the ancestor message to which the current message replies and a classifier predicted whether two messages belonged to the same thread. In dialogue parsing tasks, the pointer net is used as the backbone parsing model to construct discourse trees~\citep{aghajanyan2020conversational, lin2019unified}. \cite{tay2019simple} used a pointer-generator framework to perform machine reading comprehension over a long span, where the copy mechanism reduced the demand of including target answers in context.

\subsection{Deep Reinforcement Learning Models and Generative Adversarial Networks}
\label{Deep Reinforcement Learning Models and Generative Adversarial Network}
In recent years, two exciting approaches exhibit the potential of artificial intelligence. The first one is deep reinforcement learning, which outperforms humans in many complex problems such as large-scale games, conversations, and car-driving. Another technique is GAN, showing amazing capability in generation tasks. The data samples generated by GAN models like articles, paintings, and even videos, are sometimes indistinguishable from human creations. 

AlphaGo~\citep{silver2016mastering} stimulated the research interests again in reinforcement learning in recent years~\citep{graves2016hybrid, mnih2016asynchronous, wang2016dueling, tamar2016value, jaderberg2016reinforcement, mirowski2016learning}. Reinforcement learning is a branch of machine learning aiming to train agents to perform appropriate actions while interacting with a certain environment. It is one of the three fundamental machine learning branches, with supervised learning and unsupervised learning being the other two. It can also be seen as an intermediate between supervised learning and unsupervised learning because it only needs weak signals for training \citep{wang2016dueling}. 

\begin{figure}
\begin {center}
\includegraphics[width=0.60\textwidth]{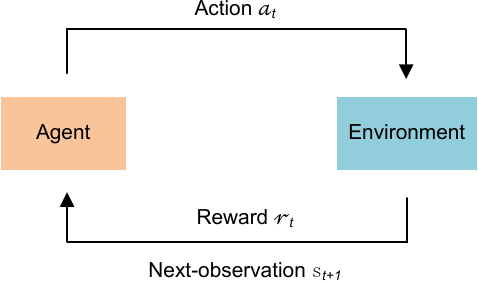}
\caption{The reinforcement learning framework}
\label{The reinforcement learning framework}
\end {center}
\end{figure} 

Figure~\ref{The reinforcement learning framework} illustrates the reinforcement learning framework, consisting of an agent and an environment. The framework is a Markov Decision Process (MDP)~\citep{puterman2014markov}, which can be described by a five-tuple M = $\langle S, A, P, R, \gamma \rangle $. $S$ denotes an infinite set of environment states; $A$ denotes a set of actions that agent chooses from conditioned on a given environment state $s$; $P$ is the transition probability matrix in MDP, denoting the probability of an environment state transfer after agent takes an action; $R$ is an average reward the agent receives from the environment after taking an action under state $s$; $\gamma$ is a discount factor. The flow of this framework is a loop of the following two steps: the agent first makes an observation on the current environment state $s_t$ and chooses an action based on its policy; then according to the transition probability matrix $P$, the environment's state transfers to $s_{t+1}$, and simultaneously provides a reward $r_t$. 

Reinforcement learning is applicable to solve many challenges in dialogue systems because of the agent-environment nature of a dialogue system. A two-party dialogue system consists of an agent, which is an intelligent chatbot, and an environment, which is usually a user or a user simulator. Here we mainly discuss deep reinforcement learning. 

Deep reinforcement learning means applying deep neural networks to model the value function or policy of the reinforcement learning framework. ``Deep model" is in contrast to the ``shallow model". The shallow model normally refers to traditional machine learning models like Decision Trees or KNN. Feature engineering, which is usually based on shallow models, is time and labor consuming, and also over-specified and incomplete. Different from that, deep neural models are easy to design and have a strong fitting capability, which contributes to many breakthroughs in recent research. Deep representation learning gets rid of human labor and exploits hierarchical features in data automatically, which strengthens the semantic expressiveness and domain correlations significantly. 

We discuss two typical reinforcement models: Deep Q-Networks~\citep{mnih2015human} and REINFORCE~\citep{williams1992simple, sutton1999policy}. They belong to \textit{Q-learning} and \textit{policy gradient} respectively, which are two families of reinforcement learning. 

\subsubsection{Deep Q-Networks}

A Deep Q-Network is a value-based RL model. It determines the best policy according to the Q-function: 
\begin{equation}
\pi^{*}(s) = arg\max_{a}Q^{*}(s, a)
\end{equation}
Where $Q^{*}(s, a)$ is an optimal Q-function and $\pi^{*}(s)$ is the corresponding optimal policy. In Deep Q-Networks, the Q function is modeled using a deep neural network, such as CNNs, RNNs, etc. 

As in \cite{gao2018neural}, the parameters of the Q model are updated using the rule: 
\begin{equation}
\theta \leftarrow \theta + \alpha \underbrace{\left (r_t+\gamma\max_{a_{t+1}}Q(s_{t+1}, a_{t+1}; \theta) - Q(s_t, a_t; \theta) \right )}_\text{temporal\ difference} \bigtriangledown_{\theta}Q(s_t, a_t; \theta)
\end{equation}
Where the $(s_t, a_t, r_t, s_{t+1})$ is an observed trajectory. $\alpha$ denotes step-size and the parameter update is calculated using temporal difference~\citep{sutton1988learning}. However, this update mechanism suffers from unstableness and demands a large number of training samples. There are two typical tricks for a more efficient and stable parameter update.

The first method is experience replay~\citep{lin1992self, mnih2015human}. Instead of using one training sample at a time to update the parameters, it uses a buffer to store training samples, and iteratively retrieves training samples from the buffer pool to perform parameter updates. It avoids encountering training samples that change too fast in distribution during training time, which increases the learning stability; further, it uses each training sample multiple times, which improves the efficiency. 

The second is two-network implementation~\citep{mnih2015human}. This method uses two networks in Q-function optimization, one being the Q-network, another being a target network. The target network is used to calculate the temporal difference, and its parameters $\theta_{target}$ are frozen while training, aligning with $\theta$ periodically. The parameters are then updated with the following rule: 
\begin{equation}
\theta \leftarrow \theta + \alpha \underbrace{\left (r_t+\gamma\max_{a_{t+1}}Q(s_{t+1}, a_{t+1}; \theta_{target}) - Q(s_t, a_t; \theta) \right )}_\text{temporal\ difference\ with\ a\ target\ network} \bigtriangledown_{\theta}Q(s_t, a_t; \theta)
\end{equation}
Since $\theta_{target}$ does not change in a period of time, the target network calculates the temporal difference in a stable manner, which facilitates the convergence of training. 

\subsubsection{REINFORCE}

REINFORCE is a policy-based RL algorithm that has no value network. It optimizes the policy directly. The policy is parameterized by a policy network, whose output is a distribution over continuous or discrete actions. A long-term reward is computed for evaluation of the policy network by collecting trajectory samples of length $H$: 
\begin{equation}
J(\theta) = E \left[ \sum_{t=1}^{H}\gamma^{t-1}r_t|a_t \sim \pi(s_t;\theta) \right]
\end{equation}
$J(\theta)$ denotes a long-term reward and the goal is to optimize the policy network in order to maximize $J(\theta)$. Here stochastic gradient ascent\footnote{Stochastic gradient ascent simply uses the negated objective function of stochastic gradient descent.} is used as an optimizer: 
\begin{equation}
\theta \leftarrow \theta + \alpha \bigtriangledown_{\theta}J(\theta)
\end{equation}
Where $\bigtriangledown_{\theta}J(\theta)$ is computed by: 
\begin{equation}
\bigtriangledown_{\theta}J(\theta) = \sum_{t=1}^{H-1}\gamma^{t-1}\left(\bigtriangledown_{\theta}log\pi(a_t|s_t; \theta) \sum_{h=t}^{H}\gamma^{h-t}r_h \right)
\label{bigtriangledown}
\end{equation} 

Both models have their advantages: Deep Q-Networks are more sample efficient while REINFORCE is more stable~\citep{li2017deep}. REINFORCE is more popular in recent works. Modern research involves larger action spaces, which means that value-based RL models like Deep Q-Networks are not suitable for problem-solving. Value-based methods ``select an action to maximize the value", which means that their action sets should be discrete and moderate in scale; while policy gradient methods such as REINFORCE are different, they predict the action via policy networks directly, which sets no restriction on the action space. As a result, policy gradient methods are more suitable for tasks involving a larger action space. 

Considering the respective benefits brought by the Q-learning and policy gradient, some work has been done combining the value- and policy-based methods. Actor-critic algorithm~\citep{konda2000actor, sutton1999policy} was proposed to alleviate the severe variance problem when calculating the gradient in policy gradient methods. It estimates a value function for term $\sum_{h=t}^{H}\gamma^{h-t}r_h$ in Equation (\ref{bigtriangledown}) and incorporates it in policy optimization. Equation (\ref{bigtriangledown}) is then transformed into the formula below:
\begin{equation}
\bigtriangledown_{\theta}J(\theta) = \sum_{t=1}^{H-1}\gamma^{t-1}\left(\bigtriangledown_{\theta}log\pi(a_t|s_t; \theta) \hat{Q}(s_t, a_t, h) \right)
\end{equation} 
Where $\hat{Q}(s_t, a_t, h)$ stands for the value function estimated.

\subsubsection{GANs}

It is easy to link the actor-critic model with another framework - GANs~\citep{goodfellow2014generative, zhang2018generating, feng2020posterior} because of their similar inner structure and logic~\citep{pfau2016connecting}. Actually, there are quite a few recent works in dialogue systems that train GANs with reinforcement learning framework~\citep{zhu2018retrieval, wu2019self, he2020amalgamating, zhu2020counterfactual, qin2020dynamic}. 

\begin{figure}
\begin {center}
\includegraphics[width=0.8\textwidth]{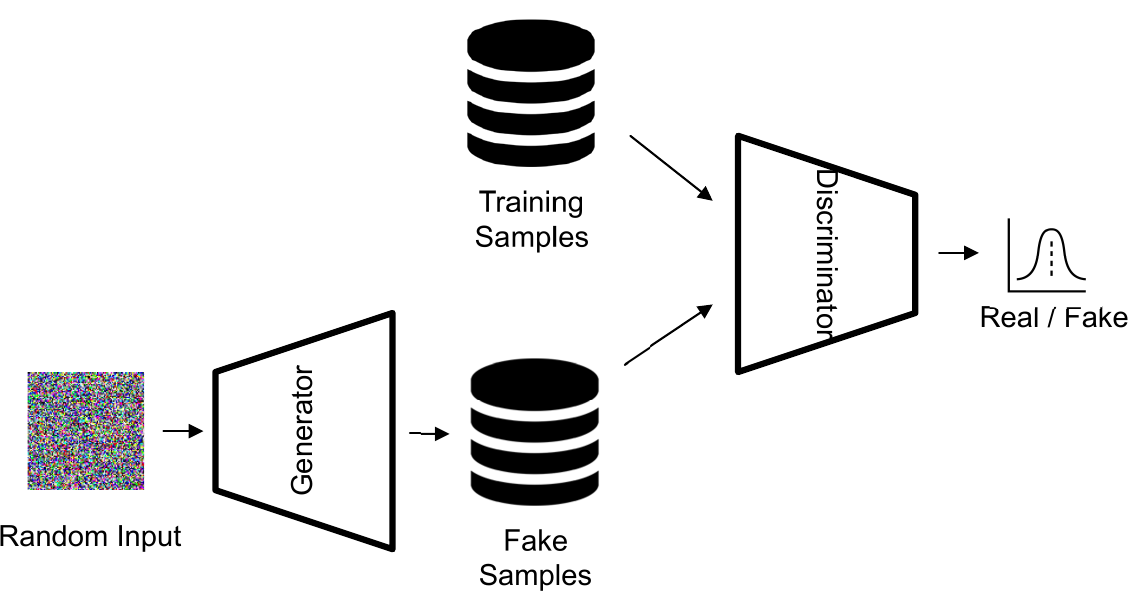}
\caption{The GAN framework}
\label{The GAN model}
\end {center}
\end{figure} 

Figure~\ref{The GAN model} represents the GAN consisting of a generator and a discriminator where the training process can be viewed as a competition between them: the generator tries to generate data distributions to fool the discriminator while the discriminator attempts to distinguish between real data (real) and generated data (fake). During training, the generator takes noise as input and generates data distribution while the discriminator takes real and fake data as input and the binary annotation as the label. The whole GAN model is trained end-to-end as a connection of generator and discriminator to minimize the following cross-entropy losses: 
\begin{equation}
L_1(D, G) = -E_{\omega\sim P_{data}}[logD(\omega)]-E_{z\sim N(0,I)}[log(1-D(G(z)))]
\end{equation} 
\begin{equation}
L_2(D, G) = -E_{z\sim N(0,I)}[logD(G(z))]
\end{equation} 
Where $L_1$ and $L_2$ denote a bilevel loss, where $D$ and $G$ being discriminator and generator respectively. $z \sim N (0, I)$ is the noise input of the generator and $w$ is the input of the discriminator. 

\paragraph*{Relationship between RL and GAN} GAN can be viewed as a special actor-critic~\citep{pfau2016connecting}. In the learning architecture of GAN, the generator acts as the actor and the discriminator acts as the critic or environment which gives the real/fake feedback as a reward. However, the actions taken by the actor cannot change the states of the environment, which means that the learning architecture of GAN is a stateless Markov decision process. Also, the actor has no access to the state of the environment and generates data distribution simply conditioned on Gaussian noise, which means that the generator in the GAN framework is a blind actor/agent. In a nutshell, GAN is a special actor-critic where the actor is blind and the whole process is a stateless MDP. \\

The interactive nature of dialogue systems motivates the wide application of reinforcement learning and GAN models in its research. 

\paragraph*{RL for task-oriented dialogue systems} One common application of reinforcement learning in dialogue systems is the reinforced dialogue management in task-oriented systems. Dialogue state tracking and policy learning are two typical modules of a dialogue manager. \cite{huang2020meta} and~\cite{li2020slot} trained the dialogue state tracker with reinforcement learning. Both of them combined a reward manager into their tracker to enhance tracking accuracy. For the policy learning module, reinforcement learning seems to be the best choice since almost all recent related works learned policy with reinforcement learning~\citep{zhang2019budgeted, wang2020task, zhu2020counterfactual, wang2020learning, takanobu2020multi, huang2020semi, xu2020conversational}. The increasing preference of reinforcement learning in policy learning tasks attributes to the characteristic of them: in policy learning tasks, the model predicts a dialogue action (action) based on the states from the DST module (state), which perfectly accords with the function of the agent in the reinforcement learning framework. 

\paragraph*{RL for open-domain dialogue systems} Due to the huge action space needed to generate language directly, many open-domain dialogue systems trained with reinforcement learning framework do not generate responses but instead select responses. Retrieval-based systems have a limited action set and are suitable to be trained in a reinforcement learning scheme. Some works achieved promising performance in retrieval-based dialogue tasks~\citep{bouchacourt2019miss, li2016dialogue, zhao2016towards}. However, retrieval systems fail to generalize in all user messages and may give unrelated responses~\citep{qiu2017alime}, which makes generation-based dialogue systems preferable. Still considering the action space problem, some works build their systems combining retrieval and generative methods~\citep{zhu2018retrieval, serban2017deep}. \cite{zhu2018retrieval} chose to first retrieve a set of n-best response candidates and then generated responses based on the retrieved results and user message. Comparatively,~\cite{serban2017deep} first generated and retrieved candidate responses with different dialogue models and then trained a scoring model with online reinforcement learning to select responses from both generated and retrieved responses. Since training a generative dialogue agent using reinforcement learning from scratch is particularly difficult, first pretraining the agent with supervised learning to warm-start is a good choice. \cite{wu2019self},~\cite{he2020amalgamating},~\cite{williams2016end} and~\cite{yao2016attentional} applied this pretrain-and-finetune strategy on dialogue learning and achieved outstanding performance, which proved that the reinforcement learning can improve the response quality of data-driven chatbots. Similarly, pretrain-and-finetune was also applicable to domain transfer problems. Some works pretrained the model in a source domain and expanded the domain area with reinforcement training~\citep{mo2018personalizing, li2016deep}. 

\paragraph*{RL for knowledge grounded dialogue systems} Some systems use reinforcement learning to select from outside information like persona, document, knowledge graph, etc., and generate responses accordingly. \cite{majumder2020like} and~\cite{jaques2020human} performed persona selection and persona-based response generation simultaneously and trained their agents with a reinforcement framework. \cite{bao2019know} and~\cite{zhao2020knowledge} built document-grounded systems. Similarly, they used reinforcement learning to accomplish document selection and knowledge-grounded response generation. There were also some works combining knowledge graphs into the dialogue systems and treated them as outside knowledge source~\citep{moon2019opendialkg, xu2020conversational}. In a reinforced training framework, the agent chooses an edge based on the current node and state for each step and then combines the knowledge into the response generation process. 

\paragraph*{RL for dialogue related tasks} Dialogue-related tasks like dialogue relation extraction~\citep{li2019entity}, question answering~\citep{hua2020few} and machine reading comprehension~\citep{guo2020interactive} benefit from reinforcement learning as well because of their interactive nature and the scarcity of annotated data. 

\paragraph*{GAN for dialogue systems} The application of GAN in dialogue systems is divided into two streams. The first sees the GAN framework applied to enhance response generation~\citep{li2017adversarial, zhu2018retrieval, wu2019self, he2020amalgamating, zhu2020counterfactual, qin2020dynamic}. The discriminator distinguishes generated responses from human responses, which incentivizes the agent, which is also the generator in GAN, to generate higher-quality responses. Another stream uses GAN as an evaluation tool of dialogue systems~\citep{kannan2017adversarial, bruni2017adversarial}. After training the generator and discriminator as a whole framework, the discriminator is used separately as a scorer to evaluate the performance of a dialogue agent and was shown to achieve a higher correlation with human evaluation compared with traditional reference-based metrics like BLEU, METEOR, ROUGE-L, etc. We discuss the evaluation of dialogue systems as a challenge in Section~\ref{Evaluation Approaches}. 

\subsection{Knowledge Graph Augmented Neural Networks}
\label{Knowledge Graph Augmented Neural Networks}	 
Supervised training with annotated data tries to learn the knowledge distribution of a dataset. However, a dataset is comparatively sparse and thus learning a reliable knowledge distribution needs a huge amount of annotated data~\citep{annervaz2018learning}. 

Knowledge Graph (KG) is attracting more and more research interests in recent years. KG is a structured knowledge source consisting of entities and their relationships~\citep{ji2020survey}. In other words, KG is the knowledge facts presented in graph format. 

\begin{figure}
\begin {center}
\includegraphics[width=0.7\textwidth]{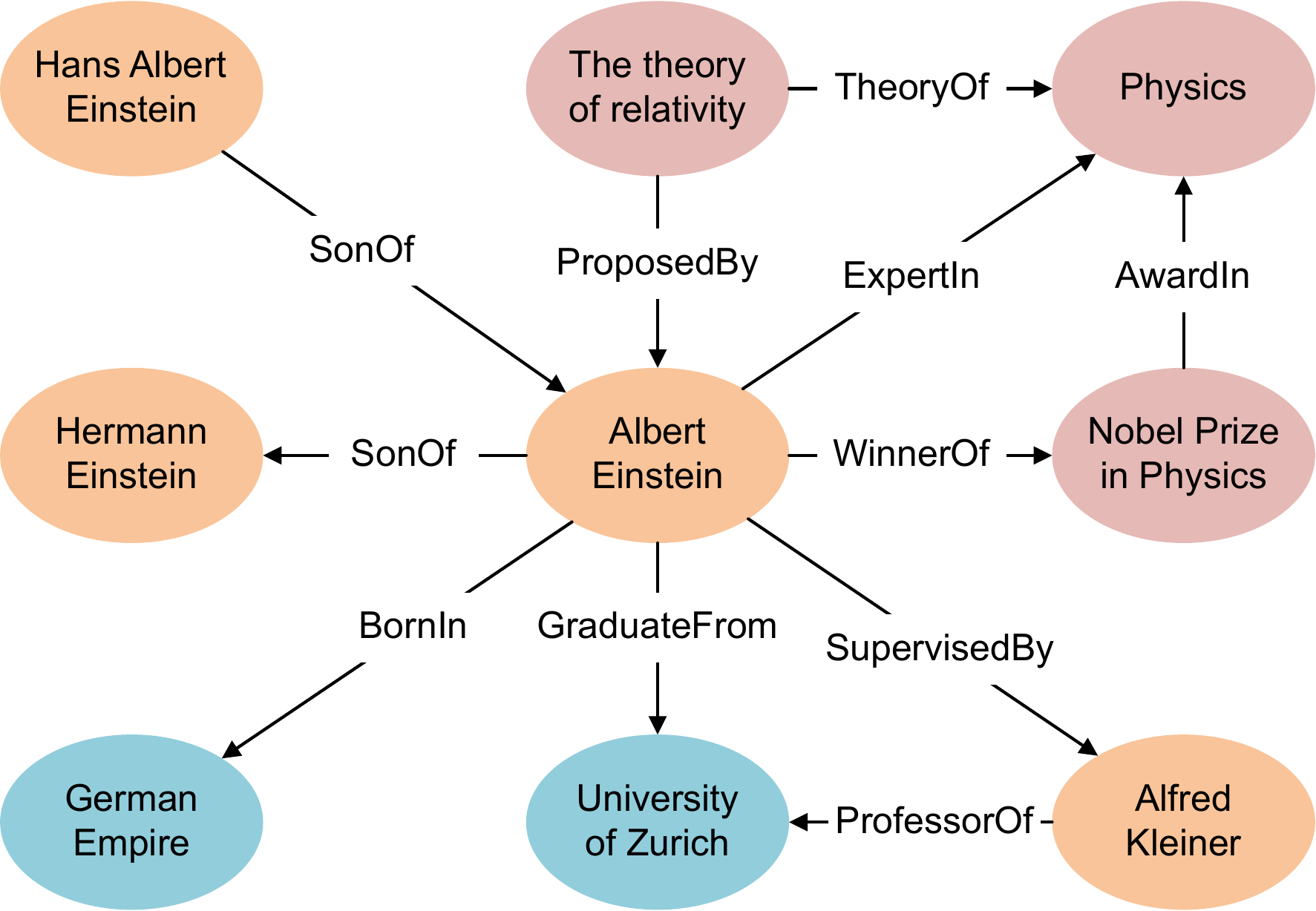}
\caption{Entities and relations in knowledge graph~\citep{ji2020survey}}
\label{Entities and relations in knowledge graph}
\end {center}
\end{figure} 

Figure~\ref{Entities and relations in knowledge graph} shows an example of a KG consisting of entities and their relationships. A KG is stored in triples under the Resource Description Framework (RDF). For example, Albert Einstein, University of Zurich, and their relationship can be expressed as $(Albert Einstein, GraduateFrom, University of Zurich)$. 

Knowledge graph augmented neural networks first represent the entities and their relations in a lower dimension space, then use a neural model to retrieve relevant facts~\citep{ji2020survey}. Knowledge graph representation learning can be generally divided into two categories: structure-based representations and semantically-enriched representations. Structure-based representations use multi-dimensional vectors to represent entities and relations. Models such as TransE~\citep{bordes2013translating}, TransR~\citep{lin2015learning}, TransH~\citep{wang2014knowledge}, TransD~\citep{ji2015knowledge}, TransG~\citep{xiao2015transg}, TransM~\citep{fan2014transition}, HolE~\citep{nickel2016holographic} and ProjE~\citep{shi2017proje} belong to this category. The semantically-enriched representation models like NTN~\citep{socher2013reasoning}, SSP~\citep{xiao2017ssp} and DKRL~\citep{xie2016representation} combine semantic information into the representation of entities and relations. The neural retrieval models also have two main directions: distance-based matching model and semantic matching model. Distance-based matching models~\citep{bordes2013translating} consider the distance between projected entities while semantic matching models~\citep{bordes2014semantic} calculate the semantic similarity of entities and relations to retrieve facts. \\

\paragraph*{Knowledge graph augmented dialogue systems} Knowledge-grounded dialogue systems benefit greatly from the structured knowledge format of KG, where facts are widely intercorrelated. Reasoning over a KG is an ideal approach for combining commonsense knowledge into response generation, resulting in accurate and informative responses~\citep{youaug}. \cite{jung2020attnio} proposed AttnIO, a bi-directional graph exploration model for knowledge retrieval in knowledge-grounded dialogue systems. Attention weights were calculated at each traversing step, and thus the model could choose a broader range of knowledge paths instead of choosing only one node at a time. In such a scheme, the model could predict adequate paths even when only having the destination node as the label. \cite{zhang2019grounded} built ConceptFlow, a dialogue agent that guided to more meaningful future conversations. It traversed in a commonsense knowledge graph to explore concept-level conversation flows. Finally, it used a gate to decide to generate among vocabulary words, central concept words, and outer concept words. \cite{majumder2020like} proposed to generate persona-based responses by first using COMET~\citep{bosselut2019comet} to expand a persona sentence in context along 9 relation types and then applied a pretrained model to generate responses based on dialogue history and the persona variable. \cite{yang2020graphdialog} used knowledge graph as an external knowledge source in task-oriented dialogue systems to incorporate domain-specified knowledge in the response. First, the dialogue history was parsed as a dependency tree and encoded into a fixed-length vector. Then they applied multi-hop reasoning over the graph using the attention mechanism. The decoder finally predicted tokens either by copying from graph entities or generating vocabulary words. \cite{moon2019opendialkg} proposed DialKG Walker for the conversational reasoning task. They computed a zero-shot relevance score between predicted KG embedding and ground KG embedding to facilitate cross-domain predictions. Furthermore, they applied an attention-based graph walker to generate graph paths based on the relevance scores. \cite{huang2020grade} evaluated the dialogue systems by combining the utterance-level contextualized representation and topic-level graph representation. They first constructed the dialogue graph based on encoded (context, response) pairs and then reasoned over the graph to get a topic-level graph representation. The final score was calculated by passing the concatenated vector of contextualized representation and graph representation to a feed-forward network.

\section{Task-oriented Dialogue Systems}
\label{Task-oriented Dialogue Systems} 

This section introduces task-oriented dialogue systems including modular and end-to-end systems. Task-oriented systems solve specific problems in a certain domain such as movie ticket booking, restaurant table reserving, etc. We focus on deep learning-based systems due to the outstanding performance. For readers who want to learn more about traditional rule-based and statistical models, there are several surveys to refer to~\citep{theune2003natural, lemon2007machine, mallios2016survey, chen2017survey, santhanam2019survey}. 

This section is organized as follows. We first discuss modular and end-to-end systems respectively by introducing the principles and reviewing recent works. After that, we comprehensively discuss related challenges and hot topics for task-oriented dialogue systems in recent research to provide some important research directions. 

\begin{figure}
\begin {center}
\includegraphics[width=1\textwidth]{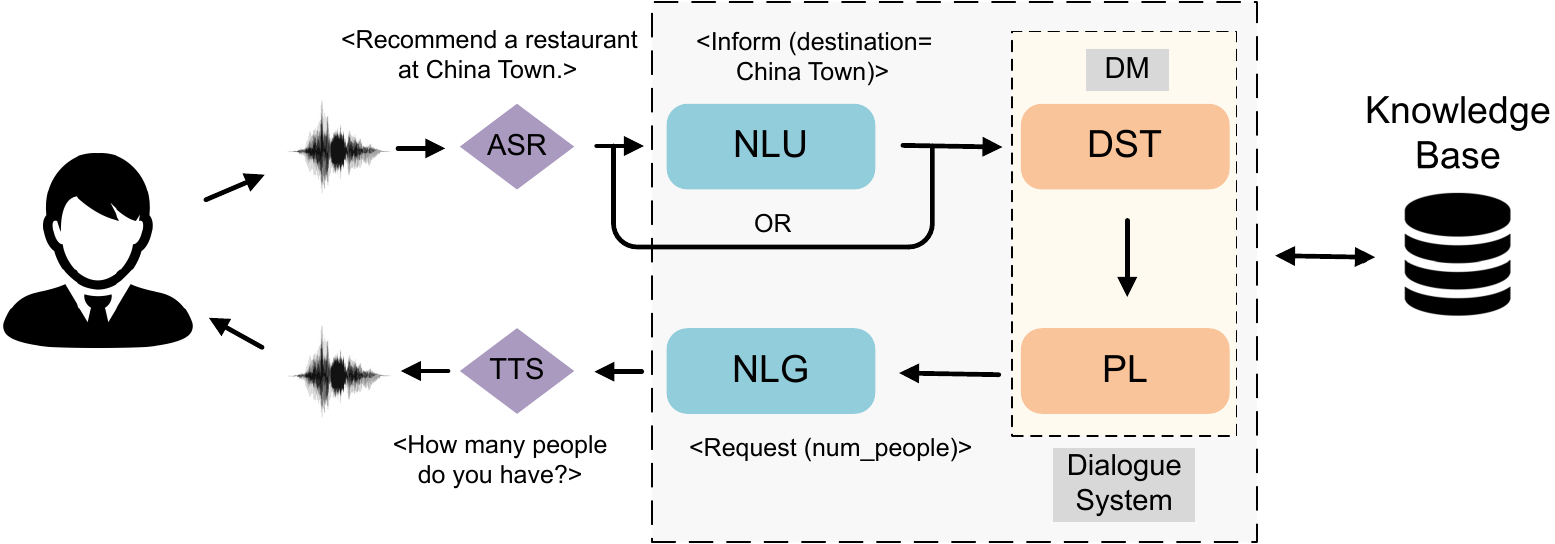}
\caption{Structure of a task-oriented dialogue system in the task-completion pipeline}
\label{Structure of a task-oriented dialogue system in the task-completion pipeline}
\end {center}
\end{figure} 

A task-oriented dialogue system requires stricter response constraints because it aims to accurately handle the user message. Therefore, modular methods were proposed to generate responses in a more controllable way. The architecture of a modular-based system is depicted in Figure~\ref{Structure of a task-oriented dialogue system in the task-completion pipeline}. It consists of four modules:

\textbf{Natural Language Understanding (NLU)}. This module converts the raw user message into semantic slots, together with classifications of domain and user intention. However, some recent modular systems omit this module and use the raw user message as the input of the next module, as shown in Figure~\ref{Structure of a task-oriented dialogue system in the task-completion pipeline}. Such a design aims to reduce the propagation of errors between modules and alleviate the impact of the original error~\citep{kim2018two}. 

\textbf{Dialogue State Tracking (DST)}. This module iteratively calibrates the dialogue states based on the current input and dialogue history. The dialogue state includes related user actions and slot-value pairs. 

\textbf{Dialogue Policy Learning}. Based on the calibrated dialogue states from the DST module, this module decides the next action of a dialogue agent. 

\textbf{Natural Language Generation (NLG)}. This module converts the selected dialogue actions into surface-level natural language, which is usually the ultimate form of response. 

Among them, Dialogue State Tracking and Dialogue Policy Learning constitute the Dialogue Manager (DM), the central controller of a task-oriented dialogue system. Usually, a task-oriented system also interacts with an external Knowledge Base (KB) to retrieve essential knowledge about the target task. For example, in a movie ticket booking task, after understanding the requirement of the user message, the agent interacts with the movie knowledge base to search for movies with specific constraints such as movie name, time, cinema, etc. 

\subsection{Natural Language Understanding}
\label{Natural Language Understanding}
It has been proven that the NLU module impacts the whole system significantly in the term of response quality~\citep{li2017end}. The NLU module converts the natural language message produced by the user into semantic slots and performs classification. Table~\ref{The output example of an NLU module} shows an example of the output format of the NLU module. The NLU module manages three tasks: domain classification, intent detection, and slot filling. Domain classification and intent detection are classification problems, which use classifiers to predict a mapping from the input language sequence to a predefined label set. In the given example, the predicted domain is ``\textit{movie}" and the intent is ``\textit{find\_movie}". Slot filling is a tagging problem, which can be viewed as a sequence-to-sequence task. It maps a raw user message into a sequence of slot names. In the example, the NLU module reads the user message ``\textit{Recommend a movie at Golden Village tonight.}" and outputs the corresponding tag sequence. It recognizes ``\textit{Golden Village}" as the place to go, which is tagged as ``\textit{B\_desti}" and ``\textit{I\_desti}" for the two words respectively. Similarly, the token ``\textit{tonight}" is converted into ``\textit{B\_time}". `B' represents the beginning of a chunk, and `I' indicates that this tag is inside a target chunk. For those unrelated tokens, an `O' is used indicating that this token is outside of any chunk of interest. This tagging method is called Inside-Outside-Beginning (IOB) tagging~\citep{ramshaw1999text}, which is a common method in Named-Entity Recognition (NER) tasks. 

\begin{table}
\caption{The output example of an NLU module}
\label{The output example of an NLU module} 
\centering
\begin{tabular}{|c|ccccccc|}
\hline
\textbf{Sentence} & \cellcolor{grey5}Recommend & \cellcolor{grey5}a & \cellcolor{grey5}movie & \cellcolor{grey5}at & \cellcolor{blue4}Golden & \cellcolor{blue4}Village & \cellcolor{orange4}tonight \\\cline{1-1}
\textbf{Slots} & \cellcolor{grey5}O & \cellcolor{grey5}O & \cellcolor{grey5}O & \cellcolor{grey5}O & \cellcolor{blue4}B-desti & \cellcolor{blue4}I-desti & \cellcolor{orange4}B-time\\\hline
\textbf{Intent} & & & & find\_movie & & & \\\hline
\textbf{Domain} & & & & movie & & & \\\hline
\end{tabular}
\end{table}

\paragraph*{Techniques for domain classification and intent detection} Domain classification and intent detection belong to the same category of tasks. Deep learning methods are proposed to solve the classification problems of dialogue domain and intent. \cite{deng2012use} and~\cite{tur2012towards} were the first who successfully improved the recognition accuracy of dialogue intent. They built deep convex networks to combine the predictions of a prior network and the current utterances as an integrated input of a current network. A deep learning framework was also used to classify the dialogue domain and intent in a semi-supervised fashion~\citep{yann2014zero}. To solve the difficulty of training a deep neural network for domain and intent prediction, Restricted Boltzmann Machine (RBM) and Deep Belief Networks (DBNs) were applied to initialize the parameters of deep neural networks~\citep{sarikaya2014application}. To make use of the strengths of RNNs in sequence processing, some works used RNNs as utterance encoders and made predictions for intent and domain categories~\citep{ravuri2015recurrent, ravuri2016comparative}. \cite{hashemi2016query} used a CNN to extract hierarchical text features for intent detection and illustrated the sequence classification capabilities of CNNs. \cite{lee2016sequential} proposed a model for intent classification of short utterances. Short utterances are hard for intent detection because of the lack of information in a single dialogue turn. This paper used RNN and CNN architectures to incorporate the dialogue history, thus obtaining the context information as an additional input besides the current turn's message. The model achieved promising performances on three intent classification datasets. More recently,~\cite{wu2020tod} pretrained Task-Oriented Dialogue BERT (TOD-BERT) and significantly improved the accuracy in the intent detection sub-task. The proposed model also exhibited a strong capability of few-shot learning and could effectively alleviate the data insufficiency issue in a specific domain. 

\paragraph*{Techniques for slot filling} The slot filling problem is also called semantic tagging, a sequence classification problem. It is more challenging for that the model needs to predict multiple objects at a time. Deep Belief Nets (DBNs) exhibit promising capabilities in the learning of deep architectures and have been applied in many tasks including semantic tagging. \cite{sarikaya2011deep} used a DBN-initialized neural network to complete slot filling in the call-routing task. \cite{deoras2013deep} built a DBN-based sequence tagger. In addition to the NER input features used in traditional taggers, they also combined part of speech (POS) and syntactic features as a part of the input. The recurrent architectures benefited the sequence tagging task in that they could keep track of the information along past timesteps to make the most of the sequential information. \cite{yao2013recurrent} first argued that instead of simply predicting words, RNN Language Models (RNN-LMs) could be applied in sequence tagging. On the output side of RNN-LMs, tag labels were predicted instead of normal vocabularies. \cite{mesnil2013investigation} and~\cite{mesnil2014using} further investigated the impact of different recurrent architectures in the slot filling task and found that all RNNs outperformed the Conditional Random Field (CRF) baseline. As a powerful recurrent model, LSTM showed promising tagging accuracy on the ATIS dataset owing to the memory control of its gate mechanism~\citep{yao2014spoken}. 
\cite{gangadharaiah2020recursive} argued that the shallow output representations of traditional semantic tagging lacked the ability to represent the structured dialogue information. To improve, they treated the slot filling task as a template-based tree decoding task by iteratively generating and filling in the templates. Different from traditional sequence tagging methods,~\cite{coope2020span} tackled the slot filling task by treating it as a turn-based span extraction task. They applied the conversational pretrained model ConveRT and utilized the rich semantic information embedded in the pretrained vectors to solve the problem of in-domain data insufficiency. The inputs of ConveRT are the requested slots and the utterance, while the output is a span of interest as the slot value. 

\paragraph*{Unifying domain classification, intent detection, and slot filling} Some works choose to combine domain classification, intent detection, and slot filling into a multitask learning framework to jointly optimize the shared latent space. \cite{hakkani2016multi} applied a bi-directional RNN-LSTM architecture to jointly perform three tasks. \cite{liu2016attention} augmented the traditional RNN encoder-decoder model with an attention mechanism to manage intent detection and slot filling. The slot filling applied explicit alignment. \cite{chen2016end} proposed an end-to-end memory network and used a memory module to store user intent and slot values in history utterances. Attention was further applied to iteratively select relevant intent and slot values at the decoding stage. Multi-task learning of three NLU subtasks contributed to the domain scaling and facilitated the zero-shot or few-shot training when transferring to a new domain~\citep{bapna2017towards, lee2019zero}. \cite{zhang2018joint} captured the hierarchical structure of dialogue semantics in NLU multi-task learning by applying a capsule-based neural network. With a dynamic routing-by-agreement strategy, the proposed architecture raised the accuracy of both intent detection and slot filling on the SNIPS-NLU and ATIS dataset. 

\paragraph*{Novel perspectives} More recently, some novel ideas appear in NLU research, which provides new possibilities for further improvements. Traditional NLU modules rely on the text converted from the audio message of the user using the Automatic Speech Recognition (ASR) module. However,~\cite{singla2020towards} jumped over the ASR module and directly used audio signals as the input of NLU. They found that by reducing the module numbers of a pipeline system, the predictions were more robust since fewer errors were broadcasted. \cite{su2019dual} argued that Natural Language Understanding (NLU) and Natural Language Generation (NLG) were reversed processes. Thus, their dual relationship could be exploited by training with a dual-supervised learning framework. The experiments exhibited improvement in both tasks.

\subsection{Dialogue State Tracking }
\label{Dialogue State Tracking }
Dialogue State Tracking (DST) is the first module of a dialogue manager. It tracks the user's goal and related details every turn based on the whole dialogue history to provide the information based on which the Policy Learning module (next module) decides the agent action to make. 

\paragraph*{Differences between NLU and DST} The NLU and DST modules are closely related. Both NLU and DST perform slot filling for the dialogue. However, they actually play different roles. The NLU module tries to make classifications for the current user message such as the intent and domain category as well as the slot each message token belongs to. For example, given a user message ``\textit{Recommend a movie at Golden Village tonight.}", the NLU module will convert the raw message into ``$inform (domain = movie;\ destination = Golden Village;\ date = today;\ time = evening)$", where the slots are usually filled by tagging each word of the user message as described in Section~\ref{Natural Language Understanding}. However, the DST module does not classify or tag the user message. Instead, it tries to find a slot value for each slot name in a pre-existing slot list based on the whole dialogue history. For example, there is a pre-existing slot list ``$intent:\_;\ domain:\_;\ name:\_;\ pricerange:\_;\ genre:\_;\ destination:\_;\ date:\_$", where the underscore behind the colon is a placeholder denoting that this place can be filled with a value. Every turn, the DST module will look up the whole dialogue history up to the current turn and decide which content can be filled in a specific slot in the slot list. If the user message ``\textit{Recommend a movie at Golden Village tonight.}" is the only message in a dialogue, then the slot list can be filled as ``$intent: inform;\ domain: movie;\ name: None;\ pricerange: None;\ genre: None;\ destination: Golden Village;\ date: today$", where the slots unspecified by the user up to current turn can be filled with ``$None$". To conclude, the NLU module tries to tag the user message while the DST module tries to find values from the user message to fill in a pre-existing form. Some dialogue systems took the output of the NLU module as the input of DST module~\citep{williams-etal-2013-dialog, henderson2014second, henderson2014third}, while others directly used raw user messages to track the state~\citep{kim2019efficient, wang2020slot, hu2020sas}. 

Dialogue State Tracking Challenges (DSTCs), a series of popular challenges in DST, provides benchmark datasets, standard evaluation frameworks, and test-beds for research~\citep{williams-etal-2013-dialog, henderson2014second, henderson2014third, kim2016fifth, kim2017fourth}. The DSTCs cover many domains such as restaurants, tourism, etc. 

A dialogue state contains all essential information to be conveyed in the response~\citep{44018}. As defined in DSTC2~\citep{henderson2014second}, the dialogue state of a given dialogue turn consists of informable slots \textit{Sinf} and requestable slots \textit{Sreq}. Informable slots are attributes specified by users to constrain the search of the database while requestable slots are attributes whose values are queried by the user. For example, the serial number of a movie ticket is usually a requestable slot because users seldom assign a specific serial number when booking a ticket. Specifically, the dialogue state has three components:
\begin{itemize}
    \item \textbf{Goal constraint corresponding with informable slots}. The constraints can be specific values mentioned by the user in the dialogue or a special value. Special values include \textit{Dontcare} indicating the user's indifference about the slot and \textit{None} indicating that the user has not specified the value in the conversation yet. 
    
    \item \textbf{Requested slots}. It can be a list of slot names queried by the user seeking answers from the agent. 
    
    \item \textbf{Search method of current turn}. It consists of values indicating the interaction categories. \textit{By constraints} denotes that the user tries to specify constraint information in his requirement; \textit{by alternatives} denotes that the user requires an alternative entity; \textit{finished} indicates that the user intends to end the conversation. 
\end{itemize}

However, considering the numerous challenges such as tracking efficiency, tracking accuracy, domain adaptability, and end-to-end training, many alternative representations have been proposed recently, which will be discussed later. 

\begin{figure}
\begin {center}
\includegraphics[width=1\textwidth]{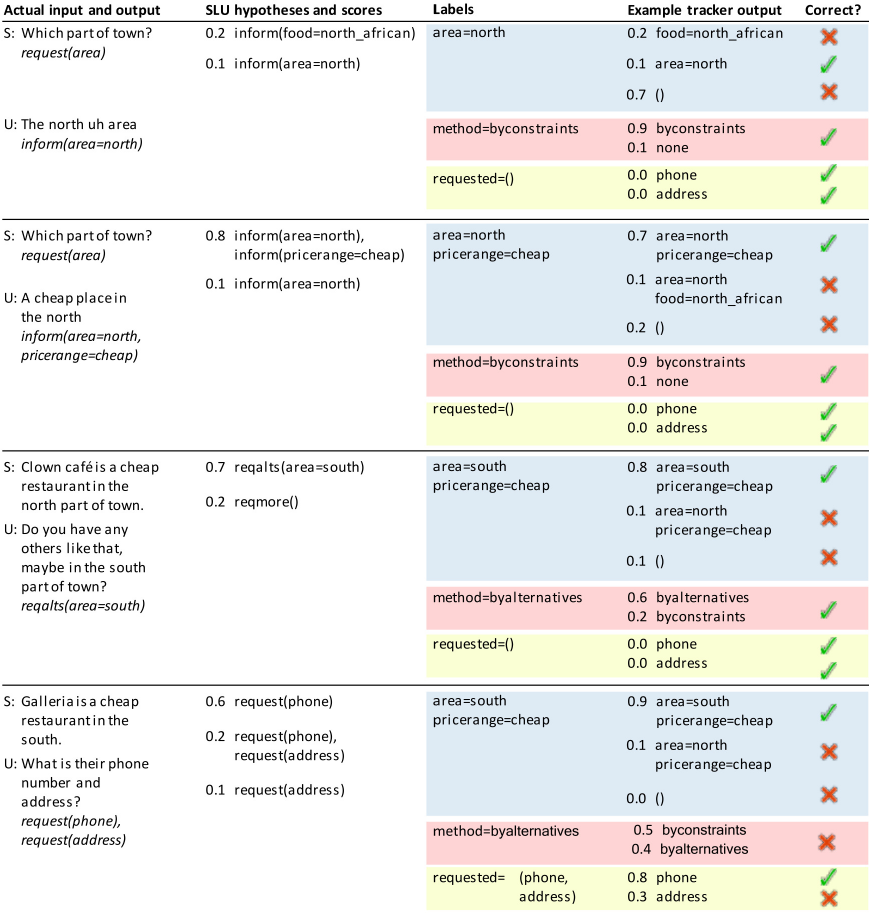}
\caption{An example of DST procedure~\citep{henderson2014second}}
\label{An example Dialogue State Tracking procedure}
\end {center}
\end{figure} 

Figure~\ref{An example Dialogue State Tracking procedure} is an example of the DST process for 4 dialogue turns in a restaurant table booking task. The first column includes the raw dialogue utterances, with $S$ denoting the system message and $U$ denoting the user message. The second column includes the N-best output lists of the NLU module and their corresponding confidence scores. The third column includes the labels of a turn, indicating the ground truth slot-value pairs. The fourth column includes the example DST outputs and their corresponding confidence scores. The fifth column indicates the correctness of the tracker output. 

Earlier works use hand-craft rules or statistical methods to solve DST tasks. While widely used in industry dialogue systems, rule-based DST methods~\citep{goddeau1996form} have many restrictions such as limited generalization, high error rate, low domain adaptability, etc~\citep{williams2014web}. Statistical methods~\citep{lee2013structured, lee2013recipe, ren2013dialog, williams2013multi, williams2014web} also suffer from noisy conditions and ambiguity~\citep{young2010hidden}. 

Recently, many neural trackers have emerged. Neural trackers have multiple advantages over rule-based and statistical trackers. In general, they are categorized into two streams. The first stream has predefined slot names and values, and each turn the DST module tries to find the most appropriate slot-value pairs based on the dialogue history; the second stream does not have a fixed slot value list, so the DST module tries to find the values directly from the dialogue context or generate values based on the dialogue context. Obviously, the latter one is more flexible and in fact, more and more works are solving DST in the second way. We discuss the works of both categories here. 

\paragraph*{Neural trackers with predefined slot names and values} The first stream can be viewed as a multi-class or multi-hop classification task. For multi-class classification DST, the tracker predicts the correct class from multiple values but this method suffers from high complexity when the value set grows large. On the other hand, for the multi-hop classification tasks, the tracker reads only one slot-value pair at a time and performs binary prediction. Working in this fashion reduces the model complexity but raises the system reaction time since for each slot there will be multiple tracking processes. \cite{henderson2013deep} was the first who used a deep learning model in the DST tasks. They integrated many feature functions (e.g., SLU score, Rank score, Affirm score, etc.) as the input of a neural network, then predict the probability of each slot-value pair. \cite{mrkvsic2015multi} applied an RNN as a neural tracker to gain awareness on dialogue context. \cite{mrkvsic2016neural} proposed a multi-hop neural tracker which took the system output and user utterances as the first two inputs (to model the dialogue context), and the candidate slot-value pairs as the third input. The tracker finally made a binary prediction on the current slot-value pair based on the dialogue history. 

\paragraph*{Neural trackers with unfixed slot names and values} The second stream attracts more attention because it not only reduces the model and time complexity of DST tasks but also facilitates end-to-end training of task-oriented dialogue systems. Moreover, it is also flexible when the target domain changes. \cite{lei2018sequicity} proposed belief span, a text span of the dialogue context corresponding to a specific slot. They built a two-stage CopyNet to copy and store slot values from the dialogue history. The slots were stored to prepare for neural response generation. The belief span facilitated the end-to-end training of dialogue systems and increased the tracking accuracy in out-of-vocabulary cases. Based on this,~\cite{lin2020mintl} proposed the minimal belief span and argued that it was not scalable to generate belief states from scratch when the system interacted with APIs from diverse domains. The proposed MinTL framework operated \textit{insertion (INS)}, \textit{deletion (DEL)} and \textit{substitution (SUB)} on the dialogue state of last turn based on the context and the minimal belief span. \cite{wu2019transferable} proposed the TRADE model. The model also applied the copy mechanism and used a soft-gated pointer-generator to generate the slot value based on the domain-slot pair and encoded dialogue context. \cite{quan2020modeling} argued that simply concatenating the dialogue context was not preferable. Alternatively, they used \textit{[sys]} and \textit{[usr]} to discriminate the system and user messages. This simple long context modeling method achieved a 7.03\% improvement compared with the baseline. \cite{cheng2020conversational} proposed Tree Encoder-Decoder (TED) architecture which utilized a hierarchical tree structure to represent the dialogue states and system acts. The TED generated tree-structured dialogue states of the current turn based on the dialogue history, dialogue action, and dialogue state of the last turn. This approach led to a 20\% improvement on the state-of-the-art DST baselines which represented dialogue states and user goals in a flat space. \cite{chen2020parallel} built an interactive encoder to exploit the dependencies within a turn and between turns. Furthermore, they used the attention mechanism to construct the slot-level context for user and system respectively, which were embedding vectors based on which the generator copied values from the dialogue context. \cite{shan2020contextual} applied BERT to perform multi-task learning and generated the dialogue state. They first encoded word-level and turn-level contexts. Then they retrieved the relevant information for each slot from the context by applying both word-level and turn-level attention. Furthermore, the slot values were predicted based on the retrieved information. Similarly,~\cite{wang2020slot} used BERT for slot value prediction. They performed Slot Attention (SA) to retrieve related spans and Value Normalization (VN) to convert the spans into final values. \cite{huang2020meta} proposed Meta-Reinforced MultiDomain State Generator (MERET), which was a dialogue state generator further finetuned with policy gradient reinforcement learning. 

\subsection{Policy Learning}
\label{Policy Learning}
The Policy learning module is the other module of a dialogue manager. This module controls which action will be taken by the system based on the output dialogue states from the DST module. Assuming that we have the dialogue state $S_t$ of the current turn and the action set $A = \{a_1, ..., a_n\}$, the task of this module is to learn a mapping function $f$: $S_t \to a_i \in A$. This module is comparatively simpler than other modules in the term of task definition but actually, the task itself is challenging~\citep{peng2017composite}. For example, in the tasks of movie ticket and restaurant table booking, if the user books a two-hour movie slot and intends to go for dinner after that, then the agent should be aware that the time gap between movie slot and restaurant slot has to be more than two hours since the commuting time from the cinema to the restaurant should be considered. 

Supervised learning and reinforcement learning are mainstream training methods for dialogue policy learning~\citep{chen2017survey}. Policies learned in a supervised fashion exhibit great decision-making ability~\citep{su2016continuously, dhingra2016towards, williams2017hybrid, liu2017iterative}. In some specific tasks, the supervised policy model can complete tasks precisely, but the training process totally depends on the quality of training data. Moreover, the annotated datasets require intensive human labor, and the decision ability is restricted by the specific task and domain, showing weak transferring capability. With the prevalence of reinforcement learning methods, more and more task-oriented dialogue systems use reinforcement learning to learn the policy. The dialogue policy learning fits the reinforcement learning setting since the agent of reinforcement learning learns a policy to map environment states to actions as well. 

Usually, the environment of reinforce policy learning is a user or a simulated user in which setting the training is called online learning. However, it is data- and time-consuming to learn a policy from scratch in the online learning scenario, so the warm-start method is needed to speed up the training process. \cite{henderson2008hybrid} used expert data to restrict the initial action space exploration. \cite{chen2017agent} applied teacher-student learning framework to transfer the teacher expert knowledge to the target network in order to warm-start the system. 

\paragraph*{Reinforcement policy learning techniques} Almost all recent dialogue policy learning works are based on reinforcement learning methods. Online learning is an ideal approach to get training samples iteratively for a reinforcement learning agent, but human labor is very limited. \cite{zhang2019budgeted} proposed Budget-Conscious Scheduling (BCS) to better utilize limited user interactions, where the user interaction is seen as the budget. The BCS used a probability scheduler to allocate the budget during training. Also, a controller decided whether to use real user interactions or simulated ones. Furthermore, a goal-based sampling model was applied to simulate the experiences for policy learning. Such a budget-controlling mechanism achieved ideal performance in the practical training process. Considering the difficulty of getting real online user interactions and the huge amount of annotated data required for training user simulators,~\cite{takanobu2020multi} proposed Multi-Agent Dialog Policy Learning, where they have two agents interacting with each other, performing both user and agent, learning the policy simultaneously. Furthermore, they incorporated a role-specific reward to facilitate role-based response generation. A High task completion rate was observed in experiments. \cite{wang2020task} introduced Monte Carlo Tree Search with Double-q Dueling network (MCTS-DDU), where a decision-time planning was proposed instead of background planning. They used the Monte Carlo simulation to perform a tree search of the dialogue states. \cite{gordon2020learning} trained expert demonstrators in a weakly supervised fashion to perform Deep Q-learning from Demonstrations (DQfD). Furthermore, Reinforced Fine-tune Learning was proposed to facilitate domain transfer. In reinforce dialogue policy learning, the agent usually receives feedback at the end of the dialogue, which is not efficient for learning. \cite{huang2020semi} proposed an innovative reward learning method that constrains the dialogue progress according to the expert demonstration. The expert demonstration could either be annotated or not, so the approach was not labor intensive. \cite{wang2020multi} proposed to co-generate the dialogue actions and responses to maintain the inherent semantic structures of dialogue. Similarly,~\cite{le2020uniconv} proposed a unified framework to simultaneously perform dialogue state tracking, dialogue policy learning, and response generation. Experiments showed that unified frameworks have a better performance both in their sub-tasks and in their domain adaptability. \cite{xu2020conversational} used a knowledge graph to provide prior knowledge of the action set and solved policy learning task in a graph-grounded fashion. By combining a knowledge graph, a long-term reward was obtained to provide the policy agent with a long-term vision while choosing actions. Also, the candidate actions were of higher quality due to prior knowledge. The policy learning was further performed in a more controllable way. 

\subsection{Natural Language Generation}
\label{Natural Language Generation}
Natural Language Generation (NLG) is the last module of a task-oriented dialogue system pipeline. It manages to convert the dialogue actions generated from the dialogue manager into a final natural language representation. E.g., Assuming ``\textit{Inform (name = Wonder Woman;\ genre = Action;\ desti = Golden Village)}" to be the dialogue action from policy learning module, then the NLG module converts it into language representations such as ``\textit{There is an action movie named Wonder Woman at Golden Village.}"

\begin{figure}
\begin {center}
\includegraphics[width=1\textwidth]{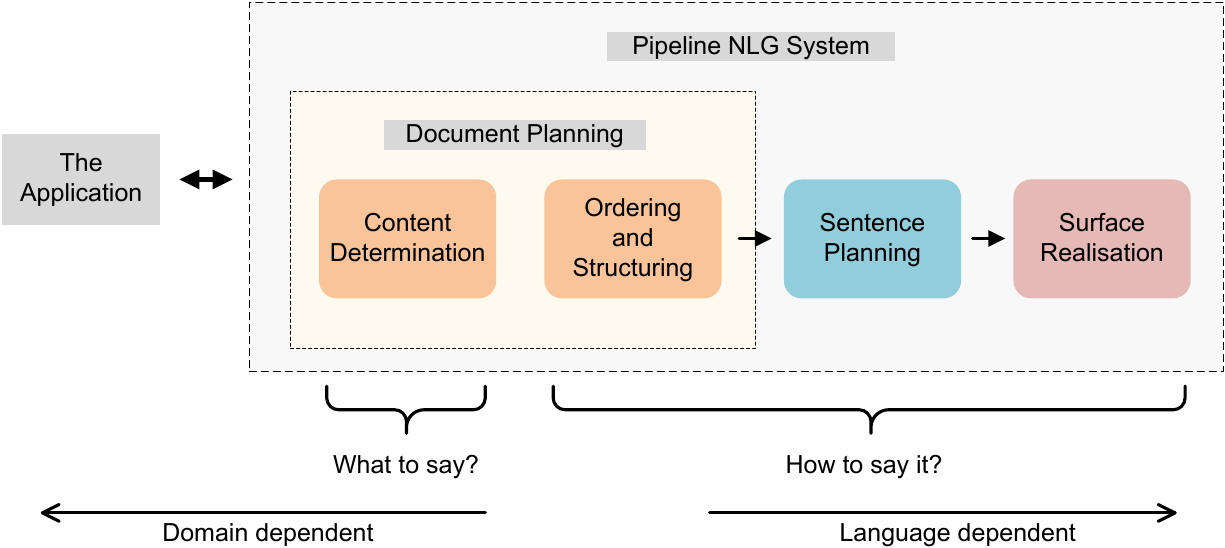}
\caption{The pipeline NLG system}
\label{The pipeline NLG system}
\end {center}
\end{figure}

Traditional NLG modules are pipeline systems. Defined by~\cite{siddharthan2001ehud}, the standard pipeline of NLG consists of four components, as shown in Figure~\ref{The pipeline NLG system}. 

The core modules of this pipeline are Content Determination, Sentence Planning, and Surface Realization, as proposed by~\cite{reiter1994has}. \cite{cahill1999search} further improved the NLG pipeline by adding three more components: lexicalization, referring expression generation, and aggregation. However, this model has a drawback that the input of the system is ambiguous. 

\paragraph*{End-to-end NLG techniques} Deep learning methods were further applied to enhance the NLG performance and the pipeline is collapsed into a single module. End-to-end natural language generation has achieved promising improvements and is the most popular way to perform NLG in recent work. \cite{wen2015stochastic} argued that language generation should be fully data-driven and not depend on any expert rules. They proposed a statistical language model based on RNNs to learn response generation with semantic constraints and grammar trees. Additionally, they used a CNN reranker to further select better responses. Similarly, an LSTM model was used by~\cite{wen2015semantically} to learn sentence planning and surface realization simultaneously. \cite{tran2017semantic} further improved the generation quality on multiple domains using GRU. The proposed generator consistently generated high-quality responses on multiple domains. To improve the domain adaptability of recurrent models,~\cite{wen2016multi} proposed to first train the recurrent language model on data synthesized from out-of-domain datasets, then finetune on a comparatively smaller in-domain dataset. This training strategy was proved effective in human evaluation. Context-awareness is important in dialogue response generation because only depending on the dialogue action of the current turn may cause illogical responses. \cite{zhou2016context} built an attention-based Context-Aware LSTM (CA-LSTM) combining target user questions, all semantic values, and dialogue actions as input to generate context-aware responses in QA. Likewise,~\cite{duvsek2016context} concatenated the preceding user utterance with the dialogue action vector and fed it into an LSTM model. \cite{duvsek2016sequence} put a syntax constraint upon their neural response generator. A two-stage sequence generation process was proposed. First, a syntax dependency tree was generated to have a structured representation of the dialogue utterance. The generator in the second stage integrated sentence planning and surface realization and produced natural language representations. 

\paragraph*{Robust natural language generation} More recent works have focused on the reliability and quality of generated responses. A tree-structured semantic representation was proposed by~\cite{balakrishnan2019constrained} to achieve better content planning and surface realization performance. They further designed a novel beam search algorithm to improve the semantic correctness of the generated response. To avoid mistakes such as slot value missing or redundancy in generated responses,~\cite{li2020slot} proposed Iterative Rectification Network (IRN), a framework trained with supervised learning and finetuned with reinforcement learning. It iteratively rectified generated tokens by incorporating slot inconsistency penalty into its reward. \cite{golovanov2019large} applied large-scale pretrained models for NLG tasks. After comparing single-input and multi-input methods, they concluded that different types of input context will cause different inductive biases in generated responses and further proposed to utilize this characteristic to better adapt a pretrained model to a new task. \cite{baheti2020fluent} solved NLG reliability problem in conversational QA. Though with different pipeline structures, they used similar methods to increase the fluency and semantic correctness of the generated response. They proposed Syntactic Transformations (STs) to generate candidate responses and used a BERT to rank their qualities. These generated responses can be viewed as an augmentation of the original dataset to be further used in NLG model learning. \cite{oraby2019curate} proposed a method to create datasets with rich style markups from easily available user reviews. They further trained multiple NLG models based on generated data to perform joint control of semantic correctness and language style. Similarly,~\cite{elder2020make} put forward a data augmentation approach which put a restriction on response generation. Though this restriction caused dull and less diverse responses, they argued that in task-oriented systems, reliability was more important than diversity.  

\subsection{End-to-end Methods}
\label{End-to-end methods}
The modules discussed above can achieve good performance in their respective tasks, with the help of recent relevant advances. However, there exist two significant drawbacks in modular systems~\citep{zhao2016towards}: (1) Modules in many pipeline systems are sometimes not differentiable, which means that errors from the end are not able to be propagated back to each module. In real dialogue systems training, usually the only signal is the user response, while other supervised signals like dialogue states and dialogue actions are scarce. (2) Though the modules jointly contribute to the success of a dialogue system, the improvement of one module may not necessarily raise the response accuracy or quality of the whole system. This causes additional training of other modules, which is labor intensive and time-consuming. Additionally, due to the handcrafted features in pipeline task-oriented systems such as dialogue states, it is usually hard to transfer modular systems to another domain, since the predefined ontologies require modification. 

There exist two main methods for the end-to-end training of task-oriented dialogue systems. One is to make each module of a pipeline system differentiable, then the whole pipeline can be viewed as a large differentiable system and the parameters can be optimized by back-propagation in an end-to-end fashion~\citep{le2020uniconv}. Another way is to use only one end-to-end module to perform both knowledge base retrieval and response generation, which is usually a multi-task learning neural model. 

\paragraph*{End-to-end trainable pipeline TOD} The increasing applications of neural models have made it possible for modules to be differentiable. While many modules are easily differentiable, there remains one task that makes differentiation challenging: the knowledge base query. Many task-oriented dialogue systems require an external knowledge source to retrieve related knowledge facts required by the user. For example, in the restaurant table booking task, the knowledge fact can be an available slot of one specific restaurant. Traditional methods use a symbolic query to match entries based on their attributes. The system performs semantic parsing on the user message to represent a symbolic query according to the user goal~\citep{li2017end, williams2016end, wen2016network}. However, this retrieval process is not differentiable, which prevents the whole framework from being end-to-end trainable. With the application of key-value memory networks~\citep{miller2016key},~\cite{eric2017key} used the key-value retrieval mechanism to retrieve relevant facts. The proposed architecture was augmented with the attention mechanism to compute the relevance between utterance representations of dialogue and key representations of the knowledge base. \cite{dhingra2016towards} presented a soft retrieval mechanism that uses a ``soft" posterior distribution over the knowledge base to replace the symbolic queries. They further combined this soft retrieval mechanism into a reinforcement learning framework to achieve complete end-to-end training based on user feedback. \cite{williams2017hybrid} proposed Hybrid Code Networks (HCNs), which encoded domain-specific knowledge into software and system action templates, achieving the differentiability of the knowledge retrieval module. They did not explicitly model the dialogue states but instead learned the latent representation and optimized the HCN using supervised learning and reinforcement learning jointly. \cite{ham2020end} used GPT-2 to form a neural pipeline and perform domain prediction, dialogue state tracking, policy learning, knowledge retrieval, and response generation in a pipeline fashion. The system could easily interact with external systems because it outputs explicit intermediate results from each module and thus being interpretable. Likewise,~\cite{hosseini2020simple} built a neural pipeline with GPT-2 and explicitly generated results for each neural module as well. 

\paragraph*{End-to-end trainable single module TOD} More recent works tend not to build their end-to-end systems in a pipeline fashion. Instead, they use complex neural models to implicitly represent the key functions and integrate the modules into one. Research in task-oriented end-to-end neural models focuses either on training methods or model architecture, which are the keys to response correctness and quality.~\cite{wang2019incremental} proposed an incremental learning framework to train their end-to-end task-oriented system. The main idea is to build an uncertainty estimation module to evaluate the confidence of appropriate responses generated. If the confidence score was higher than a threshold, then the response would be accepted, while a human response would be introduced if the confidence score was low. The agent could also learn from human responses using online learning. \cite{dai2020learning} used model-agnostic meta-learning (MAML) to improve the adaptability and reliability jointly with only a handful of training samples in a real-life online service task. Similarly,~\cite{qian2019domain} also trained the end-to-end neural model using MAML to facilitate the domain adaptation, which enables the model to first train on rich-resource tasks and then on new tasks with limited data. \cite{lin2020mintl} proposed Minimalist Transfer Learning (MinTL) to plug-and-play large-scale pretrained models for domain transfer in dialogue task completion. To maintain the sequential correctness of generated responses,~\cite{wu2019self} trained an inconsistent order detection module in an unsupervised fashion. This module detected whether an utterance pair is ordered or not to guide the task-completing agent towards generating more coherent responses. \cite{he2020amalgamating} proposed a ``Two-Teacher One-Student" training framework. At the first stage, the two teacher models were trained in a reinforcement learning framework, with the objective of retrieving knowledge facts and generating human-like responses respectively. Then at the second stage, the student network was forced to mimic the output of the teacher networks. Thus, the expert knowledge of the two teacher networks was transferred to the student network. \cite{balakrishnan2019constrained} introduced a constrained decoding method to improve the semantic correctness of the responses generated by the proposed end-to-end system. Many end-to-end task-oriented systems used a memory module to store relevant knowledge facts and dialogue history. \cite{chen2019working} argued that a single memory module was not enough for precise retrieval. They used two long-term memory modules to store the knowledge tuples and dialogue history respectively, and then a working memory was applied to control the token generation. \cite{zhang2020probabilistic} proposed LAtent BElief State (LABES) model, which treated the dialogue states as discrete latent variables to reduce the reliance on turn-level DST labels. To solve the data insufficiency problem in some tasks,~\cite{gao2020paraphrase} augmented the response generation model with a paraphrase model in their end-to-end system. The paraphrase model was jointly trained with the whole framework and it aimed to augment the training samples. \cite{yang2020graphdialog} leveraged the graph structure information of both a knowledge graph and the dialogue context-dependency tree. They proposed a recurrent cell architecture to learn representations on the graph and performed multi-hop reasoning to exploit the entity links in the knowledge graph. With the augmentation of graph information, consistent improvement was achieved on two task-oriented datasets.

\subsection{Research Challenges and Hot Topics}
\label{Research Challenges and Hot Topics}

In this section, we review recent works in task-oriented dialogue systems and point out the frequently studied topics to provide some important research directions. This section can be seen as an augmentation of the literature review in previous sections discussing techniques developed for each module, and focuses more on some specific problems to be solved in the current research community. 

\subsubsection{Pretrained Models for NLU}
The Natural Language Understanding task converts the user message into a predefined format of semantic slots. A popular way to perform NLU is by finetuning large-scale pretrained language models. \cite{wu2020probing} compared many pretrained language models including BERT-based and GPT-based systems in three subtasks of task-oriented dialogue systems - domain identification, intent detection, and slot tagging. This empirical paper is aimed to provide insights and guidelines in pretrained model selection and application for related research. \cite{wu2020tod} pretrained TOD-BERT and outperformed strong baselines in the intent detection task. The model proposed also had a strong few-shot learning ability to alleviate the data insufficiency problem. \cite{coope2020span} proposed Span-ConveRT, which was a pretrained model designed for slot filling task. It viewed the slot filling task as a turn-based span extraction problem and also performed well in the few-shot learning scenario.  

\subsubsection{Domain Transfer for NLU}
Another challenge or hot topic in NLU research is the domain transfer problem, which is also the key issue of task-oriented dialogue systems. \cite{hakkani2016multi} built an RNN-LSTM architecture for multitask learning of domain classification, intent detection, and slot-filling problem. Training samples from multiple domains were combined in a single model where respective domain data reinforces each other. \cite{bapna2017towards} used a multi-task learning framework to leverage slot name encoding and slot description encoding, thus implicitly aligning the slot-filling model across domains. Likewise,~\cite{lee2019zero} also applied slot description to exploit the similar semantic concepts between slots of different domains, which solved the sub-optimal concept alignment and long training time problems encountered in past works involving multi-domain slot-filling. 

\subsubsection{Domain Transfer for DST}
Domain adaptability is also a significant topic for dialogue state trackers. The domain transfer in DST is challenging due to three main reasons~\citep{ren2018towards}: (1) Slot values in ontologies are different when the domain changes, which accounts for the incompatibility of models. (2) When the domain changes, the slot number will also change, causing different numbers of model parameters. (3) Hand-crafted lexicons make it difficult for generalization over domains. \cite{mrkvsic2015multi} used delexicalized n-gram features to solve the domain incompatibility problem by replacing all specified slot names and values with generic symbols. \cite{lin2020mintl} introduced Levenshtein belief spans (Lev), which were short context spans relating to the user message. Different from previous methods which generated dialogue state from scratch, they performed substitution (SUB), deletion (DEL), and insertion (INS) based on past states to alleviate the dependency on annotated in-domain training samples. \cite{huang2020meta} applied model-agnostic meta-learning (MAML) to first learn on several source domains and then adapt on the target domain, while~\cite{campagna2020zero} improved the zero-shot transfer learning by synthesizing in-domain data using an abstract conversation model and the domain ontology. \cite{ouyang2020dialogue} modeled explicit slot connections to exploit the existing slots appearing in other domains. Thus, the tracker could copy slot values from the connected slots directly, alleviating the burden of reasoning and learning. \cite{wang2020slot} proposed Value Normalization (VN) to convert supporting dialogue spans into state values and could achieve high accuracy with only 30\% available ontology. 

\subsubsection{Tracking Efficiency for DST}
Tracking efficiency is another hot topic in dialogue state tracking challenges. Usually, there are multiple states within a dialogue, so how to compute the slot values without any redundant steps becomes very significant when attempting to reduce the reaction time of a system. \cite{kim2019efficient} argued that predicting the dialogue state from scratch at every turn was not efficient. They proposed to first predict the operations to be taken on each of the slots (i.e., Carryover, Delete, Dontcare, Update), and then perform respective operations as predicted. \cite{ouyang2020dialogue} used a slot connection mechanism to directly copy slot values from the source slot, which reduced the expense of reasoning. \cite{hu2020sas} and~\cite{wang2020slot} proposed slot attention to calculate the relations between the slot and dialogue context, thus only focusing on the relevant slots at each turn.  

\subsubsection{Training Environment for PL}
The environment of the Policy Learning framework has been a long-existing problem. \cite{li2017end} built a user simulator to model the user feedback as the reward signal of an environment. They modeled a stack-like user agenda to iteratively change the user goal and thus shifting the dialogue states. While using a user simulator for environment modeling seems to be promising for that it involves less human interaction,~\cite{zhang2019budgeted} argued that training a user simulator required a large amount of annotated data. \cite{takanobu2020multi} proposed Multi-Agent Dialog Policy Learning, where they have two agents interact with each other, performing both user and agent, learning policy simultaneously. Furthermore, they incorporated a role-specific reward to facilitate role-based response generation and here both agents also acted as the environment of the other one. 

\subsubsection{Response Consistency for NLG}
Response consistency in NLG is a challenging problem since it cannot be solved by simply augmenting the training samples. Instead, additional corrections or regulations should be designed. \cite{wen2015semantically} proposed the Semantically Controlled LSTM (SC-LSTM) which used a semantic planning gate to control the retention or abandonment of dialogue actions thus ensuring the response consistency. Likewise,~\cite{tran2017semantic} also applied a gating mechanism to jointly perform sentence planning and surface realization where dialogue action features were gated before entering GRU cells. \cite{li2020slot} proposed Iterative Rectification Network (IRN), which combined a slot inconsistency reward into the reinforcement learning framework. Thus, the model iteratively checked the correctness of slots and corresponding values. 

\subsubsection{End-to-end Task-oriented Dialogue Systems}
End-to-end systems are usually fully data-driven, which contributes to their robust and natural responses. However, because of the finiteness of annotated training samples, a hot research topic is figuring out how to increase the response quality of end-to-end task-oriented dialogue systems with limited data. Using rule-based methods to constrain response generation is a way to improve response quality. \cite{balakrishnan2019constrained} used linearized tree-structured representation as input to obtain control over discourse-level and sentence-level semantic concepts. \cite{kale-rastogi-2020-template} used templates to improve the semantic correctness of generated responses. They broke down the response generation into a two-stage process: first generating semantically correct but possibly incoherent responses based on the slots, with the constraint of templates; then in the second stage, pretrained language models were applied to re-organize the generated utterances into coherent ones. Training the network with reinforcement learning was another strategy to alleviate the reliance on annotated data. \cite{he2020amalgamating} trained two teacher networks using a reinforcement learning framework with the objectives of knowledge retrieval and response generation respectively. Then the student network learns to produce responses by mimicking the output of teacher networks. Training the network in a supervised way,~\cite{dai2020learning} alternatively tried to optimize the learning strategy to improve the learning efficiency of models given limited data. They combined the meta-learning algorithm with human-machine interaction and achieved significant improvement compared with strong baselines not trained with the meta-learning algorithms. A more direct way to solve the data finiteness problem in supervised learning was augmenting the dataset~\citep{elder2020make}, which also improved the response quality to some extent. Additionally, pretraining large-scale models on common corpus and then applying them in a domain that lacks annotated data is a popular approach in recent years~\citep{henderson2019training, mehri2019pretraining, bao2019plato}.

\subsubsection{Retrieval Methods for Task-oriented Dialogue Systems}
Retrieval-based methods are rare in task-oriented systems for the insufficiency of candidate entries to cover all possible responses which usually involve specific knowledge from external knowledge-base. However,~\cite{henderson2019training} argued that in some situations not relating with specific knowledge facts, retrieval-based methods were more precise and effective. They first pretrained the response selection model on general domain corpora and then finetuned on small target domain data. Experiments on six datasets from different domains proved the effectiveness of the pretrained response selection model. \cite{lu2019constructing} constructed Spatio-temporal context features to facilitate response selection, and achieved significant improvements on the Ubuntu IRC dataset. 

\section{Open-Domain Dialogue Systems}
\label{Open-Domain Dialogue Systems}

This section discusses open-domain dialogue systems, which are also called chit-chat dialogue systems or non-task-oriented dialogue systems. Almost all state-of-the-art open-domain dialogue systems are based on neural methods. We organize this section by first briefly introducing the concepts of different branches of open-domain dialogue systems, and then we focus on different research challenges and hot topics. We view these challenges and hot topics as different research directions in open-domain dialogue systems. 

Instead of managing to complete tasks, open-domain dialogue systems aim to perform chit-chat with users without the task and domain restriction~\citep{ritter2011data} and are usually fully data-driven. Open-domain dialogue systems are generally divided into three categories: generative systems, retrieval-based systems, and ensemble systems. Generative systems apply sequence-to-sequence models to map the user message and dialogue history into a response sequence that may not appear in the training corpus. By contrast, retrieval-based systems try to find a pre-existing response from a certain response set. Ensemble systems combine generative methods and retrieval-based methods in two ways: retrieved responses can be compared with generated responses to choose the best among them; generative models can also be used to refine the retrieved responses~\citep{zhu2018retrieval, song2016two, qiu2017alime, serban2017deep}. Generative systems can produce flexible and dialogue context-related responses while sometimes they lack coherence and tend to make dull responses. Retrieval-based systems select responses from human response sets and thus are able to achieve better coherence in surface-level language. However, retrieval systems are restricted by the finiteness of the response sets and sometimes the responses retrieved show a weak correlation with the dialogue context~\citep{zhu2018retrieval}. 

In the next few subsections, we discuss some research challenges and hot topics in open-domain dialogue systems. We aim to to help researchers quickly grasp the current research trends via a systematic discussion on articles solving certain problems.

\subsection{Context Awareness}
\label{Context Awareness}

Dialogue context consists of user and system messages and is an important source of information for dialogue agents to generate responses because dialogue context decides the conversation topic and user goal~\citep{serban2017hierarchical}. A context-aware dialogue agent responds not only depending on the current message but also based on the conversation history. The earlier deep learning-based systems added up all word representations in dialogue history or used a fixed-size window to focus on the recent context~\citep{sordoni2015neural, li2015diversity}. \cite{serban2016building} proposed Hierarchical Recurrent Encoder-Decoder (HRED), which was ground-breaking in building context-awareness dialogue systems. They built a word-level encoder to encode utterances and a turn-level encoder to further summarize and deliver the topic information over past turns. \cite{xing2018hierarchical} augmented the hierarchical neural networks with the attention mechanism to help the model focus on more meaningful parts of dialogue history. 

Both generative and retrieval-based systems rely heavily on dialogue context modeling. \cite{shen2019modeling} proposed Conversational Semantic Relationship RNN (CSRR) to model the dialogue context in three levels: utterance-level, pair-level, and discourse-level, capturing content information, user-system topic, and global topic respectively. \cite{zhang2019recosa} argued that the hierarchical encoder-decoder does not lay enough emphasis on certain parts when the decoder interacted with dialogue contexts. Also, they claimed that attention-based HRED models also suffered from position bias and relevance assumption insufficiency problems. Therefore, they proposed ReCoSa, whose architecture was inspired by the transformer. The model first used a word-level LSTM to encode dialogue contexts, and then self-attention was applied to update the utterance representations. In the final stage, an encoder-decoder attention was computed to facilitate the response generation process. Additionally,~\cite{mehri2019pretraining} examined several applications of large-scale pretrained models in dialogue context learning, providing guidance for large-scale network selection in context modeling.

Some works propose structured attention to improve context-awareness. \cite{qiu2020structured} learned structured dialogue context by combining structured attention with a Variational Recurrent Neural Network (VRNN). Comparatively,~\cite{ferracane2019evaluating} examined the RST discourse tree model proposed by~\cite{liu2018learning} and observed little or even no discourse structures in the learned latent tree. Thus, they argued that structured attention did not benefit dialogue modeling and sometimes might even harm the performance. 

Interestingly,~\cite{feng2020regularizing} not only utilized dialogue history, but also future conversations. Considering that in real inference situations dialogue agents cannot be explicitly aware of future information, they first trained a scenario-based model jointly on past and future context and then used an imitation framework to transfer the scenario knowledge to a target network. 

Better context modeling improves the response selection performance in retrieval-based dialogue systems \citep{jia2020multi}. \cite{tao2019one} proposed Interaction-over-Interaction network (IoI), which consisted of multiple interaction blocks to perform deeper interactions between dialogue context and candidate responses. \cite{jia2020multi} organized the dialogue history into conversation threads by performing classifications on their dependency relations. They further used a pretrained Transformer model to encode the threads and candidate responses to compute the matching score. \cite{lin2020world} argued that response-retrieval datasets should not only be annotated with relevant or irrelevant responses. Instead, a greyscale metric should be used to measure the relevance degree of a response given the dialogue context, thus increasing the context-awareness ability of retrieval models. 

Dialogue rewriting problem aims to convert several messages into a single message conveying the same information and dialogue context awareness is very crucial to this task~\citep{xu2020semantic}. \cite{su2019improving} modeled multi-turn dialogues via dialogue rewriting and benefited from the conciseness of rewritten utterances. 

\subsection{Response Coherence}
\label{Response Coherence}

Coherence is one of the qualities that a good generator seeks~\citep{stent2005evaluating}. Coherence means maintaining logic and consistency in a dialogue, which is essential in an interaction process for that a response with weak consistency in logic and grammar is hard to understand. Coherence is a hot topic in generative systems but not in retrieval-based systems because candidate responses in retrieval methods are usually human responses, which are naturally coherent. 

Refining the order or granularity of sentence functions is a popular strategy for improving the language coherence.~\cite{wu2019self} improved the response coherence via the task of inconsistent order detection. The dialogue systems learned response generation and order detection jointly, which was self-supervised multi-task learning. \cite{xu2019neural} presented the concept of meta-words. Meta-words were diverse attributes describing the response. Learning dialogue based on meta-words helped promote response generation in a more controllable way. \cite{liu2019vocabulary} used three granularities of encoders to encode raw words, low-level clusters, and high-level clusters. The architecture was called Vocabulary Pyramid Network (VPN), which performed a multi-pass encoding and decoding process on hierarchical vocabularies to generate coherent responses. \cite{shen2019modeling} also built a three-level hierarchical dialogue model to capture richer features and improved the response quality. \cite{ji2020cross} built Cross Copy Networks (CCN), which used a copy mechanism to copy from similar dialogues based on the current dialogue context. Thus, the system benefited from the pre-existing coherent responses, which alleviated the need of performing the reasoning process from scratch. 

Many work employ strategies to achieve response coherence on a higher level, which improves the overall quality of the generated responses. \cite{li2019don} improved the logical consistency of generated utterances by incorporating an unlikelihood loss to control the distribution mismatches. \cite{bao2019know} proposed a Generation-Evaluation framework that evaluated the qualities, including coherence, of the generated response. The feedback was further seen as a reward signal in the reinforcement learning framework and guided to a better dialogue strategy via policy gradient, thus improving the response quality. \cite{gao2020dialogue} raised response quality by ranking generated responses based on user feedbacks like upvotes, downvotes, and comments on social networks. \cite{zhu2018retrieval} built a retrieval-enhanced generation model, which enhanced the generated responses in two ways. First, a discriminator was trained with the help of a retrieval system, and then the generator was trained in a GAN framework under the supervision signal of a discriminator. Second, retrieved responses were also used as a part of the generator input to provide a coherent example for the generator. \cite{xu2020conversational} achieved a global coherent dialogue by constructing a knowledge graph from corpora. They further performed graph walks to decide ``what to say" and ``how to say", thus improving the dialogue flow coherence. \cite{mesgar2019dialogue} proposed an assessment approach for dialogue coherence evaluation by combining the dialogue act prediction in a multi-task learning framework and learned rich dialogue representations. 

There also evolve some data-wise methods for better response coherence.~\cite{bi2019fine} proposed to annotate sentence functions in existing conversation datasets to improve the sentence logic and coherence of generated responses. \cite{akama2020filtering} focused on data effectiveness as well. They filtered out low-quality utterance pairs by scoring the relatedness and connectivity, which was proved to be effective in improving the response coherence. \cite{akama2020filtering} presented a method for evaluating dataset utterance pairs' quality in terms of connectedness and relatedness. The proposed scoring technique is based on research findings that have been widely disseminated in the conversation and linguistics communities. \cite{lison2017not} included a weighting model in their neural architecture. The weighting model, which is based on conversation data, assigns a numerical weight to each training sample that reflects its intrinsic quality for dialogue modeling and achieved good result in experiments.

\subsection{Response Diversity}
\label{Response Diversity}

The Bland and generic response is a long-existing problem in generative dialogue systems. Because of the high frequency of generic responses like ``\textit{I don't know}" in training samples and the beam search decoding scheme of neural sequence-to-sequence models, generative dialogue systems tend to respond with universally acceptable but meaningless utterances~\citep{serban2016building, vinyals2015neural, sordoni2015neural}. For example, to respond to the user message ``\textit{I really want to have a meal}", the agent tends to choose simple responses like ``\textit{It's OK}" instead of responding with more complicated sentences like recommendations and suggestions. 

Earlier works solve this challenge by modifying the decoding objective or adding a reranking process. \cite{li2015diversity} replaced the traditional likelihood objective $p(R|C)$ with mutual information. The optimization of mutual information objective aims to achieve a Maximum Mutual Information (MMI). Specifically, the task is to find a best response $R$ based on the dialogue context $C$, in order to maximize their mutual information: 
\begin{equation}
\begin{split}
\hat{R} &= arg\max_{R}{log\frac{P(C,R)}{P(C)P(R)}} \\&= arg\max_{R}{logP(R|C)-logP(R)}
\end{split}
\label{logP(R|C)-logP(R)}
\end{equation} 
The objective $p(R|C)$ causes the model to choose responses with high probability even if the response is unconditionally frequent in the dataset, thus causing it to ignore the content of $C$. Maximizing the mutual information as Equation (\ref{logP(R|C)-logP(R)}) solves this issue by achieving a trade-off between safety and relativity. 

With a similar intuition as described above, increasing response diversity by modifying the decoding scheme at inference time has been explored in earlier works.~\cite{vijayakumar2016diverse} combined a dissimilarity term into the beam search objective and proposed Diverse Beam Search (DBS) to promote diversity. Similarly,~\cite{Shao2017GeneratingLA} proposed a stochastic beam search algorithm by performing stochastic sampling when choosing top-B responses. In the beam search algorithm, siblings sharing the same parent nodes tended to guide to similar sequences. Inspired by this,~\cite{li2016simple} penalized siblings sharing the same parent nodes using an additional term in the beam search objective. This encouraged the algorithm to search more diverse paths by expanding from different parent nodes. Some works further added a reranking stage to select more diverse responses in the generated N-best list~\citep{li2015diversity, sordoni2015neural, Shao2017GeneratingLA}. 

A user message can be mapped into multiple acceptable responses, which is also known as the one-to-many mapping problem. \cite{qiu2019training} considered the one-to-many mapping problem in open-domain dialogue systems and proposed a two-stage generation model to increase response diversity - the first stage extracting common features of multiple ground truth responses and the second stage extracting the distinctive ones. \cite{ko2020generating} solved the one-to-many mapping problem via a classification task to learn latent semantic representations. So that given one example response, different ones could be generated by exploring the semantically close vectors in the latent space. 

Different training strategies have been proposed to increase response diversity. \cite{bao2019know} used human instinct or pre-defined objective as a reward signal in a reinforcement learning setting to prompt the agent to avoid generating dull responses. Still, in a reinforcement learning framework,~\cite{zhu2020counterfactual} performed counterfactual reasoning to explore the potential response space. Given a pre-existing response, the model inferred another policy, which represented another possible response, thus increasing the response diversity.~\cite{he2019negative} used a negative training method to minimize the generation of bland responses. They first collected negative samples and then gave negative training signals based on these samples to fine-tune the model, impeding the model to generate bland responses. To achieve a better performance,~\cite{du2019boosting} synthesized different dialogue models designed for response diversity based on boosting training. The ensemble model significantly outperformed each of its base models. 

Utilizing external knowledge sources is another way to improve the diversity of generated responses because it can enrich the content. \cite{wu2020diverse} built a common-sense dialogue generation model which seeks highly related knowledge facts based on the dialogue history. Likewise,~\cite{su2020diversifying} incorporated external knowledge sources to diversify the response generation, but the difference was that they utilized non-conversational texts like news articles as relevant knowledge facts, which were obviously easier to obtain. \cite{tian2019learning} used a memory module to abstract and store useful information in the training corpus for generating diverse responses. 

Another approach to diversify the response generation is to make modifications to the training corpus. \cite{csaky2019improving} solved the challenge by filtering out the generic responses in the dataset using an entropy-based algorithm, which was simple but effective. Augmented with human feedback data,~\cite{gao2020dialogue} proposed that the generated responses could be reranked via a response ranking framework trained on the human feedback data and responses with higher quality including diversity were selected. \cite{stasaski2020more} proposed to change the data collection pipeline by iteratively computing the diversity of responses from different human participants in dataset construction and selected those participants who tend to generate informative and diverse responses. 

\subsection{Speaker Consistency and Personality-based Response}
\label{Speaker Consistency and Personality Response}
 
In open-domain dialogue systems, one big issue is that the responses are entirely learned from training data. The inconsistent response may be received when asking the system about some personal facts (e.g., age, hobbies). If the dataset contains multiple utterance pairs about the query of age, then the response generated tends to be shifting, which is unacceptable because personal facts are usually not random. Thus, for a data-driven chatbot, it is necessary to be aware of its role and respond based on a fixed persona. 

Explicitly modeling the persona is the main strategy in recent works.~\cite{liu2020you} proposed a persona-based dialogue generator consisting of a Receiver and a Transmitter. The receiver was responsible for modeling the interlocutor's persona through several turns' chat while Transmitter generated utterances based on the persona of agent and interlocutor, together with conversation content. The proposed model supported conversations between two persona-based chatbots by modeling each other's persona. Without training with additional Natural Language Inference labels,~\cite{kim2020will} built an imaginary listener following a normal generator, which reasoned over the tokens generated by the generator and predicted a posterior distribution over the personas in a certain space. After that, a self-conscious speaker generated tokens aligned with the predicted persona. Likewise,~\cite{boyd2020large} used an augmented GPT-2 to reason over the past conversations and model the target actor's persona, conditioning on which persona consistency was achieved. 

Responding with personas needs to condition on some persona descriptions. For example, to build a generous agent, descriptions like ``\textit{I am a generous person}" are needed as a part of the model input. However, these descriptions require hand-crafted feature design, which is labor intensive. \cite{madotto2019personalizing} proposed to use Model-Agnostic Meta-Learning (MAML) to adapt to new personas with only a few training samples and needed no persona description. \cite{majumder2020like} relied on external knowledge sources to expand current persona descriptions so that richer persona descriptions were obtained, and the model could associate current descriptions with some commonsense facts. 

~\cite{song2020generate} argued that traditional persona-based systems were one-stage systems and the responses they generated still contain many persona inconsistent words. To tackle this issue, they proposed a three-stage architecture to ensure persona consistency. A generate-delete-rewrite mechanism was implemented to remove the unacceptable words generated in prototype responses and rewrite them. 

\subsection{Empathetic Response}
\label{Empathetic Response}
  
Empathy means being able to sense other people's feelings~\citep{ma2020survey}. An empathetic dialogue system can sense the user's emotional changes and produce appropriate responses with a certain sentiment. This is an essential topic in chit-chat systems because it directly affects the user's feeling and to some extent decides the response quality. Industry systems such as Microsoft's Cortana, Facebook M, Google Assistant, and Amazon's Alexa are all equipped with empathy modules~\citep{wanrev}. 

There are two ways to generate utterances with emotion: one is to use explicit sentiment words as a part of input; another is to implicitly combine neural words~\citep{song2019generating}. \cite{song2019generating} proposed a unified framework that uses a lexicon-based attention to explicitly plugin emotional words and a sequence-level emotion classifier to classify the output sequence, implicitly guiding the generator to generate emotional responses through backpropagation. \cite{zhongtowards2020} used CoBERT for persona-based empathetic response selection and further investigated the impact of persona on empathetic responses. \cite{smith2020can} blended the skills of being knowledgeable, empathetic, and role-aware in one open-domain conversation model and overcame the bias issue when blending these skills. 

Since the available datasets for empathetic conversations are scarce,~\cite{rashkin2018towards} provided a new benchmark and dataset for empathetic dialogue systems. \cite{oraby2019curate} constructed a dialogue dataset with rich emotional markups from user reviews and further proposed a novel way to generate similar datasets with rich markups. 

\subsection{Controllable Generation}
\label{Controllable Generation}
Controllable dialogue generation is an important line of work in open-domain dialogue systems since solely learning from data sample distributions causes many uncertain responses. Some of the dialogue systems are grounded on some external knowledge such as knowledge graph and documents. However, grounding alone without explicit control and semantic targeting may induce output that is accurate but vague.

We may get some inspirations from the prior work on language generation and machine translation since similarly to dialogue systems they are generation-based or seq-to-seq problems. Some related work aimed to enforce user-specified constraints, most notably using lexical constraints \citep{hokamp2017lexically, hu2019parabank, miao2019cgmh}. These methods exclusively use constraints at inference time. Constraints can be included into the latent space during training, resulting in better predictions. Other studies \citep{see2019makes, keskar2019ctrl, tang2019target} have looked at non-lexical constraints, but they haven't looked into how they can help with grounding external knowledge. These publications also assume that the system can always be given (gold) constraints, which limits the ability to demonstrate larger benefits of the approaches.

Controllable text generation has also been used to extract high-level style information from contextual information in text style transfer \citep{hu2017toward} and other tasks \citep{ficler2017controlling, dong2017learning, gao2019structuring}, allowing the former to be independently modified. \cite{zhao2018unsupervised} learns an interpretable representation for dialogue systems using discrete latent actions. While existing studies employ ``style" descriptors (e.g., positive/negative, formal/informal) as control signals, \cite{wu2020controllable} use specific lexical constraints to regulate creation, allowing for finer semantic control. Content planned generation \citep{wiseman2017challenges, hua2019sentence} focuses response generation on a small number of essential words or table entries. This line of work, on the other hand, does not require consideration of the discourse context, which is critical for response generation.

\subsection{Conversation Topic}
\label{Conversation Topic}

Daily chats of people usually involve a topic or goal. Actually, a topic or goal is the key to keep each participant engaged in conversations and thus being essential to a chatbot. In real applications, a good topic model helps to retrieve related knowledge and guide the conversation instead of passively responding to the user's message~\citep{xing2017topic}. For example, if the user mentions ``\textit{I like sunny days}", a topic-aware system may reason over relevant external knowledge and produce responses like ``\textit{I know there is a nice park near the seaside, have you ever been there before?}". Thus, the agent pushes the conversation to a more engaging stage and enriches the dialogue content. 

Almost all topic-aware dialogue agents need to model explicit topics, which can be entities from external knowledge-base, or topic embeddings that have some semantic meaning.~\cite{wu2019proactive} tried to change the traditional passive response fashion and radically pursue active guidance of conversation. The dialogue agent consists of a leader and a follower, where the leader reasons over a knowledge graph and decides the conversation topic. Likewise, a common-sense knowledge graph was used by~\cite{liu2020towards} to lead the conversation topic and make recommendations. \cite{tang2019target} built a topic-aware retrieval-based chatbot. It aimed to guide the conversation topic to the target one step by step. It used a keyword predictor to predict turn-level keywords and selected the discourse-level keyword based on that. The discourse-level keyword was further fed into the retrieval model to retrieve responses regarding a certain topic. \cite{chen2020multi} built a multi-view sequence-to-sequence model to learn dialogue topics by first extracting dialogue structures of unstructured chit-chat dialogues, then generating topic summaries using BART decoder. 

In some applications of certain scenarios the conversation topic is essential, and these are where the topic-aware dialogue agents can be applied to.~\cite{zhang2020balancing} studied the topic-aware chatbot in counseling conversations. In counseling conversations, the agent led the dialogue topic by deciding between empathetically addressing a situation within the current range and moving on to a new target resolution. \cite{cao2019observing} studied chatbots in the psychotherapy treatment area and built a topic prediction model to forecast the behavior codes for upcoming conversations, thus guiding the dialogue. 

\subsection{Knowledge-Grounded System}
\label{Knowledge-Grounded System}
External knowledge such as common-sense knowledge is a significant source of information when organizing an utterance. Humans associate current conversation context with their experiences and memories and produce meaningful related responses, such capability results in the gap between human and machine chit-chat systems. As discussed, the earlier chit-chat systems are simply variants of machine translation systems, which can be viewed as sequence-to-sequence language models. However, dialogue generation is much more complicated than machine translation because of the higher freedom and vaguer constraints. Thus, chit-chat systems cannot simply consist of a sequence-to-sequence mapping since appropriate and informative responses are always related to some external common-sense knowledge. Instead, there must be a module incorporating world knowledge. 

Many researchers devoted their research efforts to building knowledge-grounded dialogue systems. A representative model is memory networks introduced in Section~\ref{Memory Networks}. Knowledge grounded systems use Memory Networks to store external knowledge and the generator retrieves relevant knowledge facts from it at the generation stage~\citep{ghazvininejad2018knowledge, vougiouklis-etal-2016-neural, yin2015neural}.~\cite{tian2019learning} built a memory-augmented conversation model. The proposed model abstracted from the training samples and stored useful ones in the memory module. \cite{zhao2020knowledge} built a knowledge-grounded dialogue generation system based on GPT-2. They combined a knowledge selection module into the language model and learned knowledge selection and response generation simultaneously.~\cite{lin2020generating} proposed Knowledge-Interaction and knowledge Copy (KIC). They performed recurrent knowledge interactions during the decoding phase to compute an attention distribution over the memory. Then they performed knowledge copy using a knowledge-aware pointer network to copy knowledge words according to the attention distribution computed. 

Documents contain large amount of knowledge facts, but they have a drawback that they are usually too long to retrieve useful information from \citep{li2019incremental}.~\cite{li2019incremental} built a multi-turn document-grounded system. They used an incremental transformer to encode multi-turns' dialogue context and respective documents retrieved. In the generation phase, they designed a two-stage generation scheme. The first stage took dialogue context as input and generated coherent responses; the second stage utilized both the utterance from the first stage and the document retrieved for the current turn for response generation. In this case, selecting knowledge based on both dialogue context and generated response was called posterior knowledge selection, while selecting knowledge with only dialogue context was called prior knowledge selection, which only utilized prior information. \cite{wang2020continuity} built a document quotation model in online conversations and investigated the consistency between quoted sentences and latent dialogue topics. 

Knowledge graph is another source of external information, which is becoming more and more popular in knowledge-grounded systems because of their structured nature. \cite{jung2020attnio} proposed a dialogue-conditioned graph traversal model for knowledge-grounded dialogue systems. The proposed model leveraged attention flows of two directions and fully made use of the structured information of knowledge graph to flexibly decide the expanding range of nodes and edges. Likewise,~\cite{zhang2019grounded} applied graph attention to traverse the concept space, which was a common-sense knowledge graph. The graph attention helped to move to more meaningful nodes conditioning on dialogue context. \cite{xu2020conversational} applied knowledge graphs as an external source to control a coarse-level utterance generation. Thus, the conversation was supported by common-sense knowledge, and the agent guided the dialogue topic in a more reasonable way. \cite{moon2019opendialkg} built a retrieval system retrieving responses based on the graph reasoning task. They used a graph walker to traverse the graph conditioning on symbolic transitions of the dialogue context. \cite{huang2020grade} proposed Graph-enhanced Representations for Automatic Dialogue Evaluation (GRADE), a novel evaluation metric for open-domain dialogue systems. This metric considered both contextualized representations and topic-level graph representations. The main idea was to use an external knowledge graph to model the conversation logic flow as a part of the evaluation criteria. 

Knowledge-grounded datasets containing context-knowledge-response triples are scarce and hard to obtain. \cite{cho2020grounding} collected a large dataset consisting of more than 26000 turns of improvised dialogues which were further grounded with a larger movie corpus as external knowledge. Also tackling the data insufficiency problem,~\cite{li2020zero} proposed a method that did not require context-knowledge-response triples for training and was thus data-efficient. They viewed knowledge as a latent variable to bridge the context and response. The variational approach learned the parameters of the generator from both a knowledge corpus and a dialogue corpus which were independent of each other. 

\subsection{Interactive Training}
\label{Interactive Training}

Interactive training, also called human-in-loop training, is a unique training method for dialogue systems. Annotated data is fixed and limited, not being able to cover all dialogue settings. Also, it takes a long time to train a good system. But in some industrial products, the dialogue systems need not be perfect when accomplishing their tasks. Thus, interactive training is desirable because the dialogue systems can improve themselves via interactions with users anywhere and anytime, which is a more flexible and cheap way to finetune the parameters. 

Training schemes with the above intuition have been developed in recent years.~\cite{li2016dialogue} introduced a reinforcement learning-based online learning framework. The agent interacted with a human dialogue partner and the partner provided feedback as a reward signal. \cite{asghar2016deep} first trained the agent with two-stage supervised learning, and then used an interaction-based reinforcement learning to finetune. Every time the user chose the best one from K responses generated by the pretrained model and then responded to this selected response. Instead of learning through being passively graded,~\cite{li2016learning} proposed a model that actively asked questions to seek improvement. Active learning was applicable to both offline and online learning settings. \cite{hancock2019learning} argued that most conversation samples an agent saw happened after it was pretrained and deployed. Thus, they proposed a framework to train the agent from the real conversations it participated in. The agent evaluated the satisfaction score of the user from the user's response of each turn and explicitly requested the user feedback when it thought that a mistake has been made. The user feedback was further used for learning. \cite{bouchacourt2019miss} placed the interactive learning in a cooperative game and tried to learn a long-term implicit strategy via Reinforce algorithm. Some of these work has been adopted by industry products and is a very promising direction for study.

\subsection{Visual Dialogue}
\label{Visual Dialogue}

More and more researchers cast their eyes to a broader space and are not only restricted to NLP. The combination of CV and NLP giving rise to tasks like visual question answering attracted lots of interest. The VQA task is to answer a question based on the content of a picture or video. Recently, this has evolved into a more challenging task: visual dialogue, which conditions a dialogue on the visual information and dialogue history. The dialogue consists of a series of queries, and the query form is usually more informal, which is why it is more complicated than VQA. 

\begin{figure}[ht]
\begin {center}
\includegraphics[width=1.0\textwidth]{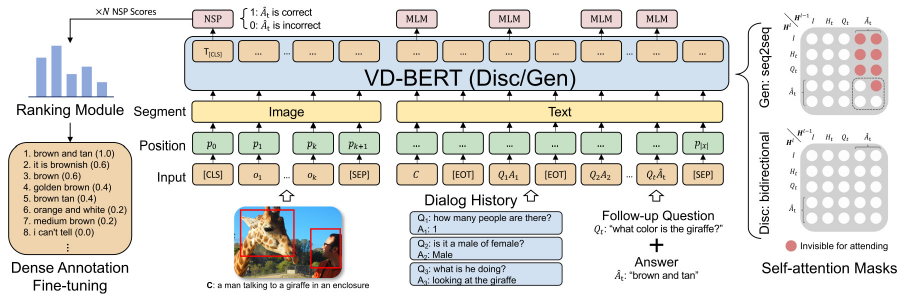}
\caption{The architecture of VD-BERT, a state-of-the-art visual dialogue system~\citep{wang2020vd}}
\label{VD-BERT}
\end {center}
\end{figure} 

\begin{figure}[ht]
\begin {center}
\includegraphics[width=1.0\textwidth]{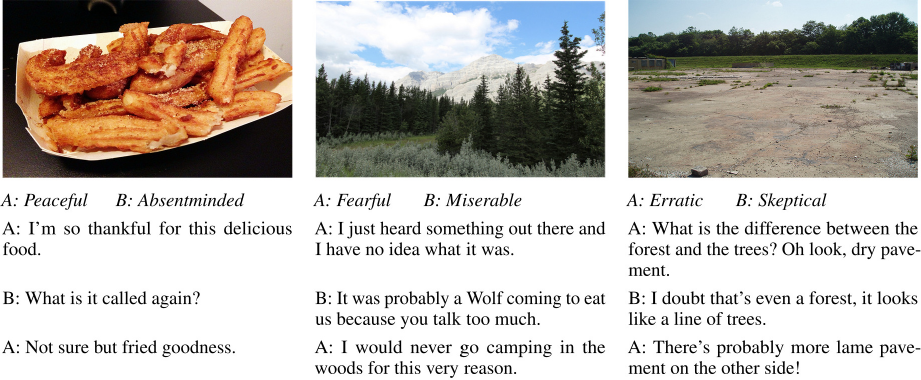}
\caption{ Three samples from the IMAGE-CHAT dataset~\citep{shuster-etal-2020-image}}
\label{IMAGECHAT}
\end {center}
\end{figure} 

Visual dialogue can be seen as a multi-step reasoning process over a series of questions \citep{gan2019multi}. \cite{gan2019multi} learned semantic representation of the question based on dialogue history and a given image, and recurrently updated the representation. \cite{shuster2019dialogue} proposed a set of image-based tasks and provided strong baselines. \cite{wang2020vd} employed R-CNN as an image encoder and fused the visual and dialogue modality with a VD-BERT. The proposed architecture achieved sufficient interactions between multi-turn dialogue and images. The proposed architecture is shown as an example model for Visual Dialogue tasks in Figure~\ref{VD-BERT}.

Compared with image-grounded dialogue systems, video-grounded systems are more interesting but also more challenging. There are two main challenges of video dialogue, as claimed by~\cite{le2020video}. One is that both spatial and temporal features exist in the video, which increases the difficulty of feature extraction. Another is that video dialogue features span across multiple conversation turns and thus are more complicated. A GPT-2 model was applied by~\cite{le2020video}, being able to fuse multi-modality information over different levels. Likewise,~\cite{le2019multimodal} built a multi-modal transformer network to incorporate information from different modalities and further applied a query-aware attention to extract context-related features from non-text modalities. \cite{le2020bist} proposed a Bi-directional Spatio-Temporal Learning (BiST) leveraging temporal-to-spatial and spatial-to-temporal reasoning process and could adapt to the dynamically evolving semantics in the video. 

Some researchers hold different opinions on the effectiveness of dialogue history in visual dialogue. \cite{takmaz2020refer} proposed that many expressions were already mentioned in previous turns and they built a visual dialogue model grounded on both image and conversation history. They further proved that better performance was achieved when grounding the model on dialogue context. However,~\cite{agarwal2020history} argued that though with dialogue history the visual dialogue model could achieve better results, in fact only a small proportion of cases benefited from the history. Furthermore, they proved that existing evaluation metrics for visual dialogue promoted generic responses. 

The visual dialogue task benefits a lot from the pretraining-based learning. The popularity of NLP pretraining sparked interest in multi-modal pretraining. VideoBERT \citep{sun2019videobert} is widely recognized as the pioneering work in the field of multimodal pretraining. It's a model that's been pre-trained on video frame features and text. CBT \citep{sun2019learning}, which is similarly pretrained on video-text pairs, is a contemporary work of VideoBERT. For video representation learning, \cite{miech2020end} used unlabeled narrated films. More researchers have focused their attention on visual-linguistic pretraining, inspired by the early work in multi-modal pretraining. For this objective, there are primarily two types of model designs. The single-stream model \citep{alberti2019fusion, chen2019uniter, gan2020large, li2020unicoder, li2019visualbert, li2020oscar, su2019vl, zhou2020unified} is one example. \citep{li2020unicoder} used a BERT model to process the concatenation of objects and words and pre-trained it with three standard tasks. Similar methods were proposed by \cite{chen2019uniter} and \cite{qi2020imagebert}, but with more pretraining tasks and larger datasets. With an adversarial training technique, \cite{gan2020large} further enhanced the model. \cite{su2019vl} employed the same architecture, but incorporated single-modal data and pre-trained the object detector. Instead of using recognized objects, \cite{huang2020pixel} sought to enter pixels directly. The object labels were used by \cite{li2020oscar} to improve cross-modal alignment. \cite{zhou2020unified} suggested a single-stream model that learns both caption generation and VQA tasks at the same time. The two-stream model \citep{lu2019vilbert, lu202012, tan2019lxmert, yu2020ernie} is another type of model architecture. \cite{tan2019lxmert} suggested a two-stream model with co-attention and solely used in-domain data to train the model. \cite{lu2019vilbert} introduced a similar architecture with a more complex co-attention model, which they pretrained with out-of-domain data, and \cite{lu202012} improved VilBERT with multi-task learning. \cite{yu2020ernie} recently added the scene graph to the model, which improved performance. Aside from these studies, \cite{singh2020we} looked at the impact of pretraining dataset selection on downstream task performance.

The annotation of visual dialogue is laborious and thus the datasets are scarce. Recently, some researchers have tried to tackle the data insufficiency problem. \cite{shuster-etal-2020-image} collected a dataset (IMAGE-CHAT, shown in Figure~\ref{IMAGECHAT}) of image-grounded human-human conversations in which speakers are asked to perform role-playing based on an emotional mood or style offered, since the usage of such characteristics is also a significant factor in engagingness. \cite{kamezawa2020visually} constructed a visual-grounded dialogue dataset. Interestingly, it additionally annotated the eye-gaze locations of the interlocutor in the image to provide information on what the interlocutor was paying attention to. \cite{cogswell2020dialog} proposed a method to utilize the VQA data when adapting to a new task, minimizing the requirement of dialogue data which is expensive to annotate.

\section{Evaluation Approaches}
\label{Evaluation Approaches}

Evaluation is an essential part of research in dialogue systems. It is not only a way to assess the performance of agents, but it can also be a part of the learning framework which provides signals to facilitate the learning~\citep{bao2019know}. This section discusses the evaluation methods in task-oriented and open-domain dialogue systems. 

\subsection{Evaluation Methods for Task-oriented Dialogue Systems}
\label{Evaluation Methods for Task-oriented Dialogue Systems}

Task-oriented systems aim to accomplish tasks and thus have more direct metrics evaluating their performance such as task completion rate and task completion cost. Some evaluation methods also involve metrics like BLEU to compare system responses with human responses, which will be discussed later. In addition, human-based evaluation and user simulators are able to provide real conversation samples. 

Task Completion Rate is the rate of successful events in all task completion attempts. It measures the task completion ability of a dialogue system. For example, in movie ticket booking tasks, the Task Completion Rate is the fraction of dialogues that meet all requirements specified by the user, such as movie time, cinema location, movie genre, etc. The task completion rate was applied in many task-oriented dialogue systems~\citep{walker1997paradise, williams2007partially, peng2017composite}. Additionally, some works~\citep{singh2002optimizing, yih2015deep} used partial success rate. 

Task Completion Cost is the resources required when completing a task. Time efficiency is a significant metric belonging to Task Completion Cost. In dialogue-related tasks, the number of conversation turns is usually used to measure the time efficiency and dialogue with fewer turns is preferred when accomplishing the same task. 

Human-based Evaluation provides user dialogues and user satisfaction scores for system evaluation. There are two main streams of human-based evaluation. One is to recruit human labor via crowdsourcing platforms to test-use a dialogue system. The crowdsource workers converse with the dialogue systems about predefined tasks and then metrics like Task Completion Rate and Task Completion Cost can be calculated. Another is computing the evaluation metrics in real user interactions, which means that evaluation is done after the system is deployed in real use. 

User Simulator provides simulated user dialogues based on pre-defined rules or models. Since recruiting human labor is expensive and real user interactions are not available until a mature system is deployed, user simulators are able to provide task-oriented dialogues at a lower cost. There are two kinds of user simulators. One is agenda-based simulators~\citep{schatzmann2009hidden, li2016user, ultes2017pydial}, which only feed dialogue systems with the pre-defined user goal as a user message, without surface realization. Another is model-based simulators~\citep{chandramohan2011user, asri2016sequence}, which generate user utterances using language models given constraint information. 

\subsection{Evaluation Methods for Open-domain Dialogue Systems}
\label{Evaluation Methods for Open-domain Dialogue Systems}

Evaluation of open-domain dialogue systems has long been a challenging problem. Unlike task-oriented systems, there is no clear metric like task completion rate or task completion cost. Both human and automatic evaluation methods are developed for ODD during these years. Human evaluation has been adopted by many works \citep{ritter2011data, shang2015neural, sordoni2015neural} to converse with and rate dialogue agents. However, human evaluation is not an ideal approach for that human labor is expensive and the evaluation results are highly subjective, varying from person to person. Researchers tend to hire crowd source workers \citep{ritter2011data, shang2015neural, sordoni2015neural} or random people \citep{moon2019opendialkg, jung2020attnio} to conduct human evaluation, both of which have two main drawbacks: 1. The evaluator group is highly random, and there exists huge gap between people with different knowledge levels or from different domains. 2. Though individual bias could be weakened by increasing the number of evaluators, the evaluator group cannot be very large because of the limited budgets (in articles mentioned above the sizes of human evaluator groups are usually 5-20). Thus, automatic and objective metrics are desirable. In general, there are two categories of automatic metrics in recent research: word-overlap metrics and neural metrics. 

Word-overlap Metrics are widely used in Machine Translation and Summarization tasks, which calculate the similarity between the generated sequence and the ground truth sequence. Representative metrics like BLEU~\citep{papineni2002bleu} and ROUGE~\citep{lin2004rouge} are n-gram matching metrics. METEOR~\citep{banerjee2005meteor} was further proposed with an improvement based on BLEU. It identified the paraphrases and synonyms between the generated sequence and the ground truth. \cite{galley2015deltableu} extended the BLEU by exploiting numerical ratings of responses. \cite{liu2016not} argued that word-overlap metrics were not correlated well with human evaluation. These metrics are effective in Machine Translation because each source sentence has a ground truth to compare with, whereas in dialogues there may be many possible responses corresponding with one user message, and thus an acceptable response may receive a low score if simply computing word-overlap metrics. 

Neural Metrics are metrics computed by neural models. Neural methods improve the evaluation effectiveness in terms of adaptability compared with word-overlap metrics, but they require an additional training process. \cite{su2015learning} used an RNN and a CNN model to extract turn-level features in a sequence and give the score. \cite{tao2018ruber} proposed Ruber, which was an automatic metric combining referenced and unreferenced components. The referenced one computed the similarity between generated response representations and ground truth representations, while the unreferenced one learned a scoring model to rate the query-response pairs. \cite{lowe2017towards} learned representations of dialogue utterances using an RNN and then computed the dot-product between generated response and ground truth response as an evaluation score. \cite{kannan2017adversarial} and~\cite{bruni2017adversarial} used the discriminator of a GAN framework to distinguish the generated responses from human responses. If a generated response achieved a high confidence score, this was indicative of a human-like response, thus desirable. 

Evaluation of open-domain dialogue systems is a hot topic at present and many researchers cast their eyes on this task recently. Some papers introduce two or more custom evaluation metrics for better evaluation, such as response diversity, response consistency, naturalness, knowledgeability, understandability, etc., to study "what to evaluate". \cite{bao2019know} evaluated the generated responses by designing two metrics. One was the informativeness metric calculating information utilization over turns. Another was the coherence metric, which was predicted by GRUs, given the response, context, and background as input. Likewise,~\cite{akama2020filtering} designed scoring functions to compute connectivity of utterance pairs and content relatedness as two evaluation metrics and used another fusion function to combine the metrics. \cite{pang2020towards} combined four metrics in their automatic evaluation framework: the context coherence metric based on GPT-2; phrase fluency metric based on GPT-2; diversity metric based on n-grams; logical self-consistency metric based on textual-entailment-inference. \cite{mehri2020usr} proposed a reference-free evaluation metric. They annotated responses considering the following qualities: Understandable (0-1), Maintains Context (1-3), Natural (1-3), Uses Knowledge (0-1), Interesting (1-3), Overall Quality (1-5). Furthermore, a transformer was trained on these annotated dialogues to compute the score of quality. 

Apart from "what to evaluate", there are also a multitude of papers studying "how to evaluate", which focus more on refining the evaluation process.~\cite{liang2020beyond} proposed a three-stage framework to denoise the self-rating process. They first performed dialogue flow anomaly detection via self-supervised representation learning, and then the model was fine-tuned with smoothed self-reported user ratings. Finally, they performed a denoising procedure by calculating the Shapley value and removed the samples with negative values. \cite{zhao2020designing} trained RoBERTa as a response scorer to achieve reference-free and semi-supervised evaluation. \cite{sato2020evaluating} constructed a test set by first generating several responses based on one user message and then human evaluation was performed to annotate each response with a score, where the response with the highest score was taken as a true response and the remainder taken as false responses. Dialogue systems were further evaluated by comparing the response selection accuracy on the test set, where a cross-entropy loss was calculated between the generated response and candidate responses to perform the selection operation. Likewise,~\cite{sinha2020learning} trained a BERT-based model to discriminate between true and false responses, where false responses were automatically generated. The model was further used to predict the evaluation score of a response based on dialogue context. \cite{huang2020grade} argued that responses should not be simply evaluated based on their surface-level features, and instead the topic-level features were more essential. They incorporated a common-sense graph in their evaluation framework to obtain topic-level graph representations. The topic-level graph representation and utterance-level representation were jointly considered to evaluate the coherence of responses generated by open-domain dialogue systems. 

Ranking is also an approach that evaluates dialogue systems effectively.~\cite{gao2020dialogue} leveraged large-scale human feedback data such as upvotes, downvotes, and replies to learn a GPT-2-based response ranker. Thus, responses were evaluated by their rankings given by the ranker. \cite{deriu2020spot} also evaluated the dialogue systems by ranking. They proposed a low-cost human-involved evaluation framework, in which different conversational agents conversed with each other and the human's responsibility was to annotate whether the generated utterance was human-like or not. The systems were evaluated by comparing the number of turns their responses were judged as human-like responses.

\section{Datasets}
\label{Data sets}

The dataset is one of the most essential components in dialogue systems study. Nowadays the datasets are not enough no matter for task-oriented or open-domain dialogue systems, especially for those tasks requiring additional annotations \citep{novikova2017e2e}. For task-oriented dialogue systems, data can be collected via two main methods. One is to recruit human labor via crowdsourcing platforms to produce dialogues in a given task. Another is to collect dialogues in real task completions like film ticket booking. For open-domain dialogue systems, apart from dialogues collected in real interactions, social media is also a significant source of data. Some social media companies such as Twitter and Reddit provide API access to a small proportion of posts, but these services are restricted by many legal terms which affect the reproducibility of research. As a result, many recent works in dialogue systems collect their own datasets for train and test. 

In this section, we review and categorize these datasets and make a comprehensive summary. To our best knowledge, Table~\ref{Datasets for Task-oriented dialogue systems} and~\ref{Datasets for Open-domain dialogue systems} cover almost all available datasets used in recent task-oriented or open-domain dialogue systems. 

\subsection{Datasets for Task-oriented Dialogue Systems}
\begin{longtable}{>{\centering\arraybackslash}y{2.5cm} y{5cm} >{\centering\arraybackslash}y{2cm} >{\centering\arraybackslash}y{2.5cm}}
\caption{Datasets for Task-oriented dialogue systems}
\label{Datasets for Task-oriented dialogue systems} 
\\\hline
\textbf{Name}	&	\centering\arraybackslash\textbf{Description}	&	\centering\arraybackslash\textbf{Task}	&	 \textbf{Origin}	 \\\hline
Schema	&	A dataset mainly for dialogue state tracking.	&	Dialogue State Tracking	&	~\cite{rastogi2020towards}	 \\\hline
MetaLWOZ	&	Collected by crowdsourcing platforms, spanning over 227 tasks and 47 domains. This dataset is designed for learning in unseen domains.	&	Domain Transfer	&	~\cite{lee2019multi}	 \\\hline
E2E	&	A dataset for end-to-end dialogue generation in restaurant domain. Data is collected in crowdsourced fashion.	&	End-to-end Task-oriented Dialogue Systems	&	~\cite{novikova2017e2e}	 \\\hline
MSR-E2E	&	Contain dialogues spanning over 3 domains: movie-ticket booking, restaurant reservation, and taxi booking.	&	End-to-end Task-oriented Dialogue Systems	&	~\cite{li2018microsoft}	 \\\hline
YELPNLG	&	A corpus consisting of utterances spanning over different restaurant attributes.	&	Natural Language Generation	&	~\cite{oraby2019curate}	 \\\hline
Clinical Conversation data set	&	It consists of conversations between physicians and participants.	&	Natural Language Understanding	&	~\cite{du2019extracting}	 \\\hline
OOS	&	A large-scale dataset for intent detection.	&	Natural Language Understanding	&	~\cite{larson2019evaluation}	 \\\hline
ATIS	&	A dataset consisting of voice calls from people who intend to make flight reservations.	&	Natural Language Understanding; Dialogue State Tracking	&	~\cite{tur2010left}	 \\\hline
MultiWOZ	&	Human-human written conversations with rich annotations spanning over multi-domains.	&	Task-oriented Dialogue	&	~\cite{budzianowski2018multiwoz}	 \\\hline
SNIPS-NLU	&	Task-oriented dialogue dataset colleted in a crowdsourced fashion. It was used to train voice assistant agents.	&	Task-oriented Dialogue	&	\url{https://github.com/snipsco/nlubenchmark}	 \\\hline
bAbI	&	Restaurant table reservation dialogues.	&	Task-oriented Dialogue	&	~\cite{bordes2016learning}	 \\\hline
JDC	&	A Chinese customer service dataset, consisting of context-response pairs.	&	Task-oriented Dialogue	&	\url{https://www.jddc.jd.com}	 \\\hline
UbuntuV2	&	It consists of dialogues collected via Ubuntu question-answering forum.	&	Task-oriented Dialogue	&	~\cite{lowe2015ubuntu}	 \\\hline
MICROSOFT DIALOGUE CHALLENGE data set	&	A task-oriented dataset collected via Amazon Mechanical Turk.	&	Task-oriented Dialogue	&	~\cite{li2018microsoft}	 \\\hline
WOZ	&	Task-oriented data collected in crowdsourced fashion.	&	Task-oriented Dialogue	&	~\cite{wen2016network}	 \\\hline
DSTC series	&	Multi-domain task-oriented dataset.	&	Task-oriented Dialogue	&	\url{https://www.microsoft.com/en-us/research/event/dialog-state-tracking-challenge/}	 \\\hline
SimDial	&	Simulated conversations spanning over multiple domains.	&	Task-oriented Dialogue	&	~\cite{zhao2018zero}	 \\\hline
SMD	&	Human-human dialogues in weather, navigation and scheduling domain.	&	Task-oriented Dialogue	&	~\cite{eric2017key}	 \\\hline
BANKING	&	Question-answer pairs with 77 categories in e-banking domain.	&	Task-oriented Dialogue	&	~\cite{henderson2019training}	 \\\hline
Weather forecast	&	A task-oriented dataset in the weather domain.	&	Task-oriented Dialogue	&	~\cite{balakrishnan2019constrained}	 \\\hline
MedDialog-(EN,CN)	&	Large scale dataset in medical domain consisting of conversations between doctors and patients	&	Task-oriented Dialogue	&	~\cite{he2020meddialog}	 \\\hline
CamRest	&	It consists of human-human multi-turn dialogues in restaurant domain.	&	Task-oriented Dialogue	&	~\cite{wen2016conditional}	 \\\hline
Taskmaster	&	Contain dialogues spanning over 6 domains. It has 22.9 average length of conversational turns.	&	Task-oriented Dialogue	&	~\cite{byrne2019taskmaster}	 \\\hline
Frames	&	Conversational dataset with annotations of semantic frame tracking.	&	Task-oriented Dialogue	&	~\cite{asri2017frames}	 \\\hline
JDDC	&	A Chinese customer service dataset, consisting of context-response pairs.	&	Task-oriented Dialogue	&	~\cite{chen2019jddc}	 \\\hline
Court Debate Dataset	&	A task-oriented dataset in judicial field containing court debate conversations.	&	Task-oriented Dialogue	&	~\cite{ji2020cross}	 \\\hline
TreeDST	&	A task-oriented dataset annotated with tree structured dialogue states and agent acts.	&	Task-oriented Dialogue	&	~\cite{cheng2020conversational}	 \\\hline
RiSAWOZ	&	Contain utterances for 12 domains, annotated with rich semantic information.	&	Task-oriented Dialogue	&	~\cite{quan2020risawoz}	 \\\hline
Cambridge Restaurant	&	A task-oriented dataset in restaurant booking field.	&	Task-oriented Dialogue	&	~\cite{wen2016network}	 \\\hline
SB-TOP	&	A task-oriented dataset with semantic parsing annotation. It spans over 4 domains: Reminder, Weather, Calling and Music.	&	Task-oriented Dialogue	&	~\cite{aghajanyan2020conversational}	 \\\hline
GSIM	&	A machine-machine task-oriented dataset. It covers two domains: restaurant table booking and movie ticket booking.	&	Task-oriented Dialogue	&	~\cite{shah2018building}	 \\\hline
SGD	&	A schema-guided dataset spanning over multiple domains.	&	Task-oriented Dialogue	&	~\cite{rastogi2020towards}	 \\\hline
cite-8K	&	A task-oriented dataset collected in restaurant booking calls.	&	Task-oriented Dialogue	&	~\cite{coope2020span}	 \\
\hline
\end{longtable}

\subsection{Datasets for Open-domain Dialogue Systems}
\begin{longtable}{>{\centering\arraybackslash}y{2.5cm} y{5cm} >{\centering\arraybackslash}y{2cm} >{\centering\arraybackslash}y{2.5cm}}
\caption{Datasets for Open-domain dialogue systems}
\label{Datasets for Open-domain dialogue systems} 
\\\hline
\textbf{Name}	&	\centering\arraybackslash\textbf{Description}	&	\centering\arraybackslash\textbf{Task}	&	 \textbf{Origin} \\\hline
Large-Scale Corpus for Conversation Disentanglement	&	A dataset consisting of messages annotated with reply-structure graphs for dialogue disentanglement.	&	Conversation Disentaglement	&	~\cite{kummerfeld2018large}	 \\\hline
DuConv	&	Collected in conversations between a conversation leader and a conversation follower. 	&	Conversation Topic	&	~\cite{wu2019proactive}	 \\\hline
PERSUASION FOR GOOD	&	A topic-oriented dataset annotated with persuasion strategies.	&	Conversation Topic	&	~\cite{wang2019persuasion}	 \\\hline
MutualFriends	&	A topic-oriented dataset based on bot-bot stratigical conversations.	&	Conversation Topic	&	~\cite{he2017learning}	 \\\hline
SAMSum	&	A large-scale dialogue summary dataset.	&	Conversation Topic	&	~\cite{gliwa2019samsum}	 \\\hline
OpenDialKG	&	It consists conversations between two agents and each dialogue corresponds with a knowledge graph path annotation.	&	Conversation Topic; Dialogue Reasoning	&	~\cite{moon2019opendialkg}	 \\\hline
doc2dial	&	A dataset consisting of conversations annotated with goals and accociated documents.	&	Conversation Topic; Knowledge-Grounded System	&	~\cite{feng2020doc2dial}	 \\\hline
DialEdit	&	A dataset constructed for image editing via conversational language instructions.	&	Conversational Image Editing	&	~\cite{manuvinakurike2018dialedit}	 \\\hline
CHART DIALOGS	&	A dataset containing dialogues describing matplotlib plot features.	&	Conversational Plotting	&	~\cite{shao2020chartdialogs}	 \\\hline
CONAN	&	A multilingual dataset for hate speech tackling.	&	Dialogue Classification	&	~\cite{chung2019conan}	 \\\hline
Dialogue NLI	&	A NLI dataset with sentences annotated with entailment (E), neutral (N), or contradiction (C).	&	Dialogue Inference	&	~\cite{welleck2018dialogue}	 \\\hline
MuTual	&	A dialogue reasoning dataset containing English listening comprehension exams.	&	Dialogue Reasoning	&	~\cite{cui2020mutual}	 \\\hline
RST-DT	&	It consists of samples from 385 news articles annotated with dialogue features.	&	Discourse Parsing	&	~\cite{carlson2002rst}	 \\\hline
NLPCC	&	A dataset consisting of emotional classification data.	&	Empathetic Response	&	\url{http://tcci.ccf.org.cn/nlpcc.php}	 \\\hline
MELD	&	A multi-party conversational dataset with emotion annotations.	&	Empathetic Response	&	~\cite{poria2018meld}	 \\\hline
EMPATHETIC DIALOGUES	&	A dataset containing conversations annotated with emotion labels.	&	Empathetic Response	&	~\cite{rashkin2018towards}	 \\\hline
IEMOCAP	&	Contain multi-party dialogues. Each dialogue is annotated with an emotion label.	&	Empathetic Response	&	~\cite{busso2008iemocap}	 \\\hline
EmoryNLP	&	Collected from Friends' TV series, annotated with emotion labels.	&	Empathetic Response	&	~\cite{zahiri2017emotion}	 \\\hline
MojiTalk	&	A largescale dataset collected from Twitter, including emojis.	&	Empathetic Response	&	~\cite{zhou2017mojitalk}	 \\\hline
CBET	&	A dialogue dataset annotated with nine emotion labels: surprise, anger, love, sadness, joy, fear, guilt, disgust and thankfulness	&	Empathetic Response	&	~\cite{yadollahi2017current}	 \\\hline
Stanford Politeness Corpus	&	A conversational dataset annotated with politeness labels.	&	Empathetic Response	&	~\cite{danescu2013computational}	 \\\hline
AIT-2018	&	Collected in SemEval-2018 Task 1: Affect in Tweets.	&	Empathetic Response	&	~\cite{mohammad2018semeval}	 \\\hline
EMOTyDA	&	A dataset containing short videos about multi-party conversations, each annotated with respective emotion.	&	Empathetic Response; Visual Dialogue	&	~\cite{saha2020towards}	 \\\hline
Wizard of Wikipedia	&	A large-sclale dataset consisting of conversations grounded with Wikipedia knowledge.	&	Knowledge-Grounded System	&	~\cite{dinan2018wizard}	 \\\hline
CMU DoG	&	A dataset consisting of conversations grounded with Wikipedia articles about popular movies.	&	Knowledge-Grounded System	&	~\cite{zhou2018dataset}	 \\\hline
Holl-E	&	Contain dialogues grounded with documents.	&	Knowledge-Grounded System	&	~\cite{moghe2018towards}	 \\\hline
Interview	&	A dataset containing multi-party conversations in the form of interviews.	&	Knowledge-Grounded System	&	~\cite{majumder2020interview}	 \\\hline
Curiosity	&	An open-domain dataset annotated with pre-existing user knowledge and dialogue acts, also grounding in Wikipedia.	&	Knowledge-Grounded System	&	~\cite{rodriguez2020information}	 \\\hline
KdConv	&	A chinese knowledge-grounded dialogue dataset. 	&	Knowledge-Grounded System	&	~\cite{zhou2020kdconv}	 \\\hline
ELI5	&	A QA dataset grounded with retrieved documents.	&	Knowledge-Grounded System	&	~\cite{fan2019eli5}	 \\\hline
Topical Chat	&	A knowledge-grounded dataset where the knowledge spans over eight different topics.	&	Knowledge-Grounded System; Conversation Topic	&	~\cite{gopalakrishnan2019topical}	 \\\hline
WHERE ARE YOU?	&	A dialogue dataset annotated with localization information.	&	Localization Dialogue	&	~\cite{hahn2020you}	 \\\hline
MMD	&	A multi-modal dataset consisting of dialogues between sales agents and shoppers.	&	Multi-modal Dialogue	&	~\cite{saha2018towards}	 \\\hline
OpenSubtitles	&	A multilingual dataset made up of movie captions, containing about 8 billion words.	&	Open-domain Dialogue	&	~\cite{tiedemann2012parallel}	 \\\hline
NTCIR	&	A social media dataset collected from Sina Weibo.	&	Open-domain Dialogue	&	\url{http://research.nii.ac.jp/ntcir/data/data-en.html}	 \\\hline
Twitter	&	A social media dataset collected from Twitter.	&	Open-domain Dialogue	&	\url{https://github.com/Marsan-Ma-zz/chat corpus}	 \\\hline
Douban Conversation Corpus	&	A social media dataset collected from Douban.	&	Open-domain Dialogue	&	~\cite{zhang2018modeling}	 \\\hline
E-commerce Dialogue Corpus	&	It consists of conversations between customers and customer service staff on Taobao.	&	Open-domain Dialogue	&	~\cite{zhang2018modeling}	 \\\hline
REDDIT	&	A social media dataset collected from REDDIT.	&	Open-domain Dialogue	&	~\cite{henderson2019repository}	 \\\hline
STC-SeFun	&	A social media dataset collected from Tieba, Zhidao, Douban and Weibo.	&	Open-domain Dialogue	&	~\cite{bi2019fine}	 \\\hline
DailyDialog	&	A dataset consisting of daily dialogues, annotated with conversation intention and emotion information.	&	Open-domain Dialogue	&	~\cite{li2017dailydialog}	 \\\hline
PDTB	&	Dialogue dataset annotated with discourse relations.	&	Open-domain Dialogue	&	~\cite{prasad2008penn}	 \\\hline
Luna	&	Dialogue dataset with Italian relation annotations.	&	Open-domain Dialogue	&	~\cite{tonelli2010annotation}	 \\\hline
Edina-DR	&	Dialogue dataset with English relation annotations, which is based on Luna data set.	&	Open-domain Dialogue	&	~\cite{ma2019implicit}	 \\\hline
Cornell Movie Dialog Corpus	&	A dialogue dataset collected via IMDB database.	&	Open-domain Dialogue	&	~\cite{danescu2011chameleons}	 \\\hline
Reddit Movie Dialogue Dataset	&	A movie dialogue dataset collected from Reddit.	&	Open-domain Dialogue	&	~\cite{liu2020mitigating}	 \\\hline
LIGHT	&	A dialogue dataset with configurable text adventure environment.	&	Open-domain Dialogue	&	~\cite{urbanek2019learning}	 \\\hline
This American Life	&	A media dialogue dataset collected in long-form expository podcast episodes.	&	Open-domain Dialogue	&	~\cite{mao2020speech}	 \\\hline
RadioTalk	&	A media dialogue dataset collected from radio transcripts.	&	Open-domain Dialogue	&	~\cite{beeferman2019radiotalk}	 \\\hline
French EPAC	&	A media dialogue dataset collected from news.	&	Open-domain Dialogue	&	~\cite{esteve2010epac}	 \\\hline
TREC Conversational Assistance 	&	An open-domain dataset spanning 30 conversation topics.	&	Open-domain Dialogue	&	~\cite{dalton2020trec}	 \\\hline
Search as a Conversation	&	A dataset for conversations with search engines.	&	Open-domain Dialogue	&	~\cite{ren2020conversations}	 \\\hline
Amazon Alexa Prize Competition	&	A dataset containing real-world conversations between Amazon Alexa customers and Gunrock, which is a champion chatbot.	&	Open-domain Dialogue	&	~\cite{ram2018conversational}	 \\\hline
SwitchBoard	&	An open-domain dataset containing English phone conversations.	&	Open-domain Dialogue	&	~\cite{jurafsky1997switchboard}	 \\\hline
Zhihu	&	A Chinese social media dataset with posts and comments.	&	Open-domain Dialogue	&	\url{https://www.zhihu.com}	 \\\hline
SPOLIN	&	A dataset containing yes-and conversations.	&	Open-domain Dialogue	&	~\cite{cho2020grounding}	 \\\hline
CRD3	&	A dataset collected in the role-playing game Dungeons and Dragons.	&	Open-domain Dialogue	&	~\cite{rameshkumar2020storytelling}	 \\\hline
Baidu Zhidao	&	A Chinese social media dataset with posts and comments.	&	Open-domain Dialogue	&	\url{https://zhidao.baidu.com/}	 \\\hline
Webis Gmane Email Corpus 2019	&	A conversational dataset collected from 153M emails.	&	Open-domain Dialogue	&	~\cite{bevendorff2020crawling}	 \\\hline
LibreSpeech Corpus	&	Contain 500 hours' speech produced by 1252 participants.	&	Open-domain Dialogue	&	~\cite{panayotov2015librispeech}	 \\\hline
Motivational Interviewing	&	A dialogue dataset about conversational psychotherapy.	&	Open-domain Dialogue	&	~\cite{tanana2016comparison}	 \\\hline
SubTle Corpus	&	Contact Ameixa for data.	&	Open-domain Dialogue	&	~\cite{lubis2018eliciting}	 \\\hline
TED-LIUM	&	TED-talk monologues.	&	Open-domain Dialogue	&	~\cite{fung2016zara}	 \\\hline
ECG NLPCC 2017 Data	&	Conversational dataset extracted from Weibo.	&	Open-domain Dialogue	&	~\cite{huang2018natural}	 \\\hline
SEMEVAL15	&	QA dataset with answer quality annotations via Amazon Mechanical Turk.	&	Question Answering	&	~\cite{nakov2019semeval}	 \\\hline
AMAZONQA	&	A QA dataset solving one-to-many problems.	&	Question Answering	&	~\cite{wan2016modeling}	 \\\hline
TGIF-QA	&	A video-grounded QA dataset.	&	Question Answering	&	~\cite{jang2017tgif}	 \\\hline
QuAC	&	A QA dataset with 14K QA dialogues.	&	Question Answering	&	~\cite{choi2018quac}	 \\\hline
SQuAD	&	A question-answering dataset collected in crowdsourced fashion.	&	Question Answering	&	~\cite{rajpurkar2018know}	 \\\hline
LIF	&	A dataset constructed based on QuAC.	&	Question Answering	&	~\cite{kundu2020learning}	 \\\hline
Yelp	&	It consists of customer reviews from Yelp Dataset Challenge	&	Response Retrieval	&	~\cite{tang2015learning}	 \\\hline
Debates	&	The dataset consists of debates on Congerssional bills.	&	Response Retrieval	&	~\cite{thomas2006get}	 \\\hline
PERSONACHAT	&	It provides profile information of the agents and background of users.	&	Speaker Consistency and Personality Response	&	~\cite{zhang2018personalizing}	 \\\hline
KvPI	&	Contain consistency annotations between response and corresponding key-value profiles.	&	Speaker Consistency and Personality Response	&	~\cite{song2020profile}	 \\\hline
ConvAI2	&	A dataset constructed on the base of Persona-Chat, each conversation having profiles from a set containing persona candidates.	&	Speaker Consistency and Personality Response	&	~\cite{dinan2019second}	 \\\hline
PEC	&	An open-domain dataset annotated with persona labels.	&	Speaker Consistency and Personality Response; Empathetic Response	&	~\cite{zhongtowards2020}	 \\\hline
GuessWhat?!	&	A visual dialogue dataset for a two-player game about object recognition.	&	Visual Dialogue	&	~\cite{de2017guesswhat}	 \\\hline
VisDial	&	A visual dialogue dataset whose images are obtained from COCO data set.	&	Visual Dialogue	&	\url{https://visualdialog.org/ data}	 \\\hline
AVSD	&	A video-grounded dialogue dataset.	&	Visual Dialogue	&	~\cite{yoshino20187th}	 \\\hline
VFD	&	A visual dialogue dataset annotated with unique eye-gaze locations.	&	Visual Dialogue	&	~\cite{kamezawa2020visually}	 \\\hline
PhotoBook	&	A dataset for task-oriented visual dialogues.	&	Visual Dialogue	&	~\cite{haber2019photobook}	 \\\hline
IGC	&	A dataset containing conversations discussing a given image.	&	Visual Dialogue	&	~\cite{mostafazadeh2017image}	 \\\hline
Image-Chat	&	Contain conversations grounded with images. The conversations are also annotated with personality.	&	Visual Dialogue; Speaker Consistency and Personality Response	&	~\cite{shuster2018image}	 \\

\hline

\end{longtable}

\section{Conclusions and Trends}
\label{Conclusions and Trends} 

More and more researchers are investigating conversational tasks. One factor contributing to the popularity of conversational tasks is the increasing demand for chatbots in industry and daily life. Industry agents like Apple's Siri, Microsoft's Cortana, Facebook M, Google Assistant, and Amazon's Alexa have brought huge convenience to people's lives. Another reason is that a considerable amount of natural language data is in the form of dialogues, which contributes to the efforts in dialogue research. 

In this paper we discuss dialogue systems from two perspectives: model and system type. Dialogue systems are a complicated but promising task because it involves the whole process of communication between agent and human. The works of recent years show an overwhelming preference towards neural methods, no matter in task-oriented or open-domain dialogue systems. Neural methods outperform traditional rule-based methods, statistical methods and machine learning methods for that neural models have the stronger fitting ability and require less hand-crafted feature engineering. 

We systematically summarized and categorized the latest works in dialogue systems, and also in other dialogue-related tasks. We hope these discussions and insights provide a comprehensive picture of the state-of-the-art in this area and pave the way for further research. Finally, we discuss some possible research trends arising from the works reviewed: 

\paragraph*{Multimodal dialogue systems} The world is multimodal and humans observe it via multiple senses such as vision, hearing, smell, taste, and touch. In a conversational interaction, humans tend to make responses not only based on text, but also on what they see and hear. Thus, some researchers argue that chatbots should also have such abilities to blend information from different modalities. There are some recent works trying to build multimodal dialogue systems~\citep{le2019multimodal, chauhan2019ordinal, saha2020towards, singla2020towards, young2020dialogue}, but these systems are still far from mature. 

\paragraph*{Multitask dialogue systems} Dialogue systems are categorized into task-oriented and open-domain systems. Such a research boundary has existed for a long time because task-oriented dialogue systems involve dialogue states, which constrain the decoding process. However, works in end-to-end task-oriented dialogue systems and knowledge-grounded open-domain systems provide a possibility of blending these two categories into a single framework, or even a single model. Such blended dialogue systems perform as assistants and chatbots simultaneously. 

\paragraph*{Corpus exploration on Internet} In Section~\ref{Data sets} we reviewed many datasets for dialogue systems training. However, data is still far from enough to train a perfect dialogue system. Many learning techniques are designed to alleviate this problem, such as reinforcement learning, meta-learning, transfer learning, and active learning. But many works ignore a significant source of information, which is the dialogue corpus on the Internet. There is a large volume of conversational corpus on the Internet but people have no access to the raw corpus because much of it is in a messy condition. In the future, dialogue agents should be able to explore useful corpus on the Internet in real-time for training. This can be achieved by standardizing online corpus access and their related legal terms. Moreover, real-time conversational corpus exploration can be an independent task that deserves study.

\paragraph*{User modeling} User modeling is a hot topic in both dialogue generation~\citep{gur2018user, serras2019goal} and dialogue systems evaluation~\citep{kannan2017adversarial}. Basically, the user modeling module tries to simulate the real decisions and actions of a human user. It makes decisions based on the dialogue state or dialogue history. In dialogue generation tasks, modeling the user helps the agent converse more coherently, based on the background information or even speaking habits. Besides that, a mature user simulator can provide an interactive training environment, which reduces the reliance on annotated training samples when training a dialogue system. In dialogue systems evaluation tasks, a user simulator provides user messages to test a dialogue agent. More recent user simulators also give feedback concerning the responses generated by the dialogue agent. However, user modeling is a challenging task since no matter explicit user simulation or implicit user modeling is actually the same in difficulty as response generation. Since response generation systems are not perfect yet, user modeling can still be a topic worthy of study.

\paragraph*{Dialogue generation with a long-term goal} Most of our daily conversations are chitchats without any purpose. However, there are quite a few scenarios when we purposely guide the conversation content to achieve a specific goal. Current open-domain dialogue systems tend to model the conversation without a long-term goal, which does not exhibit enough intelligence. There are some recent works that apply reinforcement policy learning to model a long-term reward 
which encourages the agent to converse with a long-term goal, such as the work of~\cite{xu2020conversational}. This topic will lead to strong artificial intelligence, which is useful in some real-life applications such as negotiation or story-telling chatbots.

\section*{Acknowledgements}
This research/project is supported by A*STAR under its Industry Alignment Fund (LOA Award I1901E0046).

%
%

\bibliographystyle{spbasic}   
\bibliography{Bib}  


\end{document}